\documentclass[11pt]{article}

\newif\ifreviewcopy
\ifdefined\bluereview\reviewcopytrue\fi
\ifreviewcopy
  \usepackage[review]{acl}
  \makeatletter\acl@anonymizefalse\makeatother
\else
  \usepackage[preprint]{acl}
\fi
\usepackage{amssymb}
\usepackage{times}
\usepackage{latexsym}
\usepackage{amsmath}
\usepackage{booktabs}
\usepackage{multirow}
\usepackage{subcaption}
\usepackage{algorithm}
\usepackage{algpseudocode}
\usepackage[T1]{fontenc}

\usepackage[utf8]{inputenc}

\usepackage{microtype}

\usepackage{inconsolata}

\usepackage{graphicx}
\newcommand{\RevisedGraphic}[3][]{\ifmarkchanges\includegraphics[#1]{#3}\else\includegraphics[#1]{#2}\fi}

\usepackage{etoolbox}
\usepackage{placeins}
\usepackage{needspace}
\usepackage{zref-savepos}
\definecolor{revisionblue}{RGB}{0,0,255}
\newif\ifmarkchanges
\ifreviewcopy\markchangestrue\fi
\newcommand{\revisioncolor}{\ifmarkchanges\color{revisionblue}\hypersetup{linkcolor=revisionblue,citecolor=revisionblue,urlcolor=revisionblue}\fi}

\DeclareRobustCommand{\Rev}[2]{\begingroup\revisioncolor\leavevmode\ChangeStart{#1}\ignorespaces#2\ChangeEnd{#1}\endgroup}
\DeclareRobustCommand{\RevNoLine}[2]{\begingroup\revisioncolor\zsavepos{#1-start}\label{#1-page}\ignorespaces#2\zsavepos{#1-end}\endgroup}
\newcommand{\RevFloatStart}[1]{\revisioncolor\ifmarkchanges\captionsetup{font+={color=revisionblue}}\fi\zsavepos{#1-start}\label{#1-page}}
\newcommand{\RevFloatEnd}[1]{\zsavepos{#1-end}}
\hypersetup{colorlinks=true,linkcolor=black,citecolor=black,urlcolor=black,
 pdftitle={Diagnosing Unsupervised RL for Compressing Mathematical Reasoning: Truncation Avoidance and Policy Collapse},
 pdfauthor={Zelin Zhang; Fei Cheng; Chenhui Chu},pdflang={en}}
\ifreviewcopy
 
\fi
\AtEndEnvironment{thebibliography}{\zsavepos{BIB-end-start}\label{BIB-end-page}}
\makeatletter
\AtBeginDocument{%
 \let\SACoriginalbibitem\bibitem
 \renewcommand{\bibitem}[2][]{%
  \ifboolexpr{test {\ifstrequal{#2}{unreasonable}} or test {\ifstrequal{#2}{tavenard2020tslearn}} or test {\ifstrequal{#2}{kimiteam2025kimik15scalingreinforcement}} or test {\ifstrequal{#2}{grpo}} or test {\ifstrequal{#2}{grattafiori2024llama3herdmodels}} or test {\ifstrequal{#2}{empo}} or test {\ifstrequal{#2}{ttrl}}}{\ifmarkchanges\color{revisionblue}\hypersetup{urlcolor=revisionblue}\fi}{\color{black}\hypersetup{urlcolor=black}}%
  \SACoriginalbibitem[#1]{#2}%
  \zsavepos{BIB-all-#2-start}\label{BIB-all-#2-page}%
  \ifstrequal{#2}{unreasonable}{\zsavepos{CR-bib-entry-start}\label{CR-bib-entry-page}}{}%
  \ifstrequal{#2}{tavenard2020tslearn}{\zsavepos{S1-dtw-reference-start}\label{S1-dtw-reference-page}}{}%
  \ifstrequal{#2}{kimiteam2025kimik15scalingreinforcement}{\zsavepos{CR-bib-kimi-start}\label{CR-bib-kimi-page}}{}%
  \ifstrequal{#2}{grpo}{\zsavepos{CR-bib-deepseekmath-start}\label{CR-bib-deepseekmath-page}}{}%
  \ifstrequal{#2}{empo}{\zsavepos{CR-bib-empo-start}\label{CR-bib-empo-page}}{}%
  \ifstrequal{#2}{ttrl}{\zsavepos{CR-bib-ttrl-start}\label{CR-bib-ttrl-page}}{}%
  \ifstrequal{#2}{grattafiori2024llama3herdmodels}{\zsavepos{CR-bib-llama-start}\label{CR-bib-llama-page}}{}%
 }}
\makeatother

\title{Diagnosing Unsupervised RL for Compressing Mathematical Reasoning: Truncation Avoidance and Policy Collapse}

\author{
  Zelin Zhang \hspace{3em} Fei Cheng \hspace{3em} Chenhui Chu \\
  Kyoto University \\
  \RevNoLine{CR-aff-1}{Research and Development Center for Large Language Models (LLMC),}%
   \\
  \RevNoLine{CR-aff-2}{National Institute of Informatics (NII)}%
   \\
  \texttt{zelin@nlp.ist.i.kyoto-u.ac.jp}, \texttt{\{feicheng, chu\}@i.kyoto-u.ac.jp}
}

\makeatletter
\AddToHook{shipout/after}{\ifnum\value{page}=2\global\let\@textbottom\relax\fi}
\makeatother
\begin{document}
\maketitle

\begin{abstract}
Compressing mathematical reasoning within a fixed response budget is a practical but underexplored setting, where responses that reach the generation limit without an EOS token are incomplete. We study whether unsupervised reinforcement learning (RL) with intrinsic length- and entropy-based rewards can improve this fixed-budget outcome. We operationalize compression through \textit{truncation avoidance}: the EOS-verified truncation rate decreases while mathematical accuracy is maintained or improved. Across four Large Language Models (LLMs), a direct length penalty (LP) and a cumulative token-entropy penalty (Ent) can produce a transient compression regime before policy collapse. 
\Rev{S2-abstract-scope}{Across the evaluated models, higher baseline mathematical performance is generally associated with a longer regime and later collapse.}%
{}  We further cluster token-entropy trajectories to construct a response-level entropy space. The expansion rate of its explored region increases near policy collapse, 
\Rev{S2-abstract-geometry}{with generally later and smaller increases for models with higher baseline mathematical performance.}%
{} 
\end{abstract}

\begin{figure*}[!t]
    \centering
    \includegraphics[width=\linewidth,trim=24bp 0bp 30bp 0bp,clip]{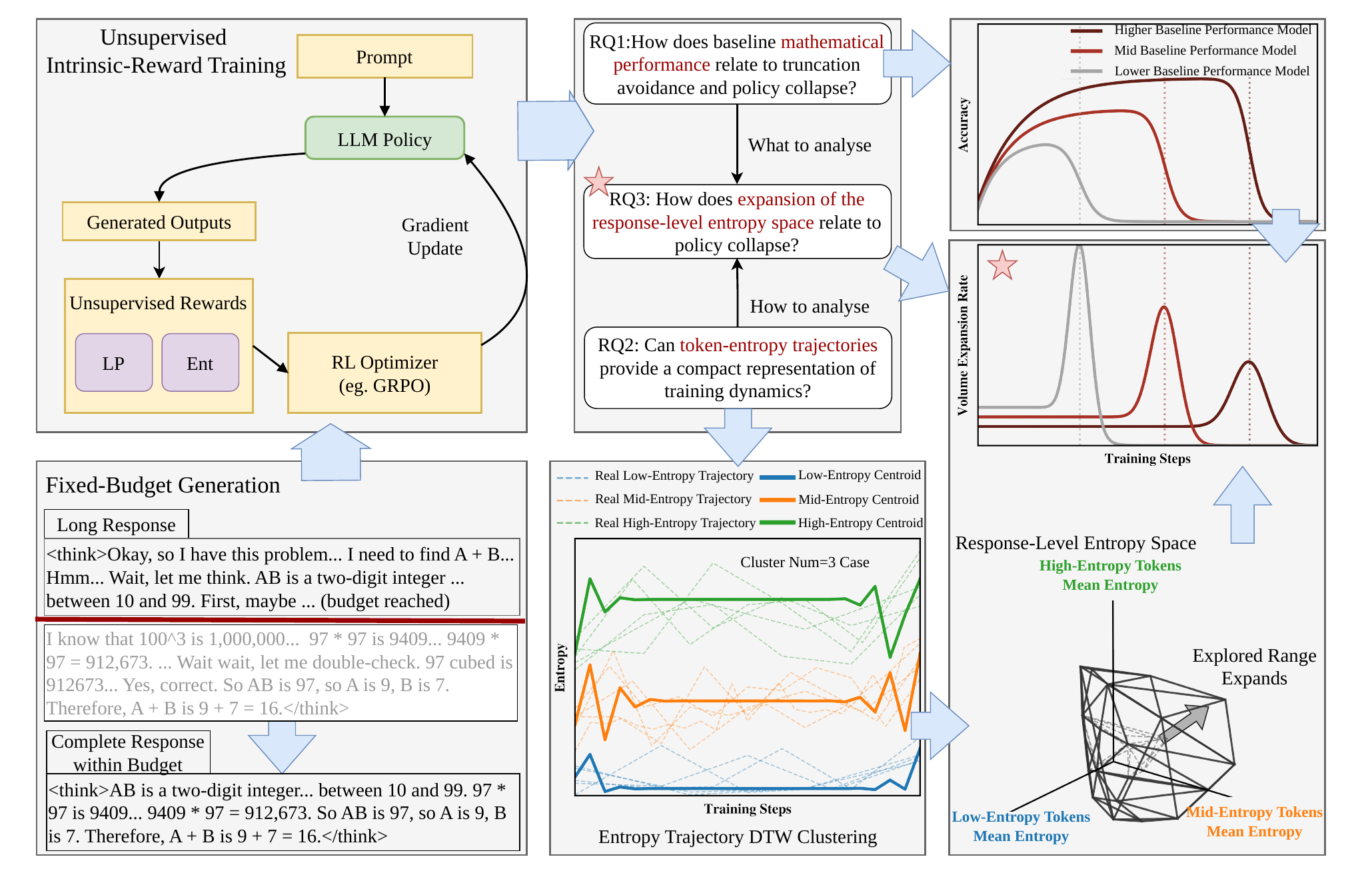}
\caption{
\RevNoLine{CR-teaser-caption}{\textbf{Research Framework.} \textbf{RQ1:} Model-dependent reasoning compression and policy collapse. \textbf{RQ2:} Low-dimensional representations of token-entropy trajectories. \textbf{RQ3:} Convex-hull expansion in the constructed response-level entropy space.}%
{}}
\label{fig:teaser}
\end{figure*}

\section{Introduction}

Long-form reasoning can improve mathematical performance, but it increases generation cost and may exceed deployment budgets. Reinforcement learning with verifiable rewards (RLVR) is widely used to train mathematical reasoning models, with Group Relative Policy Optimization (GRPO) commonly serving as the optimization backbone \citep{grpo}. Long reasoning traces can also cause ``overthinking'' \citep{sui2025stopoverthinkingsurveyefficient} and fail to finish within a fixed response budget.

A common deployment constraint is a fixed response budget, which bounds generation cost and latency. When a response reaches the generation limit without producing an EOS token, it is truncated and treated as unsuccessful in our evaluation. We therefore adopt an outcome-based view of reasoning compression: \textit{truncation avoidance}. We ask whether more responses complete within the available budget while task accuracy is maintained or improved. This operationalization focuses on the final compression outcome rather than identifying which specific tokens or reasoning components are shortened, reorganized, or omitted.

Existing methods control reasoning length through external prompts, shortened supervision traces, or teacher models 
\Rev{CR-cite-kimi-intro}{\citep{han-etal-2025-token, xia-etal-2025-tokenskip, kimiteam2025kimik15scalingreinforcement}}%
{}. Their length control is externally specified rather than learned directly from whether a response is truncated by a hard generation budget. Fixed-budget compression without external correctness or length supervision therefore remains underexplored. Recent successes of unsupervised RL in mathematical reasoning show that intrinsic signals such as entropy or self-consistency can improve reasoning behavior without ground-truth correctness labels \citep{zhao2026learning, rent, unreasonable, empo, ttrl}. This motivates our question: can intrinsic rewards also serve as compression objectives that help reasoning fit within a fixed response budget?

We study this question with the framework shown in Figure~\ref{fig:teaser}. We apply two intrinsic rewards to models with different baseline mathematical performance: LP directly penalizes response length, while Ent penalizes cumulative token entropy and therefore couples response length with token uncertainty. Both rewards admit degenerate optima, so we analyze the full training trajectory rather than the final training state. We additionally examine direct accuracy--length dynamics and controls for reward alignment and loss aggregation. Our analysis addresses three research questions.

\Needspace{2\baselineskip}
\textbf{
\Rev{LAYOUT-intro-rq1}{RQ1:}%
{} How does baseline mathematical performance relate to reasoning compression and policy collapse?}
We use EOS-verified truncation avoidance as the operational measure of fixed-budget compression and compare models trained with LP or Ent rewards using the same GRPO backbone. Across the evaluated configurations, training may enter a transient regime of maintained or improved validation accuracy before policy collapse. 
\Rev{S2-intro-rq1}{Across the evaluated models, higher baseline mathematical performance is generally associated with a longer regime and later collapse.}%
{}  This result establishes an empirical association between baseline performance and the duration of the pre-collapse regime.

\Needspace{3\baselineskip}
\textbf{RQ2: Can token-entropy trajectories provide a compact representation of training dynamics?}
We analyze the temporal evolution of token-level entropy with time-series clustering. The resulting trajectory groups are used to construct a response-level entropy space, where each response is represented by its average entropy within each group. This construction summarizes changes in output behavior during training. It does not assume that the groups form independent semantic states. The main geometric pattern is preserved under different numbers of clusters.

\Needspace{3\baselineskip}
\textbf{RQ3: How does expansion of the response-level entropy space relate to policy collapse?}
We measure the cumulative convex hull of the explored response-level entropy space and compute its expansion rate across training steps. Expansion-rate surges are associated with policy collapse. 
\Rev{S2-intro-rq3}{Models with higher baseline performance tend to show later and smaller surges.}%
{}  Together with their later collapse, these observations suggest that models with higher baseline performance explore a more restricted region of the response-level entropy space before collapse.

\Rev{LAYOUT-intro-summary}{In summary,}%
{} models trained with LP or Ent rewards can enter a transient fixed-budget compression regime before policy collapse. Appendix~\ref{sec:appendix_controls} reports direct accuracy--length evidence and control runs. The regime duration differs across models and is associated with baseline mathematical performance. The response-level entropy space summarizes these dynamics, and its expansion rate captures growth of the explored entropy region associated with collapse. 
\Rev{ADD-intro-link}{Together, these analyses connect fixed-budget task outcomes with changes in token uncertainty throughout training.}%
{}

\section{Related Work}

\paragraph{Reasoning Compression via Supervised Methods and Prompting}
Current approaches control reasoning length through prompt engineering, supervised fine-tuning (SFT), and distillation. 
\Rev{CITE-once-supervised}{Prompting methods instruct LLMs to reduce verbosity or satisfy a token budget. SFT methods train models on shortened reasoning traces.}%
{} Distillation transfers reasoning from teacher models to shorter text sequences \citep{chen-etal-2025-skip} or continuous latent representations \citep{shen-etal-2025-codi}. These methods depend on externally specified budgets, labeled shortened traces, or teacher models rather than learning directly from EOS-verified truncation under a hard response budget. Our work instead studies outcome-based reasoning compression under a fixed response budget during unsupervised RL training.

\paragraph{Length Constraints in RLVR}
RLVR combines correctness rewards with penalties on long responses. Existing methods use 
\Rev{CR-cite-kimi-related}{length penalties}%
{}  or budget constraints based on the Information Bottleneck principle \citep{massoli2026reasoningcompressionunifyingbudget}. These methods retain a verifiable correctness signal during training. Our setting removes this signal from the policy update and uses length- or entropy-based rewards as the training objective. This design allows us to study fixed-budget compression, operationalized through truncation avoidance and task accuracy, without correctness supervision in the reward.

\paragraph{Unsupervised Reinforcement Learning}
Unsupervised reinforcement learning uses rewards derived from model outputs rather than ground-truth answers. 
\Rev{CITE-once-unsupervised}{Existing methods use entropy minimization or self-consistency to modify reasoning behavior.}%
{} Our work uses intrinsic rewards as probes of fixed-budget compression behavior. We analyze token-entropy trajectories and construct a response-level entropy space from the resulting trajectory groups. The convex-hull expansion rate in this space summarizes expansion of the explored entropy region associated with policy collapse.

\section{Training Methods}

We use GRPO as the shared optimizer for all reward conditions. GRPO is representative of mathematical RLVR, supports grouped rollouts
without a separate critic, and allows us to compare reward conditions under a fixed optimization backbone.

\subsection{Policy Optimization}
\label{subsec:grpo}

For a given mathematical problem query $q$, we sample a group of $G$ response token trajectories $\{o_1, o_2, \dots, o_G\}$ from the old policy $\pi_{\theta_{old}}$. For each trajectory $o_i$, we compute a scalar reward $r_i$, as defined in Section~\ref{sec:reward_formulation}.

The advantage $A_i$ for each rollout is calculated by standardizing the rewards within the sampled group:
\begin{equation}
    A_i = \frac{r_i - \mu_{\mathbf{r}}}{\sigma_{\mathbf{r}}}
\end{equation}
where $\mu_{\mathbf{r}}$ and $\sigma_{\mathbf{r}}$ are the mean and standard deviation of the group rewards $\mathbf{r} = \{r_1, \dots, r_G\}$, respectively. The policy model $\pi_{\theta}$ is then optimized by maximizing the GRPO objective:
\begin{equation}
\begin{aligned}
\mathcal{J}(\theta)&=\mathbb{E}\Bigg[\frac{1}{G}\sum_{i=1}^G\\
&\min\big(\rho_i(\theta)A_i,
\RevNoLine{FIX-clip}{\operatorname{clip}(\rho_i(\theta),1-\epsilon,1+\epsilon)}%
A_i\big)\Bigg]\end{aligned}
\end{equation}
where $\rho_i(\theta) = \pi_{\theta}(o_i \mid q)/\pi_{\theta_{old}}(o_i \mid q)$ is the importance sampling ratio and $\epsilon$ is the clipping hyperparameter. We exclude the KL-divergence term $\mathbb{D}_{KL}$ in all settings. This choice allows us to observe the adaptation and collapse dynamics without reference-policy regularization.

\subsection{Rewards}
\label{sec:reward_formulation}

We evaluate two intrinsic rewards: LP and Ent. LP applies a uniform penalty to response length. Ent applies a cumulative penalty based on token entropy and response length.

During RL training, for a generated response of length $T$, let $p_t(v)$ denote the predicted probability of token $v$ at generation step $t$.

\paragraph{LP.}
LP is defined as
\begin{equation}
R_{\mathrm{LP}} = -T/T_{\mathrm{max}}
\end{equation}
where $T_{\mathrm{max}}$ denotes the maximum response length.

\paragraph{Ent.}
Ent is defined using Shannon entropy:
\begin{equation}
R_{\mathrm{Ent}}=-\sum_{t=1}^{T}\mathcal{H}(p_t)=\sum_{t=1}^{T}\sum_{v\in\mathcal{V}}p_t(v)\log p_t(v)
\end{equation}
\Rev{FIX-vocabulary}{Here, $\mathcal{V}$ is the token vocabulary.}%
{} Because $\mathcal{H}(p_t)$ is non-negative, $R_{\mathrm{Ent}}$ decreases as cumulative token entropy increases. It therefore penalizes long responses and high-entropy token distributions.

\begin{table}[!t]
\centering
\small
\begin{tabular}{lr}
\toprule
\textbf{Model} & \textbf{AIME26 Pass@1} \\
\midrule
Llama-3.1-8B-Instruct & 3.3 \\
DeepSeek-R1-Distill-Llama-8B & 33.3 \\
\RevNoLine{S2-baseline-row}{Qwen3-1.7B}%
 & {\revisioncolor 35.8} \\
Qwen3-8B & \textbf{63.3} \\

\bottomrule
\end{tabular}
\caption{\textbf{Baseline Mathematical Performance.} Pass@1 accuracy on AIME26 under a 32K response budget.}
\label{tab:f_reasoning}
\end{table}

\section{Experimental Settings}
\label{sec:experiments}
The experiments study fixed-budget compression behavior across models with different baseline mathematical performance. We treat each reward and optimizer configuration as a complete training condition and analyze the resulting compression outcome, measured by truncation avoidance and task accuracy, together with the associated entropy-space dynamics. The main experiments use the loss-aggregation setting reported in Appendix~\ref{sec:appendix_exp}
\Rev{FIX-settings-period}{.}%
{} Appendix~\ref{sec:appendix_controls} reports direct accuracy--length evidence and controls for reward alignment and loss aggregation.

\paragraph{Models.}
We evaluate four models: Llama-3.1-8B-Instruct~\cite{grattafiori2024llama3herdmodels}, Qwen3-1.7B, Qwen3-8B~\cite{yang2025qwen3technicalreport}, and DeepSeek-R1-Distill-Llama-8B~\cite{deepseekai2025deepseekr1incentivizingreasoningcapability}. Table~\ref{tab:f_reasoning} reports the baseline mathematical performance 
\Rev{S2-baseline-description}{of all four models on AIME26}%
{}
\Rev{CR-cite-aime26-1}{\footnote{\label{fn:aime26}\url{https://huggingface.co/datasets/math-ai/aime26}}}%
{} under a 32K response budget. The model set provides complementary comparisons. Llama-3.1-8B-Instruct and DeepSeek-R1-Distill-Llama-8B form a same-scale Llama-family contrast with substantially different baseline mathematical performance. Qwen3-1.7B and Qwen3-8B provide a within-family comparison, while Qwen3-8B also serves as a cross-family high-performance reference at the 8B scale. These comparisons allow us to examine whether the training dynamics track observed baseline mathematical performance across and within model families. All models use their native \texttt{think} mode when available or prompted Chain-of-Thought reasoning otherwise.

\paragraph{Training and Evaluation Setup.}
The models are trained on DAPO-Math-17K~\cite{dapo} under a 4K response budget. LP and Ent use no correctness labels in their rewards or policy updates. We evaluate validation performance on AIME24
\Rev{CR-cite-aime24}{\footnote{\label{fn:aime24}\url{https://huggingface.co/datasets/math-ai/aime24}}}%
{} every five training steps. Because AIME24 contains a limited number of problems, we generate eight independent responses per problem and report mean Pass@1 accuracy.

AIME24 accuracy is used to monitor training and select the peak checkpoint for cross-benchmark evaluation. The policy optimization is therefore unsupervised with respect to correctness, while evaluation and checkpoint selection use labeled answers. For reporting, the method label \textit{GRPO} denotes the supervised correctness-reward reference. LP and Ent use the same GRPO optimizer but are labeled by their unsupervised rewards. We evaluate the selected checkpoints on MATH500~\cite{hendrycks2021measuringmathematicalproblemsolving}, Minerva Math~\cite{minerva}, OlympiadBench~\cite{olympiadbench}, AIME24, AMC23,
\Rev{DATA-amc-source}{\footnote{\label{fn:amc23}\url{https://maa.org/math-competitions/}}}%
{} and AIME26.
\begin{table*}[!t]
\centering
\resizebox{\textwidth}{!}{
\setlength{\tabcolsep}{2pt}
\begin{tabular}{l l ccc ccc ccc ccc ccc ccc ccc}
\toprule
\multirow{2}{*}{\textbf{Model}} & \multirow{2}{*}{\textbf{Method}} & 
\multicolumn{3}{c}{\textbf{MATH500}} & 
\multicolumn{3}{c}{\textbf{OlympiadBench}} & 
\multicolumn{3}{c}{\textbf{Minerva Math}} & 
\multicolumn{3}{c}{\textbf{AIME24}} & 
\multicolumn{3}{c}{\textbf{AMC23}} & 
\multicolumn{3}{c}{\textbf{AIME26}} & 
\multicolumn{3}{c}{\textbf{AVG}}\\
\cmidrule(lr){3-5} \cmidrule(lr){6-8} \cmidrule(lr){9-11} \cmidrule(lr){12-14} \cmidrule(lr){15-17} \cmidrule(lr){18-20} \cmidrule(lr){21-23}
 & & P@1 & P@5 & P@10 & P@1 & P@5 & P@10 & P@1 & P@5 & P@10 & P@1 & P@5 & P@10 & P@1 & P@5 & P@10 & P@1 & P@5 & P@10 & P@1 & P@5 & P@10\\
\midrule

\multirow{4}{*}{\shortstack[l]{\textbf{Qwen3-1.7B}}}
& Base & 70.2 & 81.4 & 84.9 & 30.1 & 41.6 & 46.9 & 30.6 & 42.0 & 46.0 & 6.0 & 14.2 & 19.1 & 47.2 & 61.4 & 68.0 & 7.1 & 15.5 & 20.0 & 42.7 & 54.0 & 58.5 \\
& GRPO & 78.8 & 89.5 & 91.8 & 39.5 & 52.0 & 56.0 & \textbf{35.4} & \textbf{49.2} & \textbf{53.9} & 17.9 & 31.0 & 37.8 & 56.6 & 76.0 & 82.1 & 12.7 & 26.0 & 31.1 & 51.0 & 63.3 & 67.0 \\
\cmidrule{2-23}
& LP & \textbf{\underline{80.0}} & \textbf{\underline{90.4}} & \textbf{\underline{92.4}} & \textbf{\underline{42.6}} & \textbf{\underline{55.3}} & \textbf{\underline{59.1}} & \underline{34.4} & 47.9 & 52.4 & 20.0 & 34.2 & 39.3 & 57.8 & 79.7 & 86.8 & \textbf{\underline{14.4}} & \textbf{\underline{28.0}} & \textbf{\underline{33.6}} & \textbf{\underline{52.7}} & 65.0 & 68.5 \\
& Ent & 79.8 & \textbf{\underline{90.4}} & \textbf{\underline{92.4}} & 41.4 & 54.9 & 58.8 & \underline{34.4} & \underline{48.3} & \underline{52.9} & \textbf{\underline{22.5}} & \textbf{\underline{42.6}} & \textbf{\underline{49.3}} & \textbf{\underline{61.3}} & \textbf{\underline{81.8}} & \textbf{\underline{88.1}} & 14.0 & 27.9 & 31.6 & 52.2 & \textbf{\underline{65.1}} & \textbf{\underline{68.7}} \\
\midrule

\multirow{4}{*}{\shortstack[l]{\textbf{Qwen3-8B}}}
& Base & 67.9 & 80.5 & 84.4 & 28.4 & 41.2 & 46.7 & 29.8 & 39.9 & 44.8 & 7.7 & 19.9 & 23.1 & 41.7 & 64.1 & 73.6 & 9.2 & 16.6 & 22.1 & 41.0 & 53.4 & 58.3 \\
& GRPO & 88.4 & 93.8 & 95.0 & 51.7 & 60.1 & 63.4 & 46.0 & 55.7 & 58.3 & 36.2 & 52.7 & \textbf{60.4} & 76.2 & 87.9 & 89.1 & 36.2 & 48.8 & 50.0 & 62.6 & 70.6 & 73.1 \\
\cmidrule{2-23}
& LP & \textbf{\underline{88.7}} & \textbf{\underline{94.2}} & \textbf{\underline{95.2}} & \textbf{\underline{52.3}} & \textbf{\underline{61.1}} & \textbf{\underline{64.2}} & \textbf{\underline{46.1}} & \textbf{\underline{56.2}} & \textbf{\underline{59.1}} & \textbf{\underline{39.2}} & \textbf{\underline{55.8}} & \underline{59.1} & \textbf{\underline{78.8}} & \textbf{\underline{93.0}} & \textbf{\underline{94.9}} & \textbf{\underline{36.7}} & \textbf{\underline{51.1}} & 54.1 & \textbf{\underline{63.1}} & \textbf{\underline{71.5}} & \textbf{\underline{73.8}} \\
& Ent & 86.6 & 93.1 & 94.5 & 49.9 & 58.2 & 61.0 & 43.9 & 54.4 & 58.2 & 31.7 & 48.0 & 53.5 & 74.1 & 89.4 & 91.5 & 34.2 & 49.8 & \textbf{\underline{54.6}} & 60.7 & 69.3 & 71.9 \\
\midrule

\multirow{4}{*}{\shortstack[l]{\textbf{DeepSeek-R1-Distill}\\\textbf{Llama-8B}}}
& Base & 69.4 & 87.1 & 90.8 & 35.7 & 52.9 & 58.8 & 22.7 & 40.2 & 47.8 & 16.5 & 30.3 & 37.9 & 50.8 & 80.6 & 86.9 & 14.4 & 27.4 & 32.7 & 43.9 & 61.5 & 67.0 \\
& GRPO & \textbf{72.7} & \textbf{90.2} & \textbf{93.7} & \textbf{39.3} & \textbf{56.9} & \textbf{62.4} & \textbf{24.4} & \textbf{42.7} & \textbf{49.9} & \textbf{24.4} & \textbf{44.9} & \textbf{54.9} & \textbf{59.4} & \textbf{86.9} & \textbf{91.1} & \textbf{22.5} & \textbf{40.0} & \textbf{44.6} & \textbf{47.4} & \textbf{65.4} & \textbf{70.6} \\
\cmidrule{2-23}
& LP & \underline{70.5} & \underline{88.5} & \underline{92.5} & \underline{36.6} & 53.9 & 59.5 & \underline{22.2} & \underline{40.7} & \underline{48.3} & \underline{21.7} & \underline{43.1} & \underline{52.8} & 50.8 & 81.4 & 87.0 & \underline{17.5} & \underline{33.8} & \underline{39.4} & \underline{44.7} & \underline{62.9} & \underline{68.4} \\
& Ent & 70.0 & 88.1 & 91.9 & 36.4 & \underline{54.6} & \underline{60.6} & 21.8 & 39.9 & 47.2 & 18.1 & 37.9 & 47.5 & \underline{53.4} & \underline{83.8} & \underline{88.5} & 17.3 & 31.2 & 36.1 & 44.4 & 62.8 & \underline{68.4} \\
\midrule

\multirow{4}{*}{\shortstack[l]{\textbf{Llama-3.1-8B-}\\\textbf{Instruct}}}
& Base & 45.5 & \textbf{69.4} & \textbf{77.2} & 13.2 & 29.8 & 38.2 & 21.4 & 39.8 & 47.1 & 5.2 & 14.3 & 20.8 & 22.7 & \textbf{51.6} & \textbf{66.4} & 0.4 & 2.1 & 4.2 & 24.9 & 44.1 & 52.1 \\
& GRPO & \textbf{46.3} & 69.0 & 76.4 & \textbf{14.1} & \textbf{31.0} & 38.8 & 25.7 & \textbf{45.5} & \textbf{52.3} & 4.6 & 13.2 & 18.7 & 23.6 & 46.8 & 56.1 & \textbf{1.2} & 3.1 & 3.3 & \textbf{26.4} & \textbf{45.4} & 52.7 \\
\cmidrule(lr){2-23} 
& LP  & \multicolumn{21}{c}{\textit{Collapse} ($\times$)}  \\
& Ent & \underline{45.8} & \textbf{\underline{69.4}} & \underline{77.0} & \underline{13.9} & \underline{30.6} & \textbf{\underline{38.9}} & \textbf{\underline{25.8}} & \underline{43.7} & \underline{50.1} & \textbf{\underline{6.5}} & \textbf{\underline{16.7}} & \textbf{\underline{23.3}} & \textbf{\underline{24.1}} & \underline{46.9} & \underline{56.9} & \underline{1.0} & \textbf{\underline{4.9}} & \textbf{\underline{9.2}} & \underline{26.2} & \underline{45.1} & \textbf{\underline{52.8}} \\
\bottomrule
\end{tabular}
}
\caption{\textbf{Peak Results on Mathematical Reasoning Benchmarks under a 4K Response Budget.} \textit{GRPO} denotes the supervised correctness-reward reference. LP and Ent use the same GRPO optimizer and are labeled by their unsupervised rewards. \textbf{Bold} denotes the overall best performance, and \underline{underlined} indicates the best result among unsupervised methods within each model. \textit{P@k} means Pass@k.}
\label{tab:main_results_comprehensive}
\end{table*}

\begin{figure}[!t]
    \centering
\zsavepos{CR-figure2-labels-start}\label{CR-figure2-labels-page}

    \begin{subfigure}{\linewidth}
        \centering
        \RevisedGraphic[width=\linewidth]{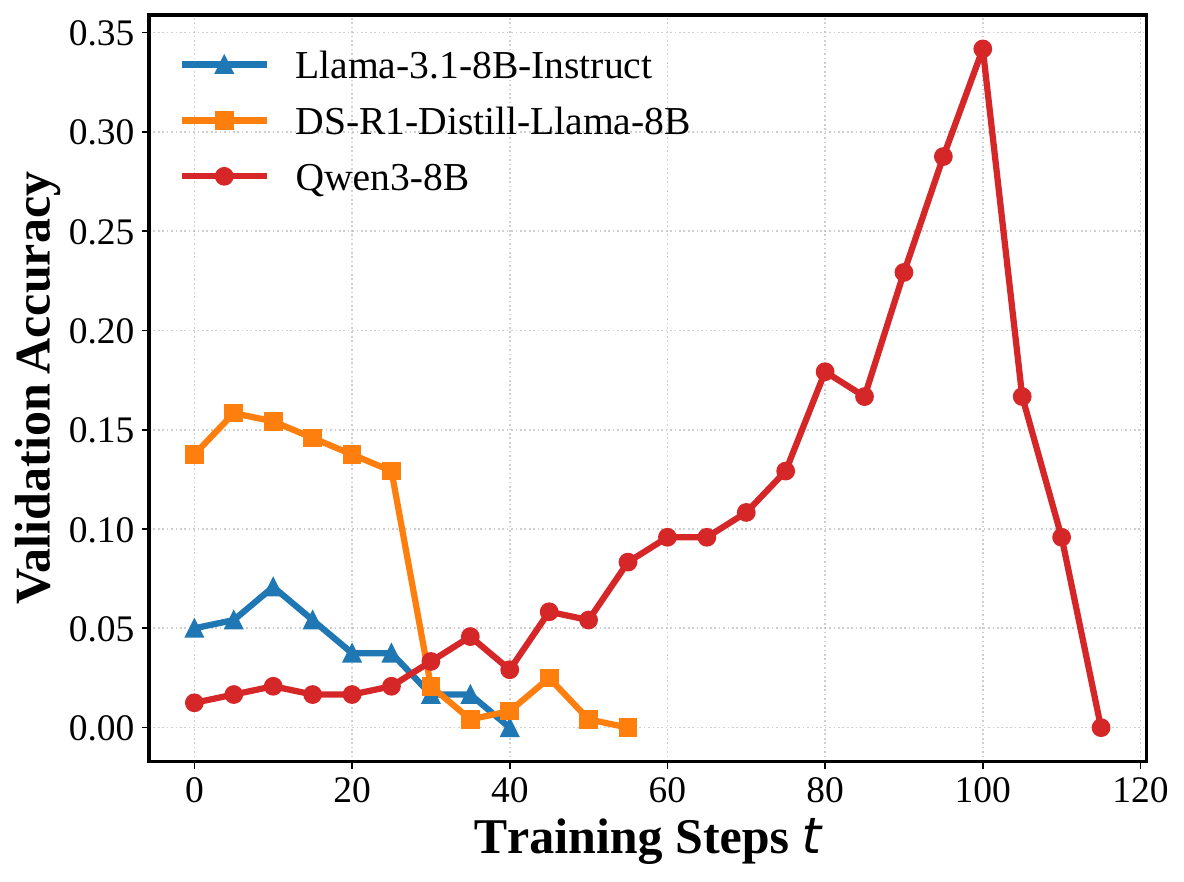}{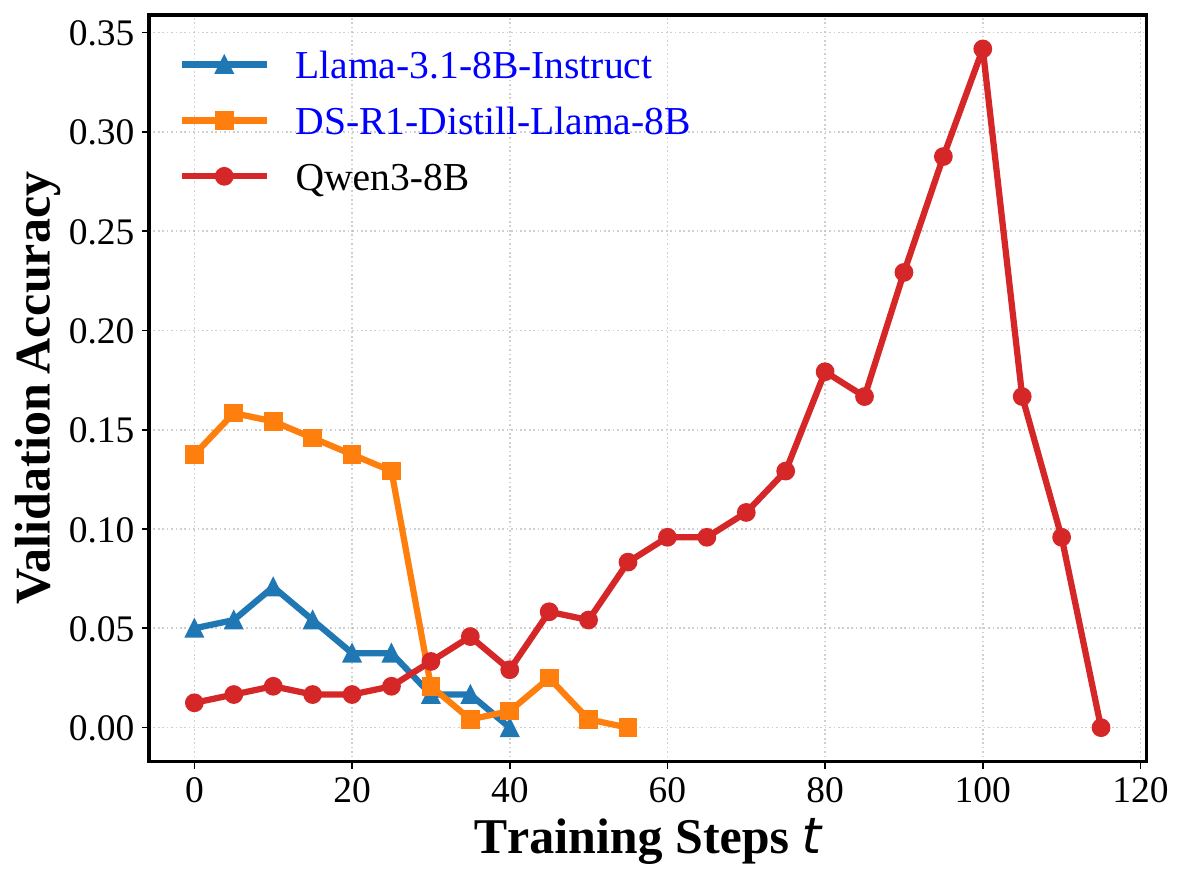}
        \caption{Method: Ent}
        \label{fig:acc_collapse_ent}
    \end{subfigure}
    
    \vspace{0.5cm} 
    
    \begin{subfigure}{\linewidth}
        \centering
        \RevisedGraphic[width=\linewidth]{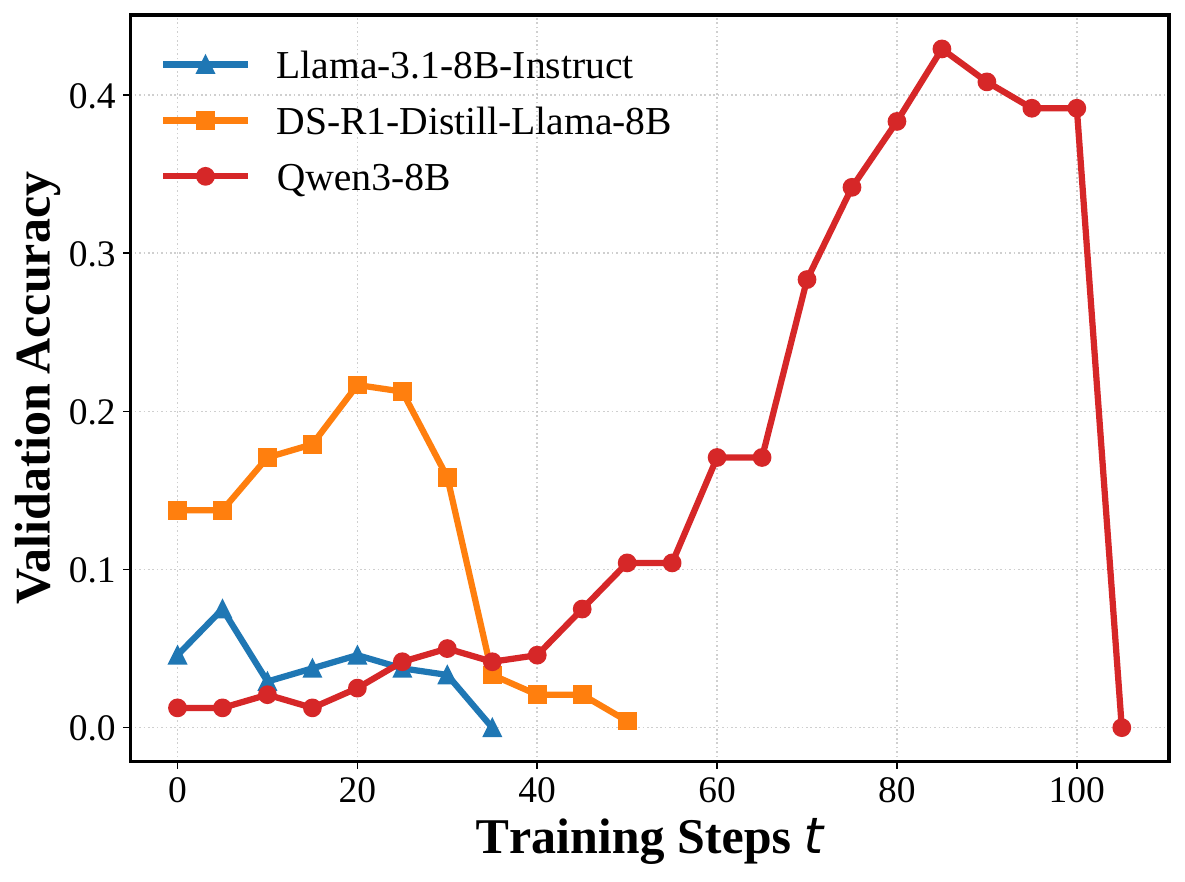}{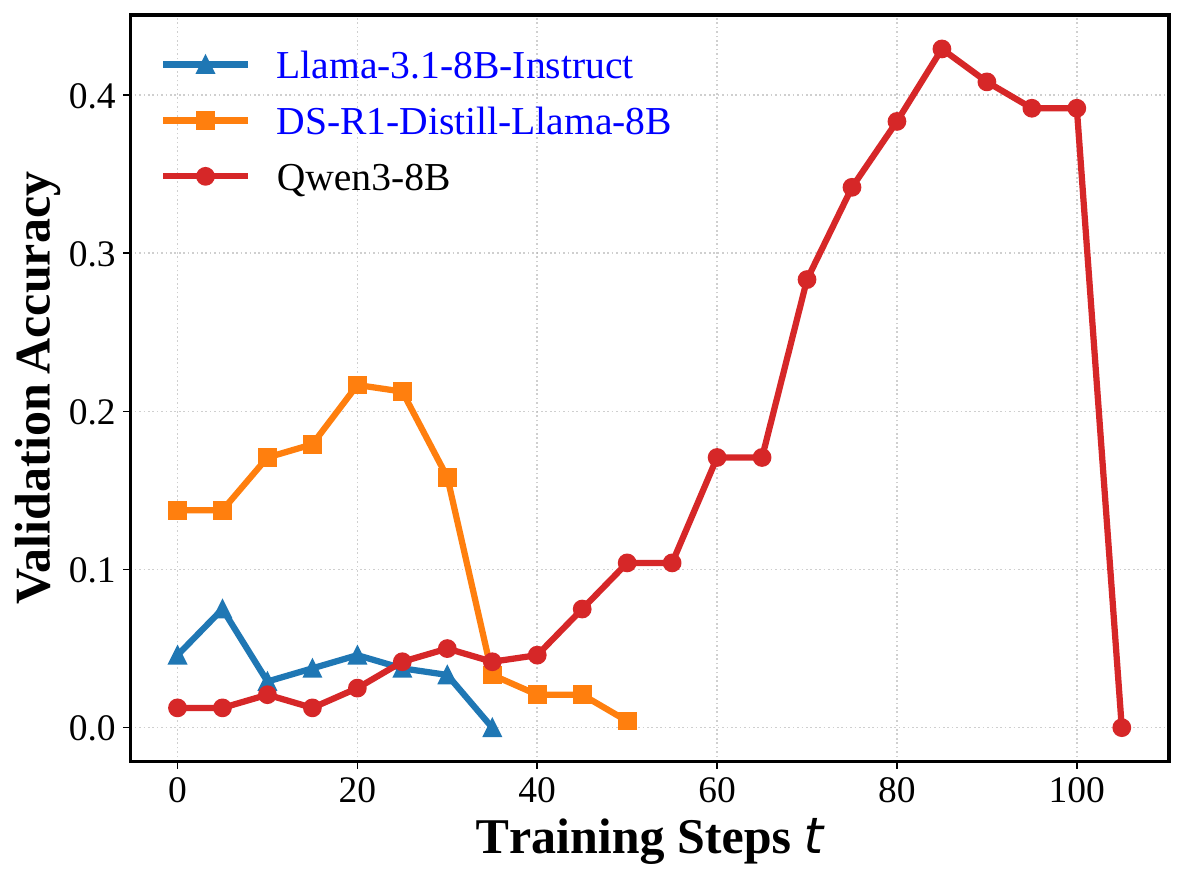}
        \caption{Method: LP}
        \label{fig:acc_collapse_lp}
    \end{subfigure}

    \zsavepos{CR-figure2-labels-end}
\caption{\textbf{Validation accuracy curves under a 4K response budget.} The vertical axis reports AIME24 validation accuracy as expected Pass@1, averaged across eight independent generations per problem.}    \label{fig:accuracy_collapse_comparison}
\end{figure}

\section{RQ1: Baseline Performance and Reasoning Compression}
\label{sec:results}

We first examine validation accuracy across training steps under Ent and LP, as shown in Figure~\ref{fig:accuracy_collapse_comparison}. Most settings show a transient accuracy peak before policy collapse, while the weakest model may collapse before a meaningful improvement. The timing of this transition differs across models. Qwen3-8B maintains its performance for more training steps, DeepSeek-R1-Distill-Llama-8B reaches an earlier and lower peak, and Llama-3.1-8B-Instruct collapses early under LP and shows limited improvement under Ent. 
\Rev{S2-rq1-scope}{Across the evaluated models, higher baseline mathematical performance is generally associated with a longer pre-collapse regime and later policy collapse.}%
{} 
\Rev{S2-qwen-timing}{The Qwen3 trajectories in Appendix~\ref{sec:appendix_budget_dynamics} have similar peak times (within one five-step evaluation interval), while Qwen3-8B reaches higher peak accuracy under both rewards.}%
{} 

\begin{figure}[!t]
    \centering
    \includegraphics[width=\linewidth]{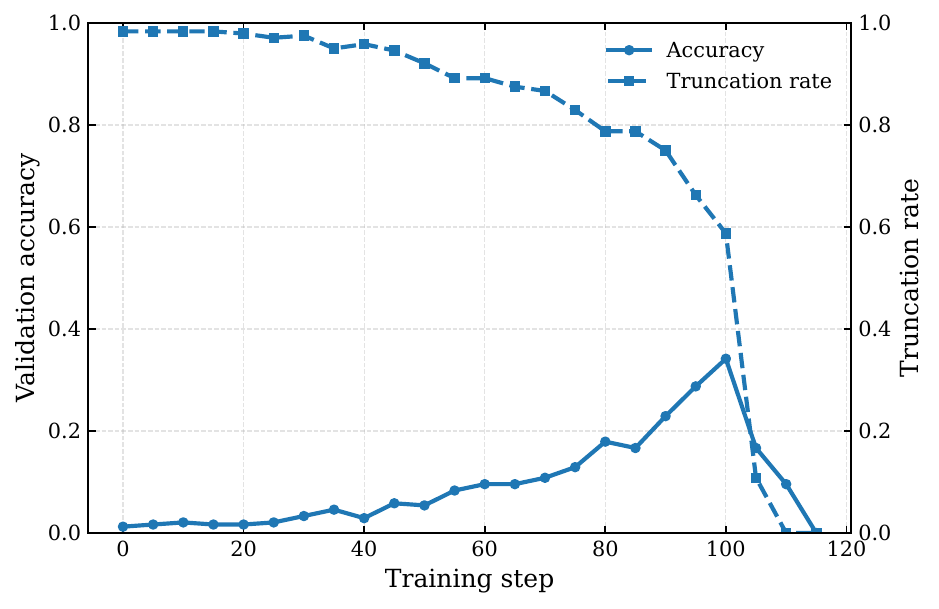}
    \caption{\textbf{Accuracy and truncation rate during Ent training for Qwen3-8B.} The truncation rate is the fraction of validation responses that reach the 4K generation limit without EOS tokens. Validation accuracy increases as the truncation rate decreases, followed by policy collapse under continued optimization.}
    \label{fig:qwen8b_ent_truncation}
\end{figure}

Under our outcome-based definition, a decrease in EOS-verified truncation while accuracy is maintained or improved indicates that reasoning is more successfully accommodated within the fixed response budget. Figure~\ref{fig:qwen8b_ent_truncation} reports validation accuracy and truncation rate for Qwen3-8B under Ent. The truncation rate is the fraction of responses that reach the 4K generation limit without producing an EOS token. It decreases as validation accuracy increases, indicating that more responses terminate before reaching the generation limit. Accuracy subsequently collapses under continued optimization. The corresponding Ent and LP trajectories are provided in Appendix~\ref{sec:appendix_budget_dynamics}.

We select the checkpoint with the highest AIME24 validation accuracy from each training trajectory and evaluate it on additional mathematical benchmarks. As shown in Table~\ref{tab:main_results_comprehensive}, the selected checkpoints preserve the model-dependent pattern observed during validation. 
\Rev{S2-benchmark-scope}{Qwen3-8B obtains the strongest cross-benchmark results under LP and Ent}%
{}, while the gains are smaller for DeepSeek-R1-Distill-Llama-8B and limited for Llama-3.1-8B-Instruct. These results show that the transient states selected on AIME24 can generalize across mathematical benchmarks. Appendix~\ref{sec:appendix_controls} provides direct and control evidence with global token-mean and shuffled-LP runs.

\subsection{
\RevNoLine{ADD-length-head}{Direct Evidence of Response Shortening}%
}
\label{sec:ADD-length-head}
\Rev{ADD-length}{All eight LP/Ent runs produce shorter responses at their AIME24 accuracy peak than at the initial checkpoint (Table~\ref{tab:sac_all_lengths}). Mean length falls by 0.7\% to 24.9\%. For Qwen3-8B LP, length falls from 4,091.9 to 3,072.7 tokens as accuracy rises from 1.25\% to 42.9\% and truncation falls from 98.3\% to 47.1\%. These joint changes directly support the temporary fixed-budget compression outcome.}%

\begin{table*}[!t]
\RevFloatStart{S3-main-table}
\centering
\small
\setlength{\tabcolsep}{3pt}
\begin{tabular}{llrrr}
\toprule
\textbf{Model} & \textbf{Reward} & \textbf{Acc. (\%)} & \textbf{Mean tokens} & \textbf{Trunc. (\%)} \\
\midrule
Qwen3-1.7B & LP & 1.67 $\to$ 25.00 & 4,076.8 $\to$ 3,208.3 & 97.1 $\to$ 56.2 \\
Qwen3-1.7B & Ent & 1.67 $\to$ 21.67 & 4,076.8 $\to$ 3,511.4 & 97.1 $\to$ 68.8 \\
Qwen3-8B & LP & 1.25 $\to$ 42.92 & 4,091.9 $\to$ 3,072.7 & 98.3 $\to$ 47.1 \\
Qwen3-8B & Ent & 1.25 $\to$ 34.17 & 4,091.9 $\to$ 3,343.6 & 98.3 $\to$ 58.8 \\
DS-R1-Distill-Llama-8B & LP & 13.75 $\to$ 21.67 & 3,930.7 $\to$ 3,775.2 & 84.2 $\to$ 74.2 \\
DS-R1-Distill-Llama-8B & Ent & 13.75 $\to$ 15.83 & 3,930.7 $\to$ 3,903.3 & 84.2 $\to$ 81.7 \\
Llama-3.1-8B-Instruct & LP & 4.58 $\to$ 7.50 & 2,062.7 $\to$ 1,829.5 & 32.5 $\to$ 24.2 \\
Llama-3.1-8B-Instruct & Ent & 5.00 $\to$ 7.08 & 2,176.6 $\to$ 1,910.2 & 37.5 $\to$ 30.0 \\
\bottomrule
\end{tabular}
\caption{\textbf{AIME24 accuracy, response length and truncation from the initial checkpoint to peak accuracy under a 4K budget.} Accuracy is mean Pass@1 over eight responses per problem. Peak is the earliest checkpoint with maximum AIME24 monitoring accuracy in each run. Mean length includes all responses. Truncation is the fraction reaching the budget without EOS. DS-R1-Distill-Llama-8B abbreviates DeepSeek-R1-Distill-Llama-8B.}
\label{tab:sac_all_lengths}
\RevFloatEnd{S3-main-table}
\end{table*}


\section{RQ2: Entropy Dynamics as a Low-Dimensional Proxy}
\label{sec:ent_analysis}

Due to space limitations, we focus on Qwen3-8B with Ent in this section. Results for other models and reward formulations are provided in 
\Rev{CR-cluster-results-link}{Appendix~\ref{sec:appendix_cok_clustering}}%
{}.

We analyze the temporal evolution of token entropy during training. We use 
\Rev{FIX-dtw-main}{dynamic time warping (DTW)~\citep{tavenard2020tslearn}}%
{} to measure the distance between entropy trajectories and Time-Series K-means to cluster them. The algorithm is described in 
\Rev{CR-cluster-method-links}{Appendices~\ref{sec:appendix_clustering_algo} and~\ref{sec:ADD-cluster-head}}%
{}.

\subsection{Sensitivity to the Number of Clusters}

We evaluate the number of clusters $K$ using DTW Inertia and the Silhouette Score. DTW Inertia measures the sum of squared distances between each trajectory and its cluster centroid:
\begin{equation}
\text{Inertia}=\sum_{k=1}^{K}\sum_{x_i\in C_k}\text{DTW}(x_i,\mu_k)^2
\end{equation}
where $C_k$ is the $k$-th cluster and $\mu_k$ is its centroid. The Silhouette Score measures cluster separation and cohesion:
\begin{equation}
S=\frac{1}{N}\sum_{i=1}^{N}\frac{b_i-a_i}{\max(a_i,b_i)}
\end{equation}
where $a_i$ is the mean intra-cluster distance and $b_i$ is the mean distance to the nearest cluster.

\begin{figure}[!t]
    \centering
    \includegraphics[width=\linewidth]{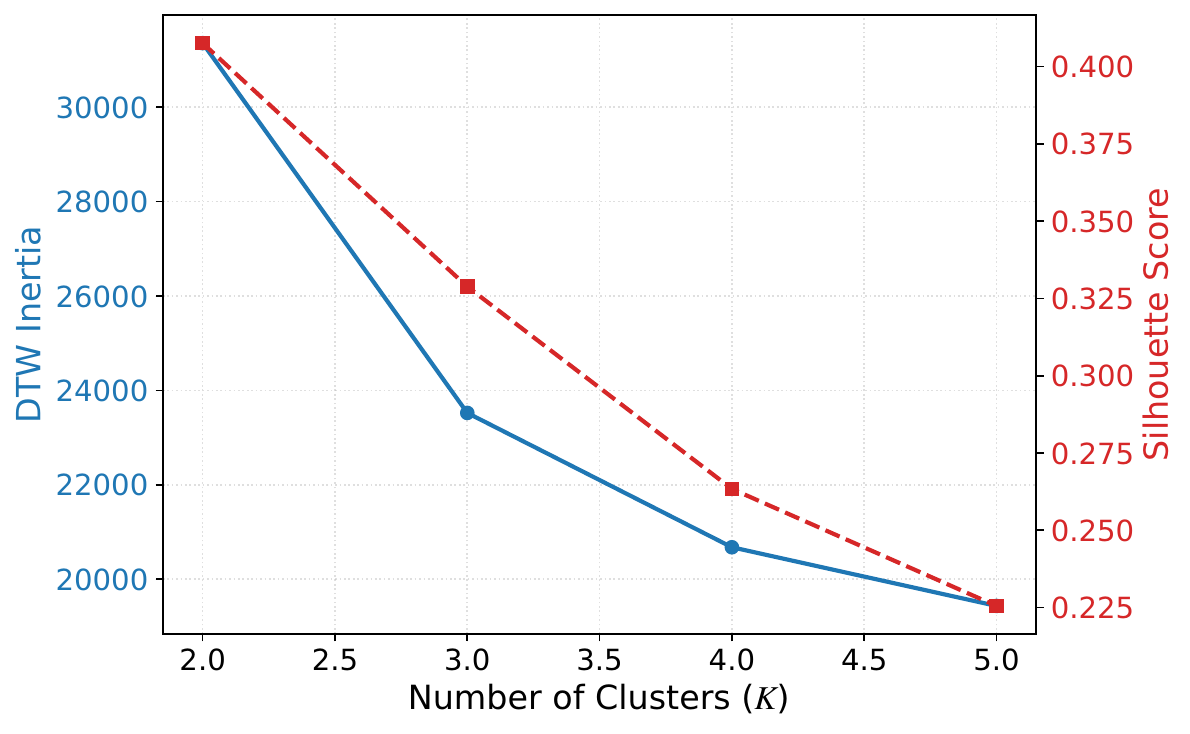}
    \caption{\textbf{Evaluation of $K$ for Qwen3-8B with Ent.} $K=2$ achieves the highest Silhouette Score, while the DTW Inertia curve shows an elbow at $K=3$.}
    \label{fig:k_choice_q8b}
\end{figure}

Figure~\ref{fig:k_choice_q8b} shows that the two criteria favor different resolutions. $K=2$ provides stronger cluster separation, while $K=3$ provides a finer partition near the DTW Inertia elbow. We therefore retain both values as complementary representations and evaluate the geometric analysis under both settings. Results across all model and reward settings are provided in Appendix~\ref{sec:appendix_cok}. 

\begin{figure}[!t]
    \centering

    \begin{subfigure}[t]{\linewidth}
        \centering
        \includegraphics[width=\linewidth]{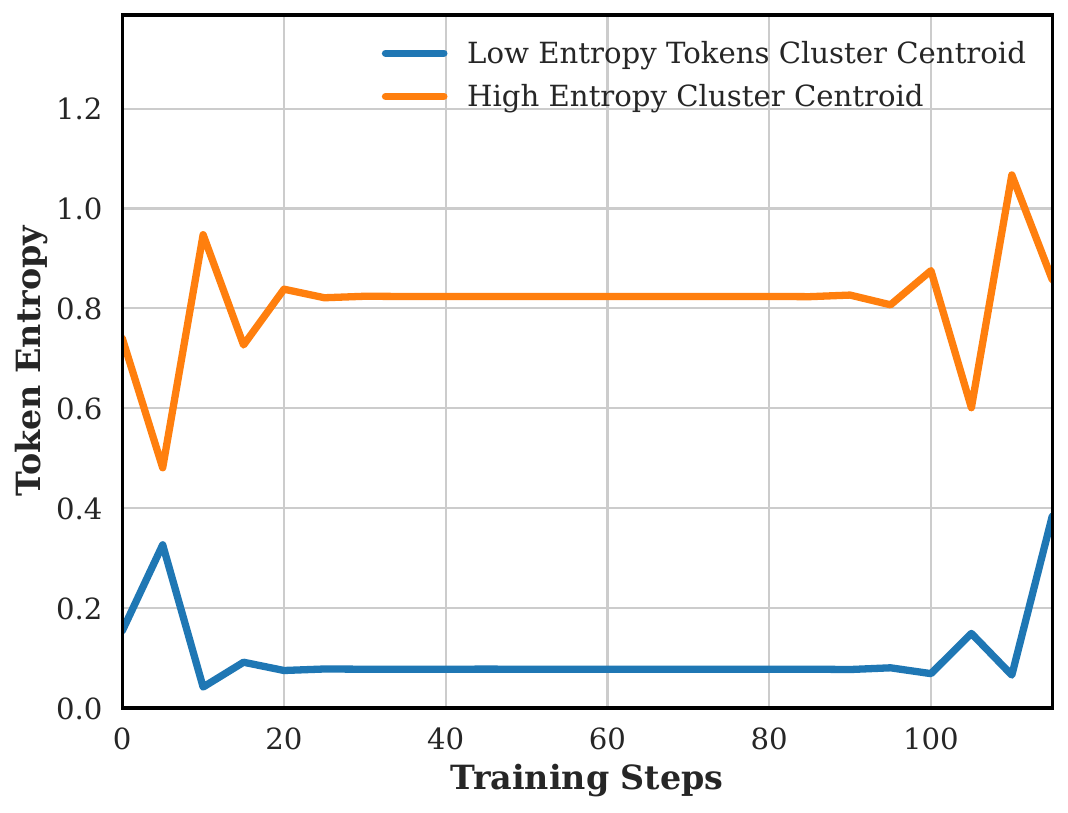}
        \caption{$K=2$}
        \label{fig:token_entropy_centroids_k2}
    \end{subfigure}

    \vspace{0.4em}

    \begin{subfigure}[t]{\linewidth}
        \centering
        \includegraphics[width=\linewidth]{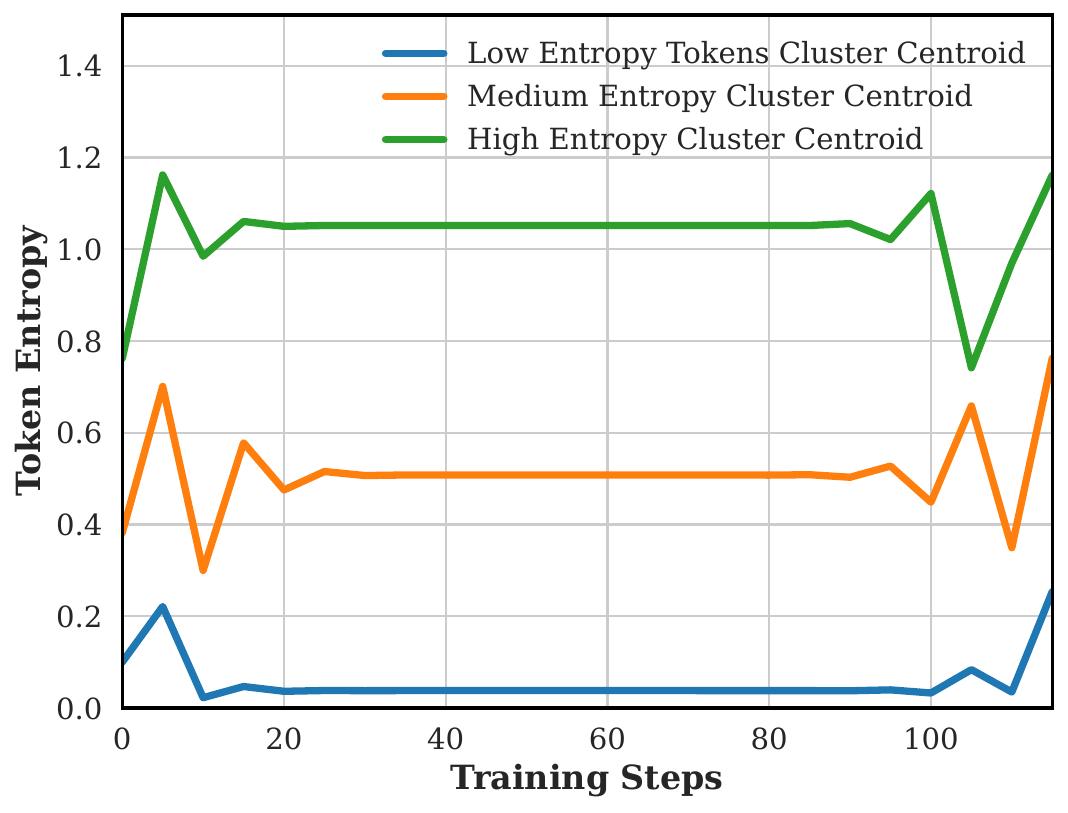}
        \caption{$K=3$}
        \label{fig:token_entropy_centroids_k3}
    \end{subfigure}

    \caption{\textbf{Token-entropy trajectory clusters for Qwen3-8B with Ent.} $K=2$ gives a coarse partition of the trajectories, while $K=3$ separates them into low-, mid-, and high-entropy groups. Both settings provide low-dimensional summaries of the training dynamics.}
    \label{fig:token_entropy_centroids_short}
\end{figure}

Time-Series K-means groups token-entropy trajectories according to their temporal patterns. Figure~\ref{fig:token_entropy_centroids_short} shows the results for $K=2$ and $K=3$. The exact partition changes with $K$, but both settings separate trajectories with different entropy levels and temporal trends. We use these groups to construct the response-level entropy space in Section~\ref{sec:volume_expansion}.

Appendix~\ref{sec:appendix_cok_semantics} provides a qualitative interpretation of the $K=3$ entropy groups based on representative tokens.

\begin{figure*}[!t]
    \centering
\zsavepos{CR-figure6-labels-start}\label{CR-figure6-labels-page}

    \RevisedGraphic[width=\textwidth]{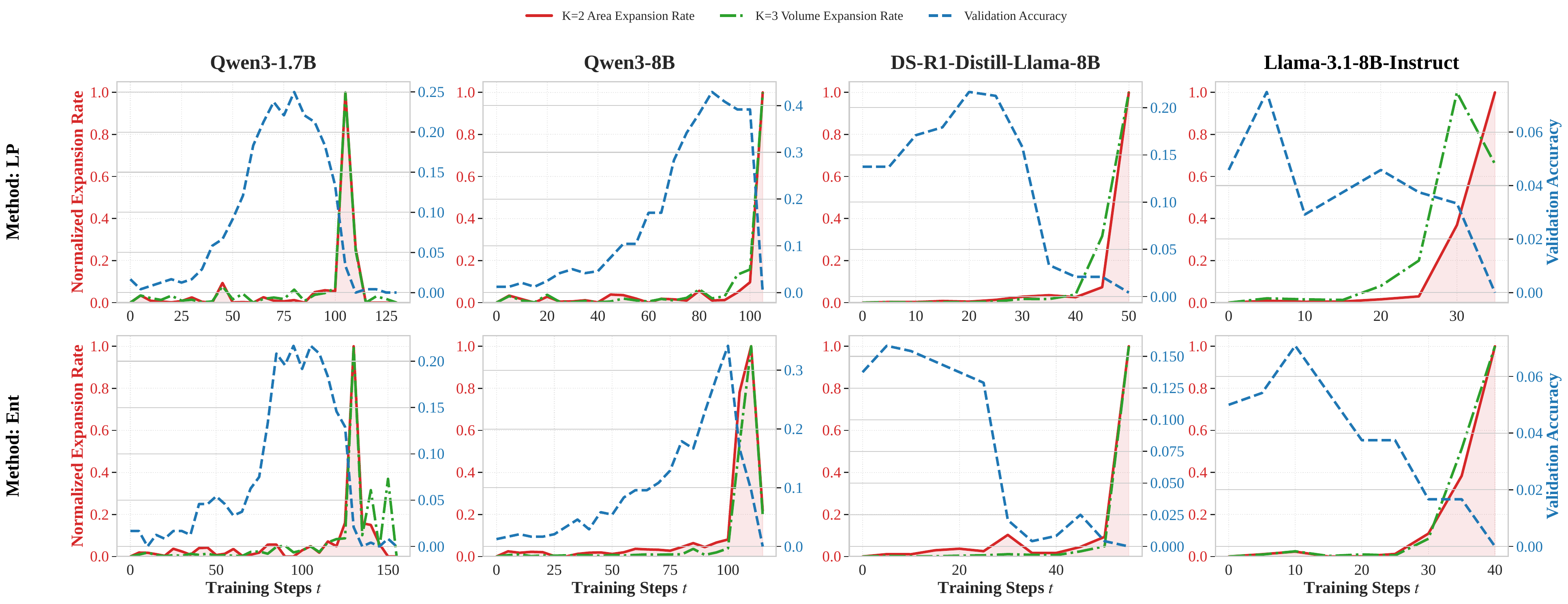}{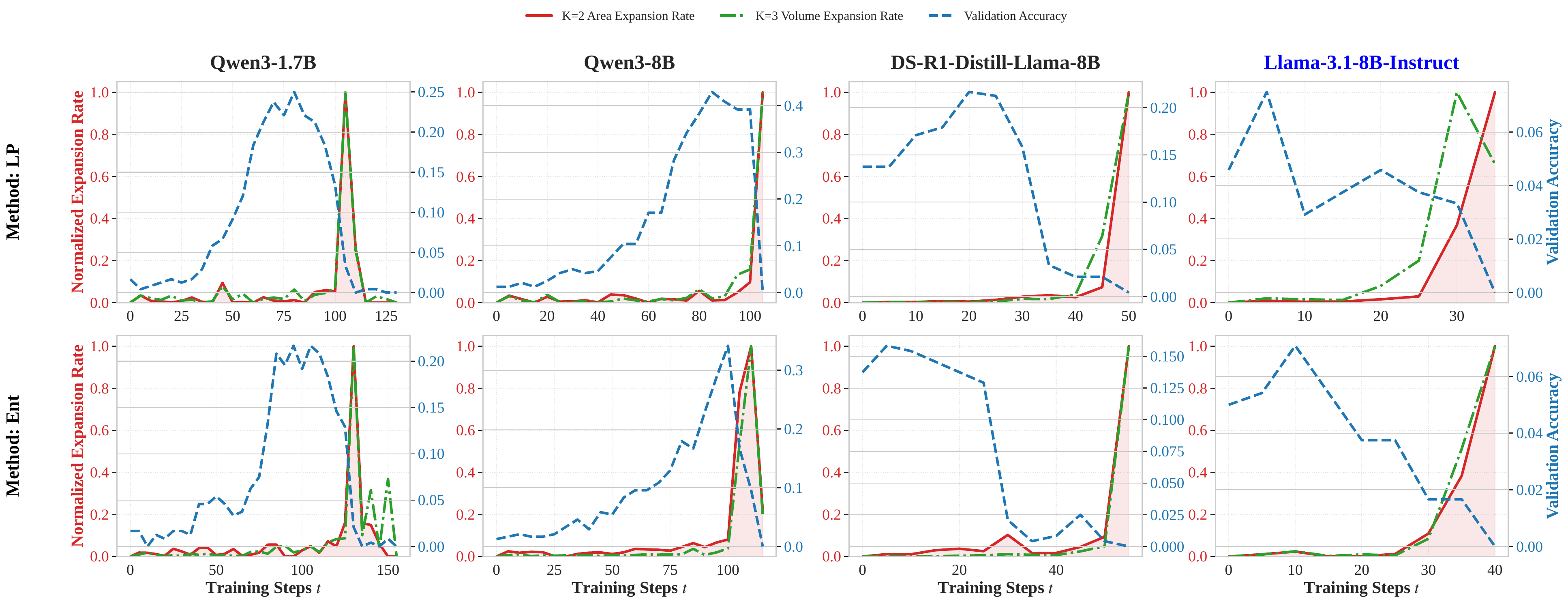}
    \zsavepos{CR-figure6-labels-end}
\caption{\textbf{Validation accuracy and convex-hull expansion under $K=2$ and $K=3$.} $R_2(t)$ measures area expansion and $R_3(t)$ measures volume expansion. Expansion-rate surges occur around accuracy collapse under both representations.}
    \label{fig:k2_k3_accuracy}
\end{figure*}

\begin{figure*}[!t]
    \centering
\zsavepos{CR-figure7-labels-start}\label{CR-figure7-labels-page}

    \begin{subfigure}[t]{0.99\textwidth}
        \centering
        \RevisedGraphic[width=\linewidth]{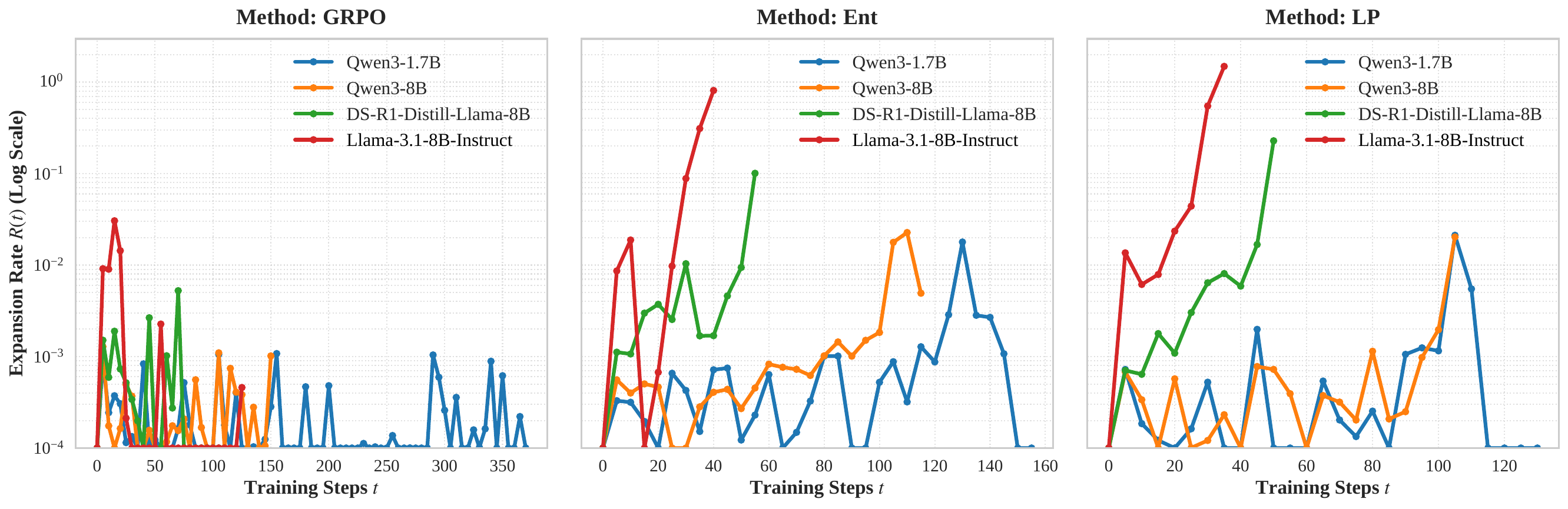}{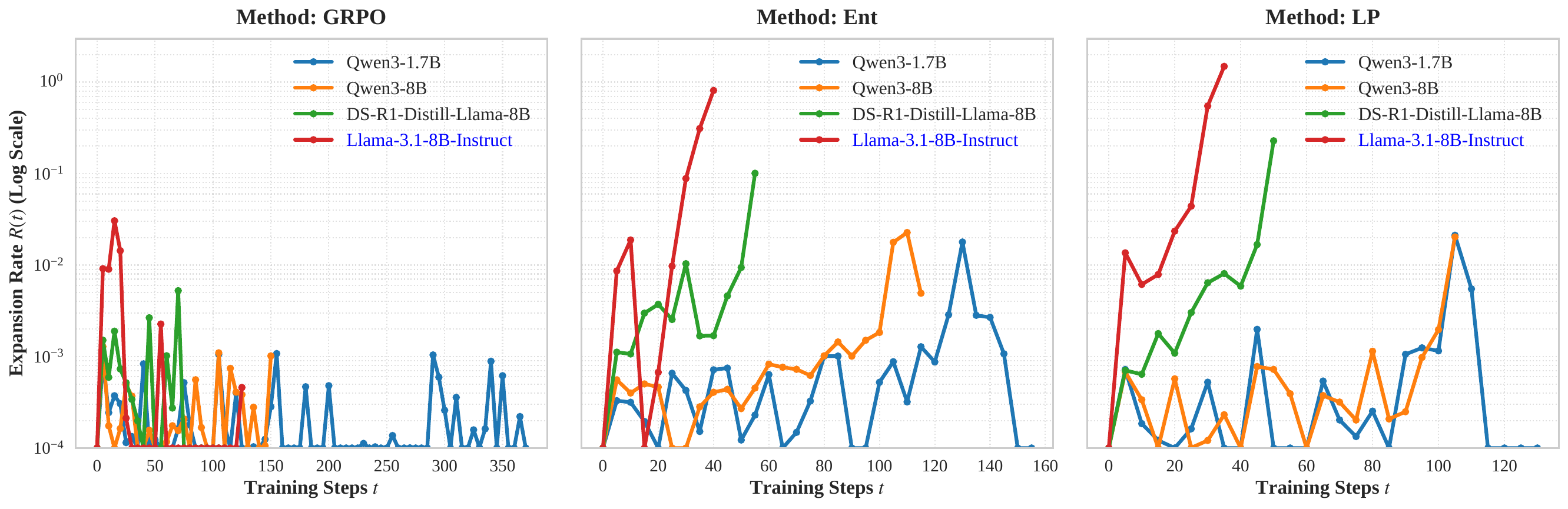}
        \caption{$K=2$}
        \label{fig:k2_cross_model}
    \end{subfigure}
    \vfill
    \begin{subfigure}[t]{0.99\textwidth}
        \centering
        \RevisedGraphic[width=\linewidth]{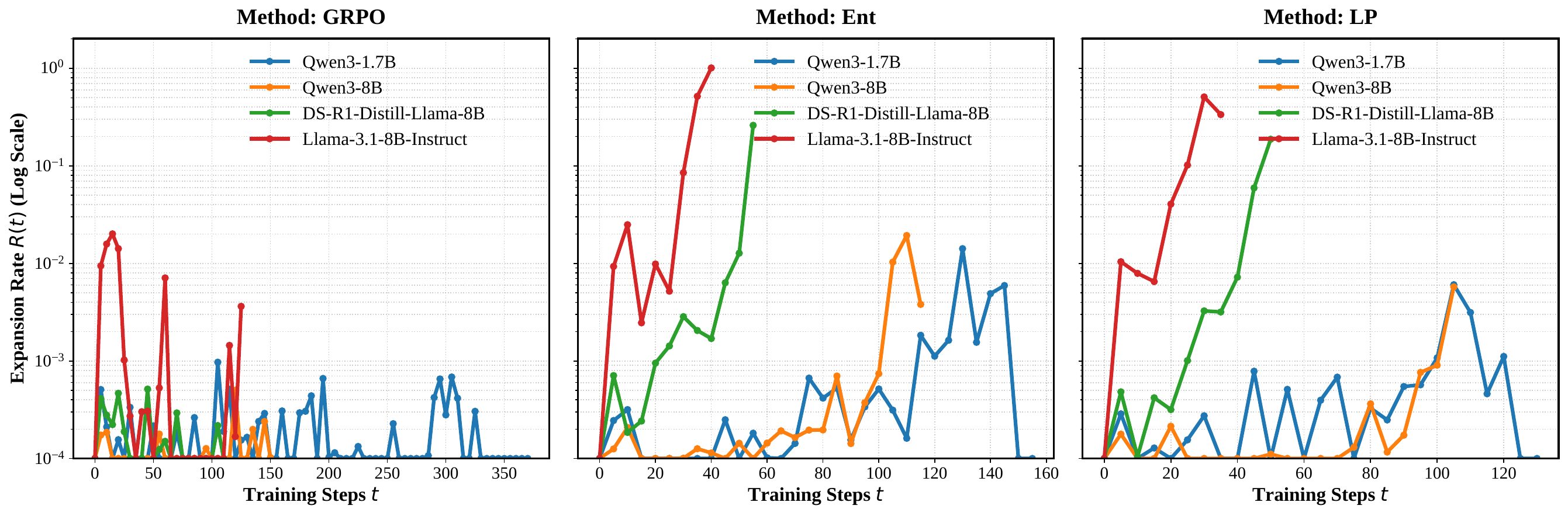}{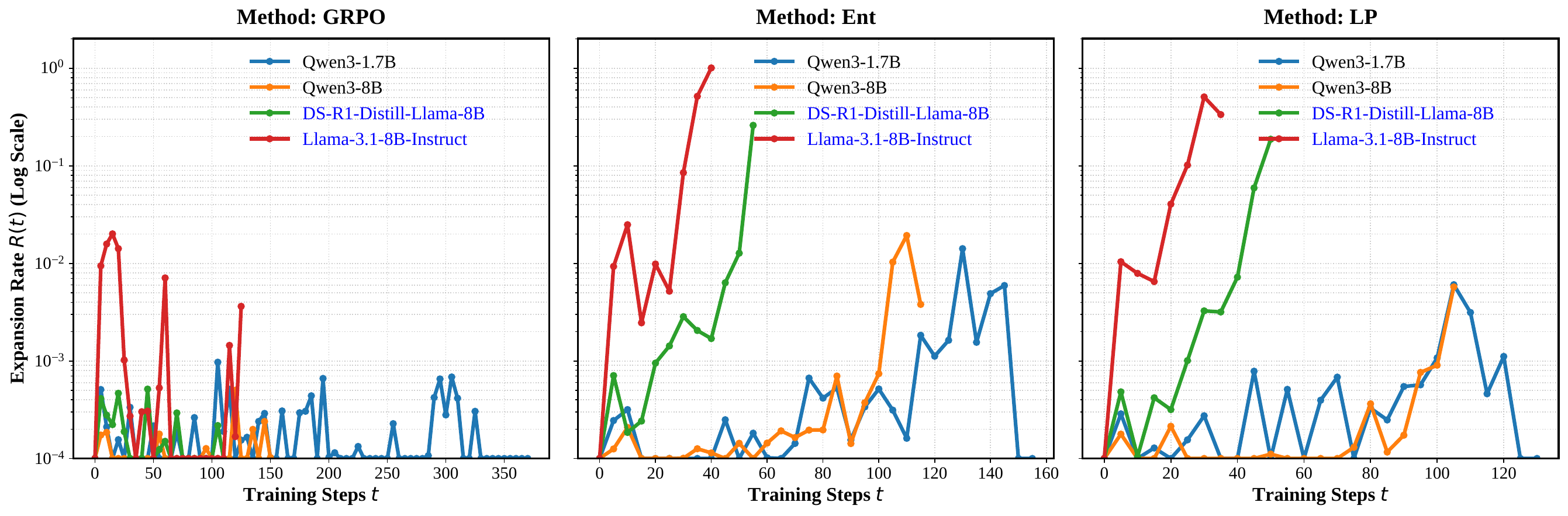}
        \caption{$K=3$}
        \label{fig:k3_cross_model}
    \end{subfigure}
    \zsavepos{CR-figure7-labels-end}
\caption{\textbf{Cross-model comparison of the convex-hull expansion rate.} 
\RevNoLine{S2-crossmodel-caption}{Models with higher baseline mathematical performance tend to show later and lower-magnitude surges under LP and Ent.}%
{}  The same cross-model pattern is observed with $K=2$ and $K=3$.}
    \label{fig:cross_model_expansion}
\end{figure*}

\section{RQ3: Entropy-Space Expansion and Policy Collapse}
\label{sec:volume_expansion}

\Rev{S2-rq3-scope}{RQ1 shows an overall association between higher baseline mathematical performance, a longer transient regime and later collapse across the evaluated models.}%
{}  We examine whether this difference is reflected in the range of entropy patterns explored during training.

\subsection{Constructing a Response-Level Entropy Space}

We evaluate each model every five training steps on AIME24. For a clustering result with $K$ groups, let $C_k$ denote the $k$-th entropy-trajectory group. For a response $r$, let $T_k(r)$ be the set of its tokens assigned to $C_k$. We define the response-level entropy coordinate
\begin{equation}
E_k(r)=\frac{1}{|T_k(r)|}\sum_{x\in T_k(r)}H(x)
\end{equation}
and set $E_k(r)=0$ when $T_k(r)$ is empty. Each response is represented as
\begin{equation}
\mathbf{p}_r=(E_1(r),\dots,E_K(r)).
\end{equation}

Let $S_t$ be the cumulative set of response coordinates from step 0 to step $t$:
\begin{equation}
S_t=\bigcup_{\tau=0}^{t}\{\mathbf{p}_r\mid r\text{ is generated at step }\tau\}.
\end{equation}
We measure the full cumulative range reached by the policy using the convex hull of $S_t$:
\begin{equation}
V_K(t)=\operatorname{Vol}_K\!\left(\operatorname{ConvHull}(S_t)\right),
\label{eq:vol}
\end{equation}
where $\operatorname{Vol}_2$ denotes area and $\operatorname{Vol}_3$ denotes volume. Each evaluation step contributes the same number of responses. Because the target quantity is the complete range reached during training, any response that extends the hull is counted as newly explored behavior rather than removed as an outlier.

The expansion rate is
\begin{equation}
R_K(t)=\frac{V_K(t)-V_K(t-\Delta t)}{\Delta t},
\label{eq:vr}
\end{equation}
where $\Delta t=5$ and $R_K(0)=0$. This construction produces a response-level entropy space.

\subsection{
\RevNoLine{ADD-geometry-head}{Interpreting Entropy-Space Expansion}%
}
\label{sec:ADD-geometry-head}
\Rev{ADD-means}{Cluster-wise means separate entropy intensity from token counts. Only tokens assigned to a group contribute to its coordinate, and an absent group receives zero. Responses of different lengths can therefore share a coordinate when their assigned tokens have the same mean entropy. A sum would combine entropy intensity with token counts. The coordinates summarize uncertainty among assigned tokens together with group presence.}%

\Rev{ADD-hull}{The cumulative hull is the smallest convex set enclosing all observed response coordinates, including earlier checkpoints. It preserves the extrema of every linear combination of entropy coordinates, while its area or volume summarizes their joint extent. For example, four corners of a unit square and four points split evenly between opposite corners have identical coordinate means, ranges and variances, but hull areas of one and zero. The hull distinguishes these joint arrangements.}%

\Rev{ADD-responsechecks}{Coverage filtering retains prompt--token pairs observed at the specified fraction of checkpoints, and the complete-coordinate check retains responses with every group present. Fitted group assignments remain fixed. Each filtered or resampled expansion peak is compared with its original full-sample peak.}%

\Rev{S1-stability-prose}{Expansion-peak timing remains stable under response filtering and resampling. Across 16 conditions (four models, two rewards and $K=2,3$), prompt--token coverage thresholds of 0\%, 10\%, 25\% and 50\% retain the original peak. Requiring every group to be present places all peaks within ten steps. Removing the most coordinate-extreme 5\% or 10\% of responses preserves the exact peak in 14/16 conditions and places all peaks within five steps. With 50\% response subsampling over 100 repetitions, at least 99\% of peaks remain within ten steps in each condition. These checks support timing stability across the observed response set.}%

\Rev{S1-joint-checks}{The surge also involves many responses: a median 47.1\% lie outside the previous cumulative hull at peak expansion. For Qwen3-8B Ent with $K=3$, hull growth between steps 100 and 105 persists after matching response-length bins and group-presence patterns. This supports joint entropy variation among comparable responses.}%

\subsection{Expansion Near Policy Collapse}

Figure~\ref{fig:k2_k3_accuracy} compares validation accuracy with the expansion rate under $K=2$ and $K=3$. Sharp increases in $R_K(t)$ occur around policy collapse under both settings. 
\Rev{S2-retrospective-timing}{At the five-step evaluation resolution, the dominant surge coincides with the sharp accuracy drop in most runs and follows it in the remaining runs. The timing supports a retrospective description of collapse.}%
{}  The same expansion--collapse association is observed under both clustering resolutions. After the generated responses become degenerate, they add few new points to the cumulative explored region and the expansion rate decreases.

\subsection{Cross-Model Comparison}

Figure~\ref{fig:cross_model_expansion} compares the expansion rate across models under both $K=2$ and $K=3$. 
\Rev{S2-crossmodel-scope}{Under LP and Ent, models with higher baseline mathematical performance tend to show later and lower-magnitude expansion-rate surges.}%
{}  These models also collapse later, as reported in Section~\ref{sec:results}. Under the supervised GRPO reference, the expansion rate remains lower across the evaluated models. The same pattern appears under both clustering resolutions.

\section{Discussion}
\label{sec:discussion}

\paragraph{Entropy-trajectory structure.}
The cluster centroids are approximately horizontal during much of the intermediate training regime. 
\Rev{FIX-discussion-level}{This pattern suggests stable entropy levels within the trajectory groups.}%
{} The stable centroid levels summarize the central entropy of each trajectory group, whereas hull expansion records newly reached combinations of response-level entropy coordinates. Thus, the explored region can expand even when the centroid levels remain approximately stable.

\paragraph{Possible policy constraints.}
Because the cumulative hull counts every newly reached region, a low expansion rate indicates that generated responses remain within the previously explored range, whereas a surge indicates that the policy has reached new combinations of token-entropy statistics. The expansion rate may therefore reflect how tightly a policy constrains changes in token uncertainty. Because the entropy space is constructed from output distributions, this interpretation concerns observable policy behavior rather than a direct measurement of hidden states.

\paragraph{
\RevNoLine{ADD-scope-head}{Scope of the Findings.}%
}
\label{sec:ADD-scope-head}
\Rev{ADD-scope}{Compression means completing responses within a fixed token budget while preserving accuracy. Each reward and loss-aggregation configuration defines a training condition. Cross-model comparisons describe empirical associations with baseline performance. Geometry provides a retrospective description in the entropy coordinates fitted for each run.}%


\section{Conclusion}
\label{sec:conclusion}

We study compressing mathematical reasoning under a fixed response budget, operationalized through EOS-verified truncation avoidance and 
\Rev{S3-conclusion-length}{supported by direct measurements of response length}%
{}. With LP or Ent rewards optimized by GRPO, models can enter a transient compression regime before collapse, 
\Rev{S2-conclusion-scope}{and higher baseline performance is generally associated with a longer regime.}%
{}  Across both $K=2$ and $K=3$, convex-hull expansion in the response-level entropy space increases around collapse, linking model-dependent collapse resistance to the range of entropy patterns explored during training.

\FloatBarrier


\clearpage
\section*{Limitations}

\paragraph{Model Coverage.}
We evaluate four models from the Qwen and Llama families, all at or below 8B parameters. The model set includes within-family and same-scale comparisons, but larger models and additional families are needed to test the generality of the observed performance ordering.

\paragraph{Scope of the Compression Measure.}
We operationalize fixed-budget compression through EOS-verified truncation behavior, task accuracy, and an additional direct accuracy--length trajectory. These outcome-based measures do not identify which tokens or reasoning components are shortened, reorganized, omitted, or preserved.

\paragraph{Unsupervised Training and Supervised Evaluation.}
LP and Ent use no correctness labels in the reward or policy update. Labeled AIME24 accuracy is used for training monitoring and checkpoint selection. The policy optimization is unsupervised with respect to correctness, while evaluation and model selection are supervised. 

\paragraph{Expansion Metric.}
Convex-hull expansion intentionally emphasizes responses that reach previously unexplored regions because the target quantity is the cumulative explored range. It should therefore be interpreted as a range-based measure rather than a density-weighted summary of typical responses. Its estimated magnitude may still depend on validation-set and sampling coverage. 
\Rev{ADD-limits}{Numerical comparisons across models require calibration of the fitted entropy coordinates.}%
{}

\section*{
\RevNoLine{CR-ack-head}{Acknowledgments}%
}
\label{sec:CR-ack-head}
\Rev{CR-ack}{This work was supported by JST BOOST Program, Japan Grant Number JPMJBY25D2. We used ABCI 3.0 provided by AIST and AIST Solutions with support from ``ABCI 3.0 Development Acceleration Use.''}%


\bibliography{custom}

\appendix

\begin{table*}[hbt!]
\centering
\small
\begin{tabular}{lc}
\toprule
\textbf{Hyperparameter} & \textbf{Value} \\
\midrule
\multicolumn{2}{c}{\textit{Optimization}} \\
\midrule
Actor Learning Rate & $1 \times 10^{-6}$ \\
Learning Rate Warmup Steps & 10 \\
Weight Decay & 0.1 \\
Mini-batch Size & 24 \\
Prompt Batch Size & 8 \\
\midrule
\multicolumn{2}{c}{\textit{Reinforcement Learning}} \\
\midrule
Responses per Prompt ($G$) & 16 \\
PPO Clip Ratio ($\epsilon$) & 0.2 \\
Main Loss Aggregation Mode & seq-mean-token-mean \\
Control Loss Aggregation Mode & global token-mean \\
Max Prompt Length & 1024 \\
\midrule
\multicolumn{2}{c}{\textit{Generation (Rollout \& Validation)}} \\
\midrule
Training Temperature & 0.6 \\
Training Top-$p$ & 1.0 \\
Training Top-$k$ & -1 \\
Validation Top-$p$ & 0.95 \\
Validation Top-$k$ & 20 \\
\bottomrule
\end{tabular}
\caption{Hyperparameters for RL training.}
\label{tab:rl_hyperparameters}
\end{table*}

\section{
\RevNoLine{FIX-appendix-settings-title}{Experimental Settings}%
{}}
\label{sec:appendix_exp}

The training hyperparameters are reported in Table~\ref{tab:rl_hyperparameters}. The validation generation settings follow the 
\Rev{CITE-once-qwen-report}{Qwen3 Technical Report}%
{}.

\subsection{
\RevNoLine{ADD-evalstats-head}{Evaluation Statistics and Table Conventions}%
}
\label{sec:ADD-evalstats-head}
\Rev{ADD-evalstats}{Training monitoring uses eight responses per AIME24 problem, while the selected checkpoint is evaluated with 16 responses per problem on the six mathematical benchmarks. For $c$ correct answers among $n=16$ samples, Pass@$k$ is estimated by $1-\binom{n-c}{k}/\binom{n}{k}$, with a zero numerator when $n-c<k$. Scores are averaged over problems, and AVG weights the six datasets by their problem counts. Baseline and benchmark accuracies are percentages. The 16K and global token-mean control tables report accuracy as a fraction and length in tokens. All main 4K monitoring and entropy-trajectory plots use AIME24.}%


\section{Additional Budget-Adaptation Trajectories}
\label{sec:appendix_budget_dynamics}

\begin{figure*}[t]
    \centering

    \begin{subfigure}[t]{0.48\textwidth}
        \centering
        \includegraphics[width=\linewidth]{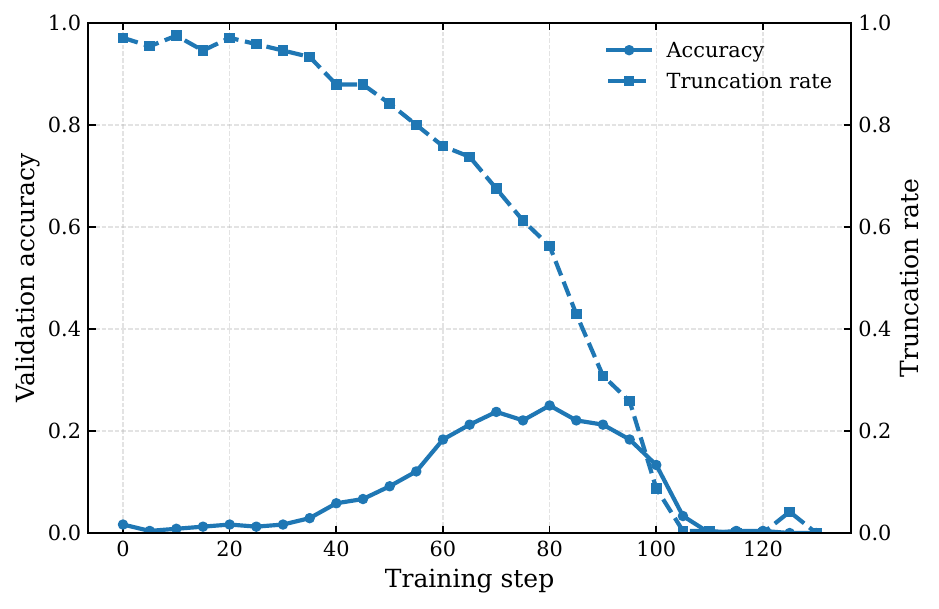}
        \caption{Qwen3-1.7B, LP}
    \end{subfigure}
    \hfill
    \begin{subfigure}[t]{0.48\textwidth}
        \centering
        \includegraphics[width=\linewidth]{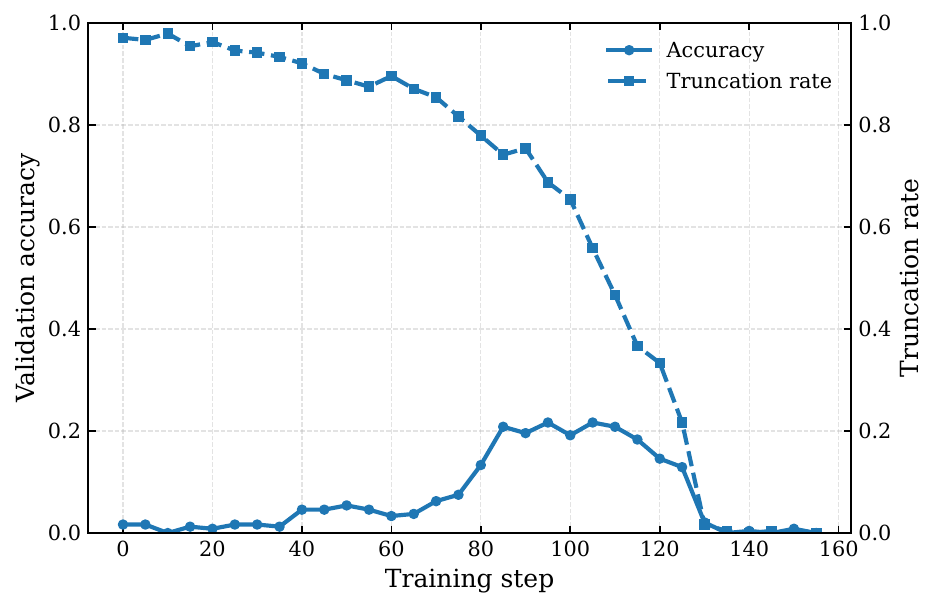}
        \caption{Qwen3-1.7B, Ent}
    \end{subfigure}

    \vspace{0.4em}

    \begin{subfigure}[t]{0.48\textwidth}
        \centering
        \includegraphics[width=\linewidth]{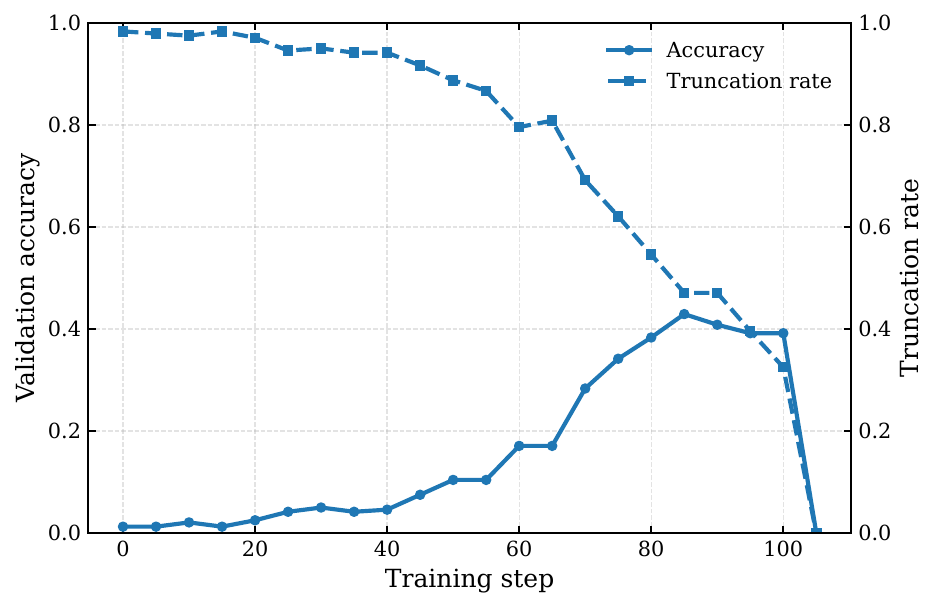}
        \caption{Qwen3-8B, LP}
    \end{subfigure}
    \hfill
    \begin{subfigure}[t]{0.48\textwidth}
        \centering
        \includegraphics[width=\linewidth]{acc_tru/Qwen3-8B-em-dapo-aime-new.pdf}
        \caption{Qwen3-8B, Ent}
    \end{subfigure}

    \vspace{0.4em}

    \begin{subfigure}[t]{0.48\textwidth}
        \centering
        \includegraphics[width=\linewidth]{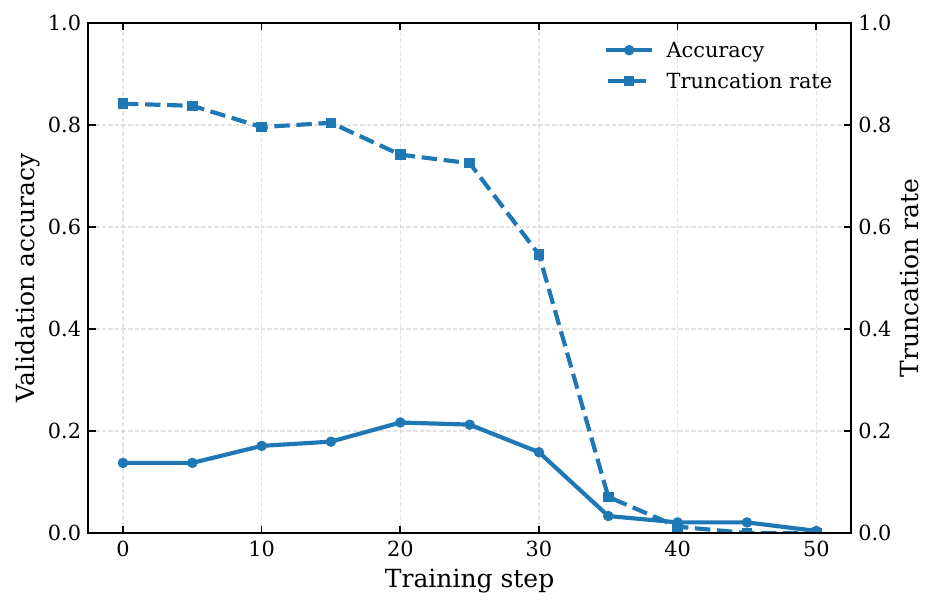}
        \caption{DeepSeek-R1-Distill-Llama-8B, LP}
    \end{subfigure}
    \hfill
    \begin{subfigure}[t]{0.48\textwidth}
        \centering
        \includegraphics[width=\linewidth]{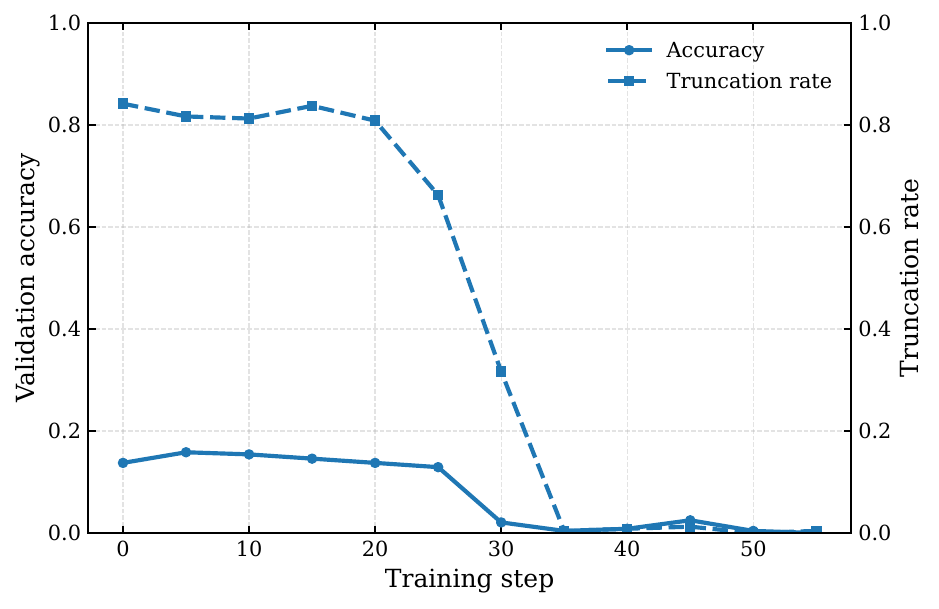}
        \caption{DeepSeek-R1-Distill-Llama-8B, Ent}
    \end{subfigure}

    \vspace{0.4em}

    \begin{subfigure}[t]{0.48\textwidth}
        \centering
        \includegraphics[width=\linewidth]{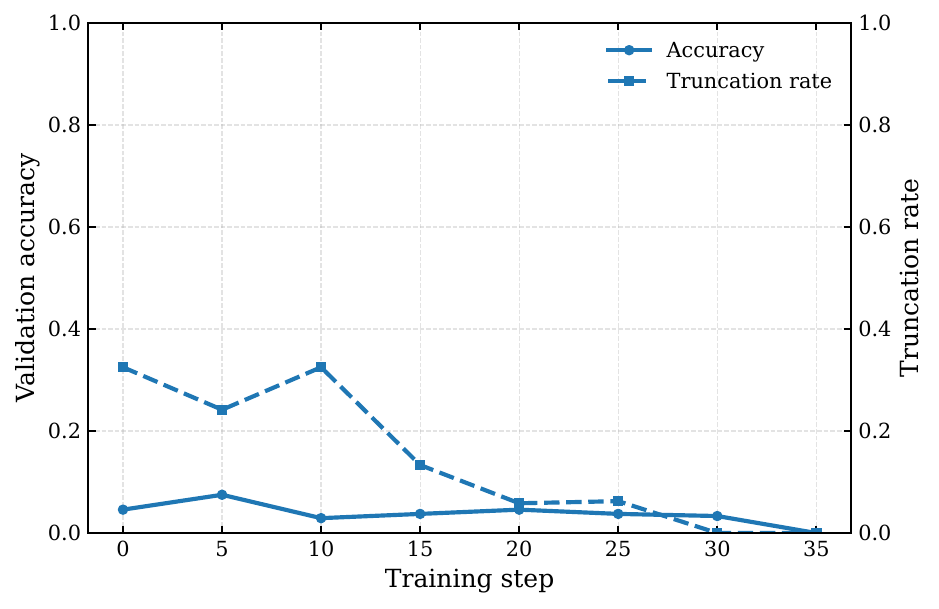}
        \caption{Llama-3.1-8B-Instruct, LP}
    \end{subfigure}
    \hfill
    \begin{subfigure}[t]{0.48\textwidth}
        \centering
        \includegraphics[width=\linewidth]{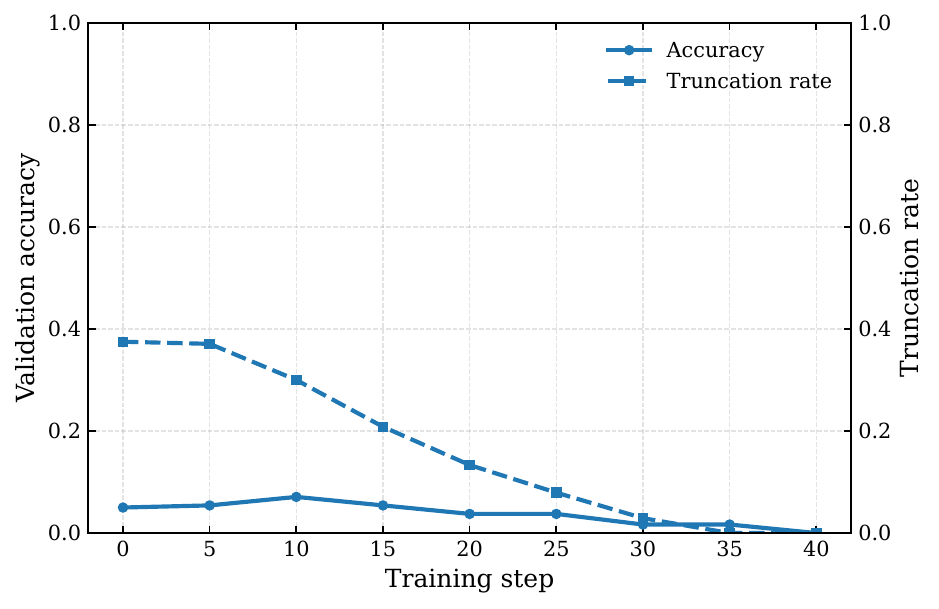}
        \caption{Llama-3.1-8B-Instruct, Ent}
    \end{subfigure}

    \caption{\textbf{Accuracy and truncation-rate trajectories under LP and Ent.} Each panel reports AIME24 validation accuracy and the fraction of responses that reach the 4K generation limit without producing an EOS token. The models differ in the duration of the transient performance regime and the onset of policy collapse.}    \label{fig:appendix_budget_dynamics}
\end{figure*}

Figure~\ref{fig:appendix_budget_dynamics} reports validation accuracy and truncation rate under Ent and LP. The truncation rate is the fraction of responses that reach the 4K generation limit without producing an EOS token. Each row presents the two intrinsic rewards for the same model.

\section{Clustering}
\label{sec:appendix_clustering}

\subsection{Trajectory Construction}
\label{sec:appendix_clustering_algo}

Algorithm~\ref{alg:trajectory_pipeline} describes the construction and clustering of token-entropy trajectories. The response to a prompt may change across training steps. We therefore use each prompt--token pair $(p,y)$ as an anchor and record the entropy assigned to token $y$ whenever it appears in the response to prompt $p$.

\begin{algorithm*}[t]
\caption{Token-Level Entropy Trajectory Construction and Clustering}
\label{alg:trajectory_pipeline}
\begin{algorithmic}[1]
\Require Validation dataset $\mathcal{D}_{val}$, evaluation steps $\mathcal{S}=\{t_1,t_2,\dots,t_M\}$, number of clusters $K$
\Ensure Cluster assignments $\mathcal{C}$ and cluster centroids
\State Initialize an empty map $\mathcal{T}$ indexed by $(p,y)$
\For{\textbf{each} evaluation step $t\in\mathcal{S}$}
    \State Load model checkpoint $\mathcal{M}_t$
    \For{\textbf{each} prompt $p\in\mathcal{D}_{val}$}
        \State Generate response $R_{p,t}$ using $\mathcal{M}_t$
        \State Extract tokens $Y_{p,t}=[y_1,\dots,y_N]$ and entropies $H_{p,t}=[h_1,\dots,h_N]$
        \For{$i=1$ \textbf{to} $N$}
            \State Append $(t,h_i)$ to $\mathcal{T}[(p,y_i)]$
        \EndFor
    \EndFor
\EndFor
\State Initialize $\mathcal{T}_{valid}\leftarrow\emptyset$
\For{\textbf{each} anchor $(p,y)$ in $\mathcal{T}$}
    \State Extract entropy sequence $\mathcal{E}_{p,y}$ from $\mathcal{T}[(p,y)]$
    \If{$|\mathcal{E}_{p,y}|\geq2$}
        \State Normalize its time axis to $[0,1]$
        \State $\mathcal{T}_{valid}\leftarrow\mathcal{T}_{valid}\cup\{\mathcal{E}_{p,y}\}$
    \EndIf
\EndFor
\State Fit Time-Series K-means with 
\RevNoLine{FIX-dtw-algorithm}{DTW}%
{} and $K$ clusters to $\mathcal{T}_{valid}$
\State \Return cluster assignments $\mathcal{C}$ and centroids
\end{algorithmic}
\end{algorithm*}

We record validation responses every five training steps. Each trajectory contains the entropy values associated with one prompt--token pair across the steps in which the token appears. Time-Series K-means with 
\Rev{FIX-dtw-appendix}{DTW}%
{} groups trajectories by their temporal shape rather than by entropy values at fixed steps.

\subsection{Sensitivity to the Number of Clusters}
\label{sec:appendix_cok}

\begin{figure*}[t]
    \centering

    \begin{subfigure}{0.32\textwidth}
        \includegraphics[width=\linewidth]{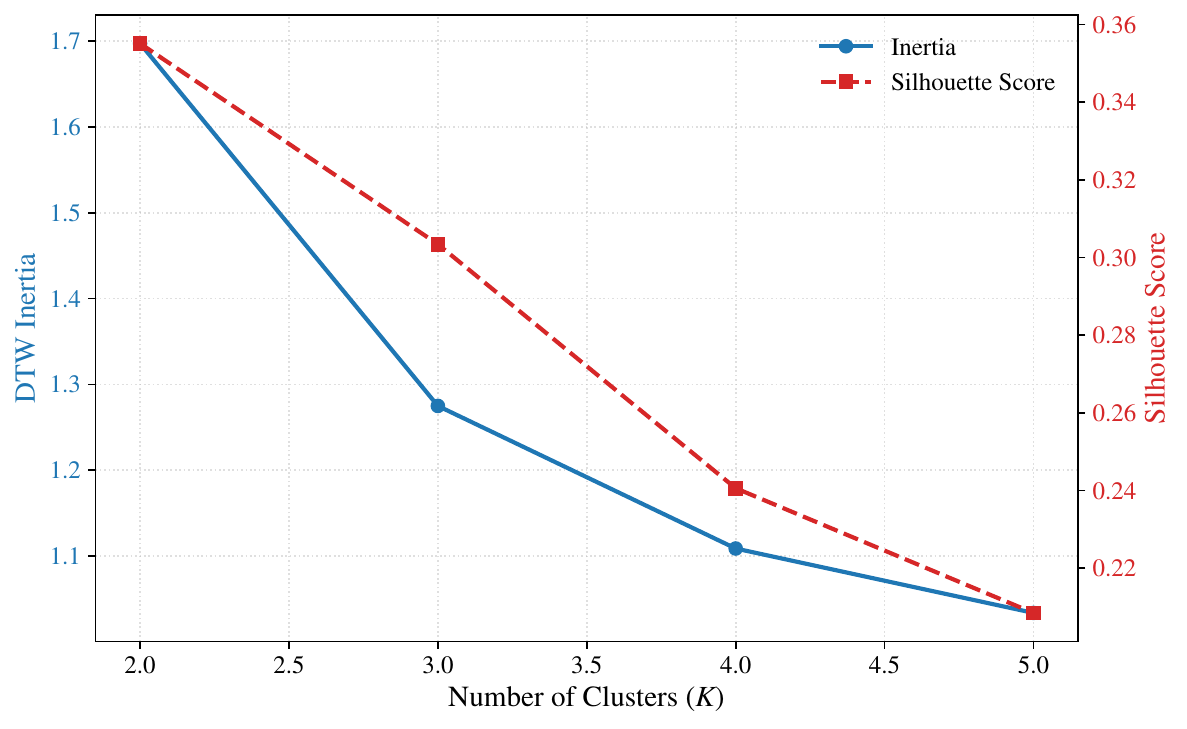}
        \caption{Qwen3-1.7B, GRPO}
    \end{subfigure}\hfill
    \begin{subfigure}{0.32\textwidth}
        \includegraphics[width=\linewidth]{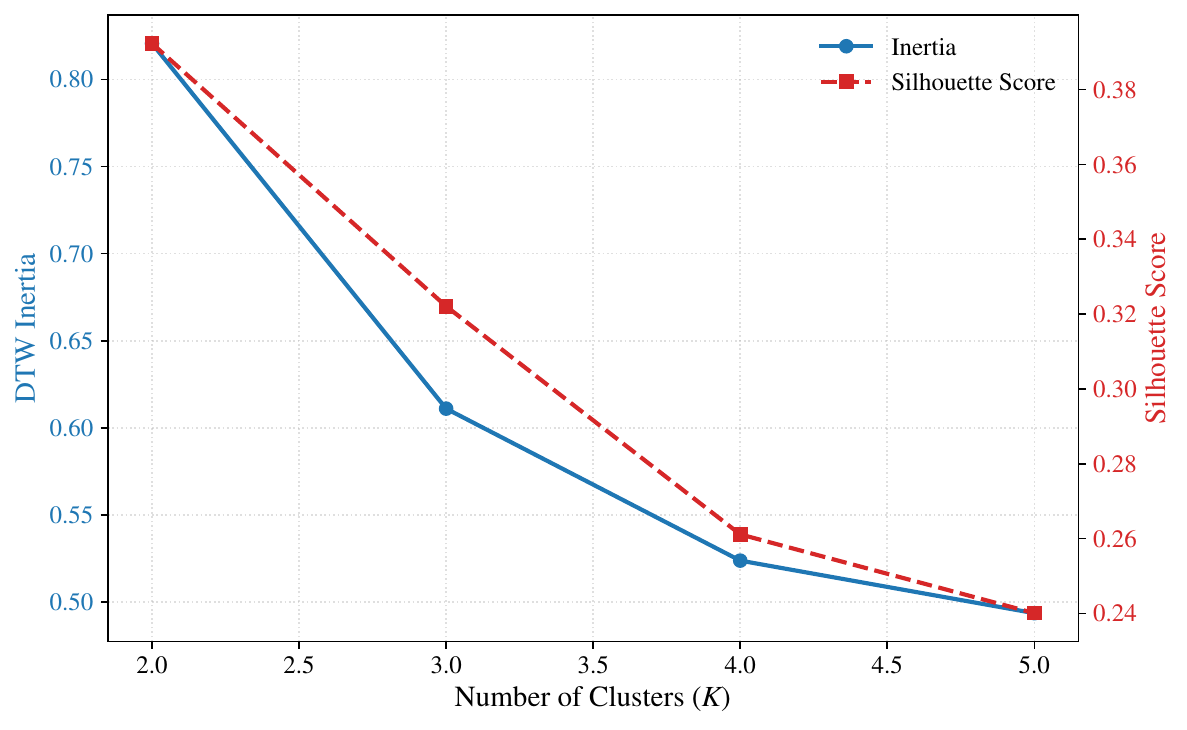}
        \caption{Qwen3-1.7B, Ent}
    \end{subfigure}\hfill
    \begin{subfigure}{0.32\textwidth}
        \includegraphics[width=\linewidth]{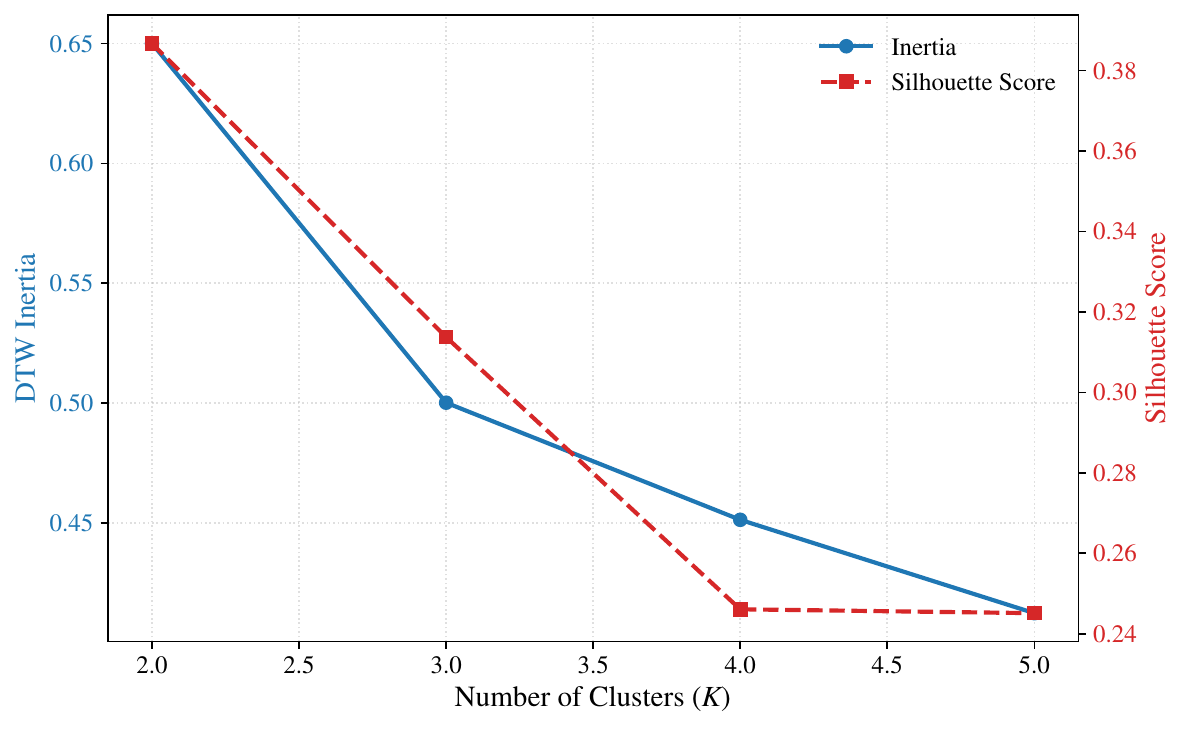}
        \caption{Qwen3-1.7B, LP}
    \end{subfigure}

    \vspace{0.3cm}

    \begin{subfigure}{0.32\textwidth}
        \includegraphics[width=\linewidth]{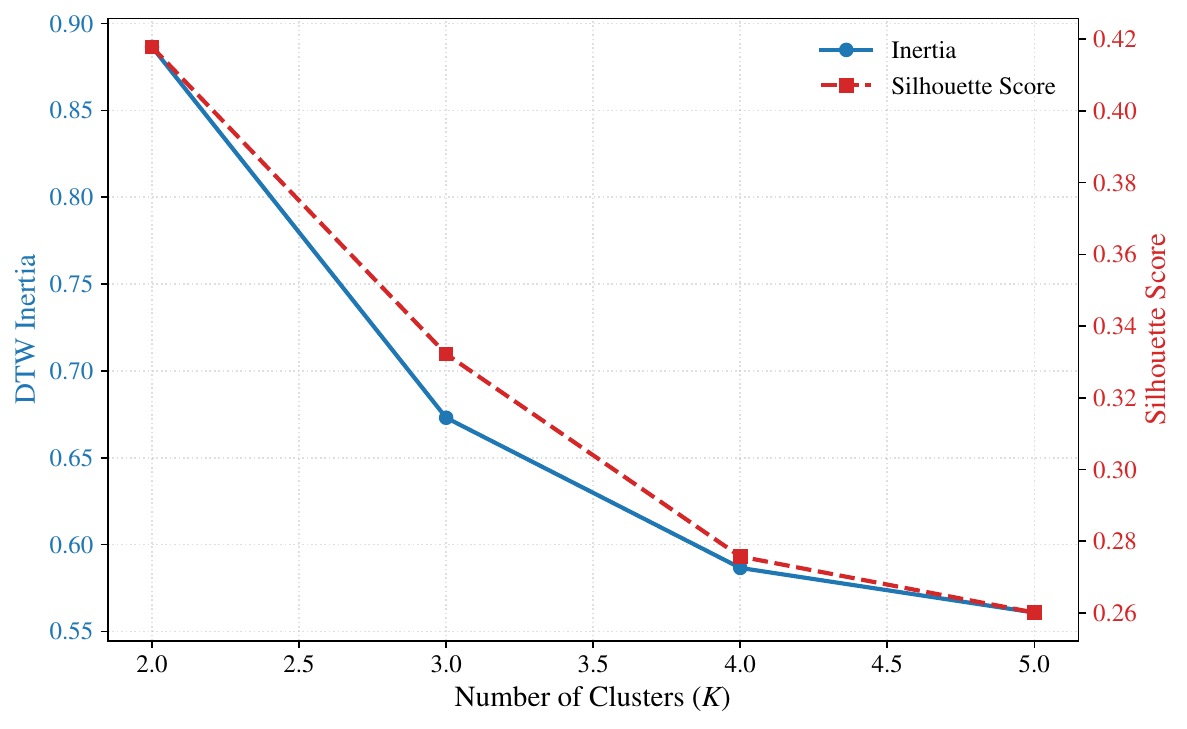}
        \caption{Qwen3-8B, GRPO}
    \end{subfigure}\hfill
    \begin{subfigure}{0.32\textwidth}
        \includegraphics[width=\linewidth]{k_select/k_evaluation_entropy_Qwen3-8B-em-dapo-aime-new.pdf}
        \caption{Qwen3-8B, Ent}
    \end{subfigure}\hfill
    \begin{subfigure}{0.32\textwidth}
        \includegraphics[width=\linewidth]{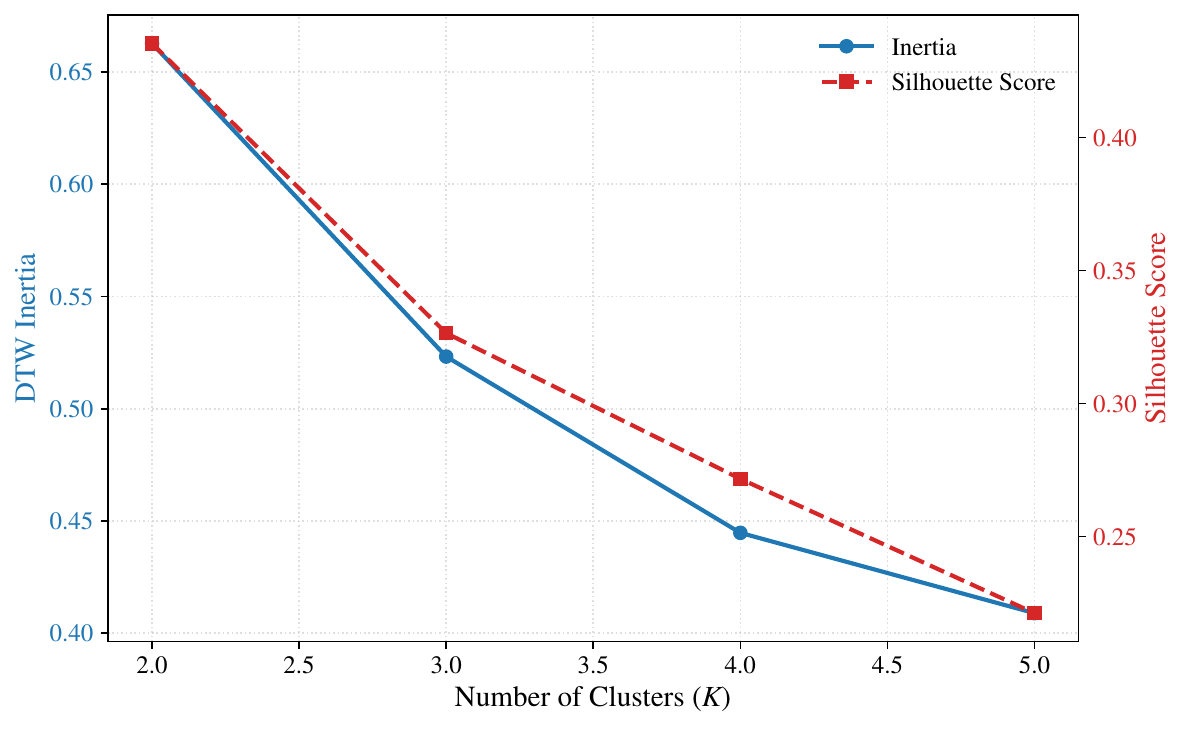}
        \caption{Qwen3-8B, LP}
    \end{subfigure}

    \vspace{0.3cm}

    \begin{subfigure}{0.32\textwidth}
        \includegraphics[width=\linewidth]{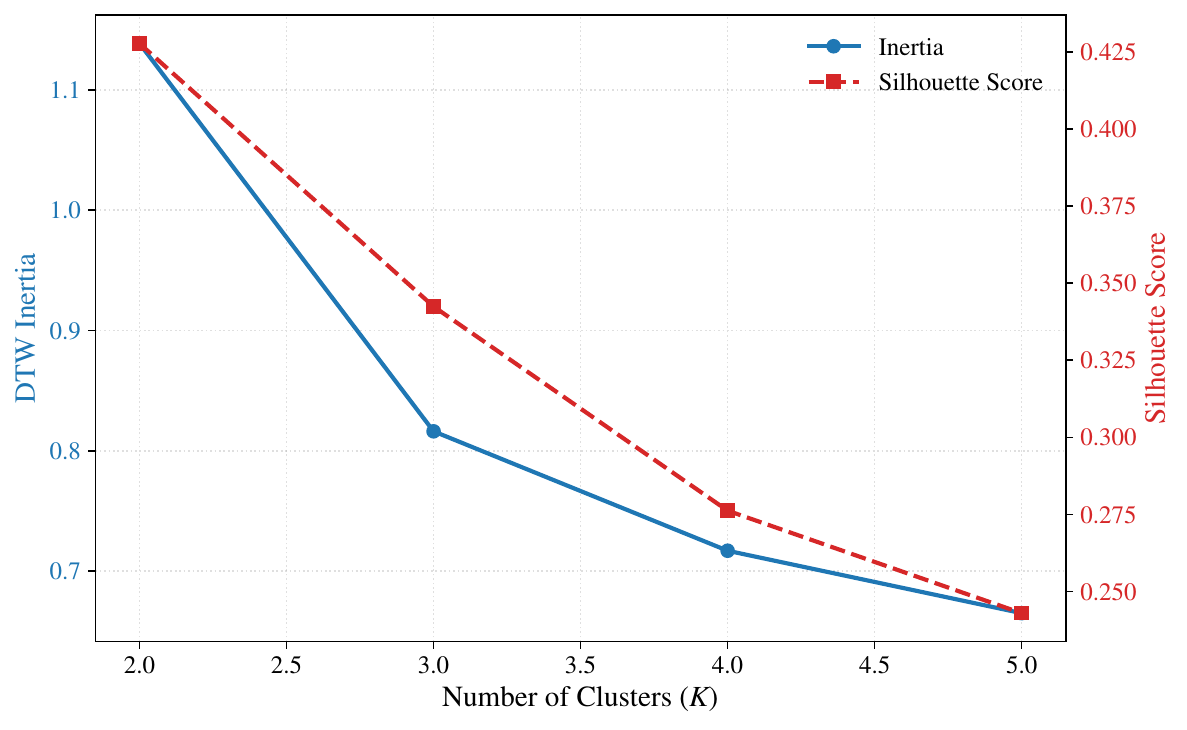}
        \caption{DeepSeek-R1-Distill-Llama-8B, GRPO}
    \end{subfigure}\hfill
    \begin{subfigure}{0.32\textwidth}
        \includegraphics[width=\linewidth]{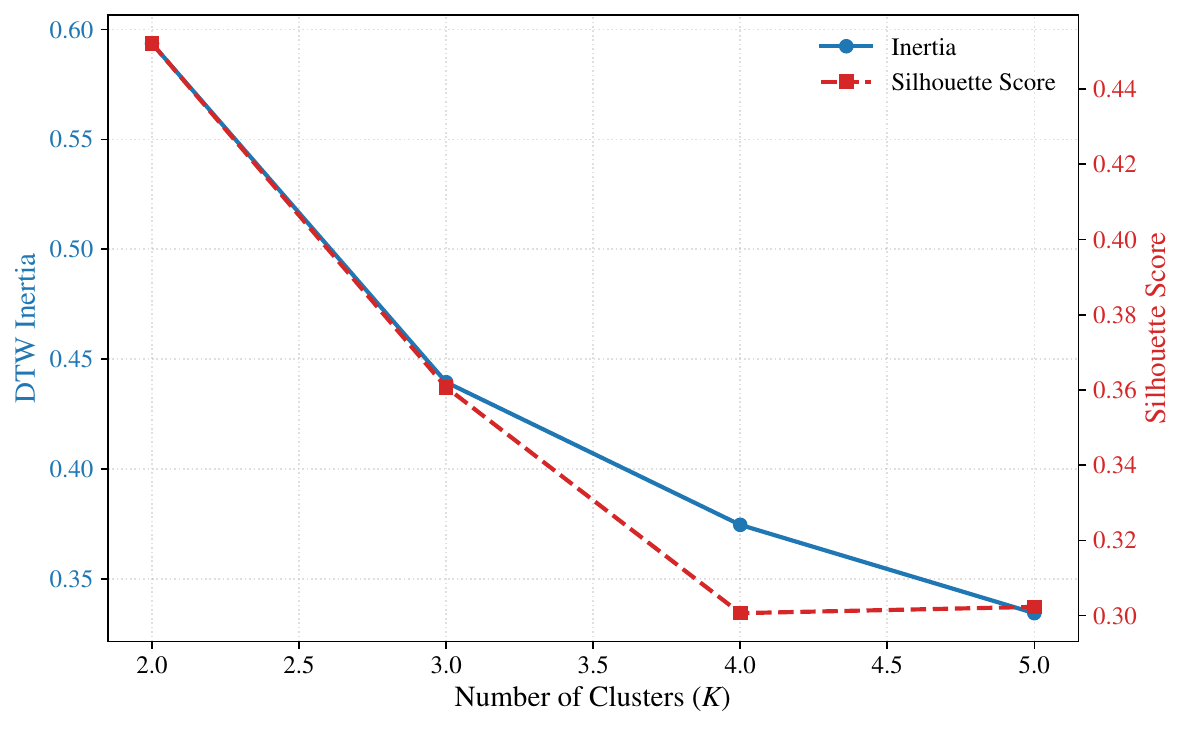}
        \caption{DeepSeek-R1-Distill-Llama-8B, Ent}
    \end{subfigure}\hfill
    \begin{subfigure}{0.32\textwidth}
        \includegraphics[width=\linewidth]{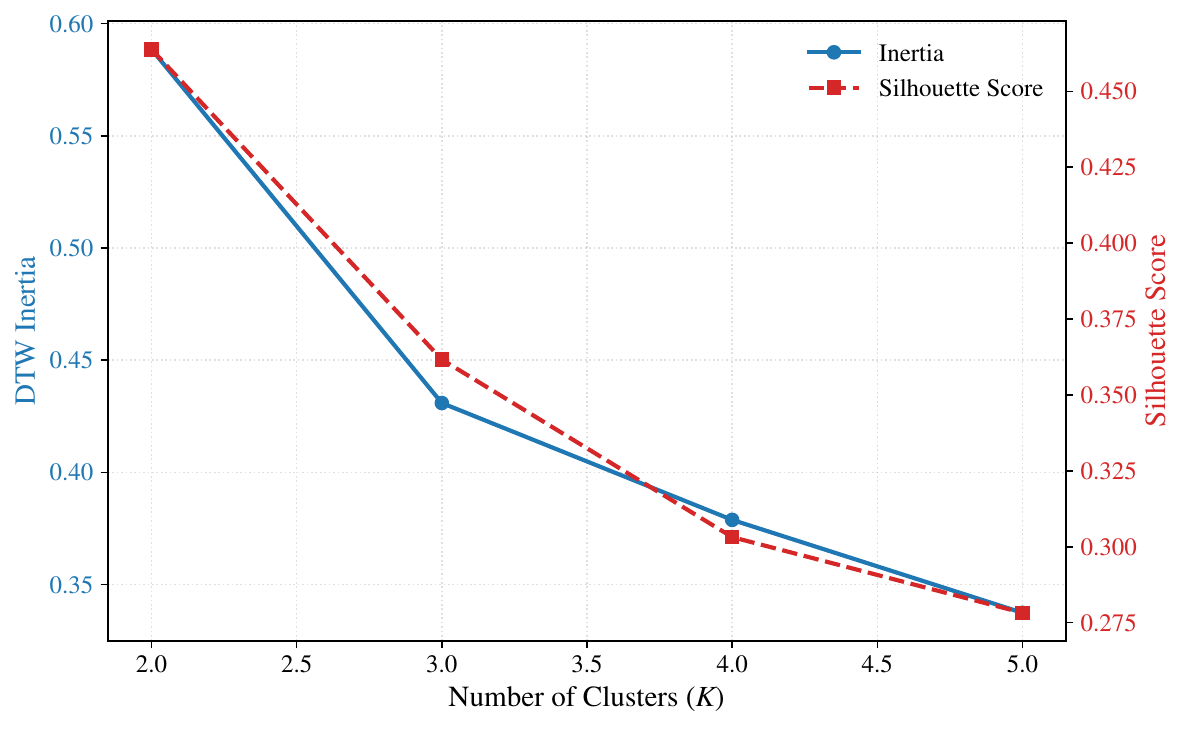}
        \caption{DeepSeek-R1-Distill-Llama-8B, LP}
    \end{subfigure}

    \vspace{0.3cm}

    \begin{subfigure}{0.32\textwidth}
        \includegraphics[width=\linewidth]{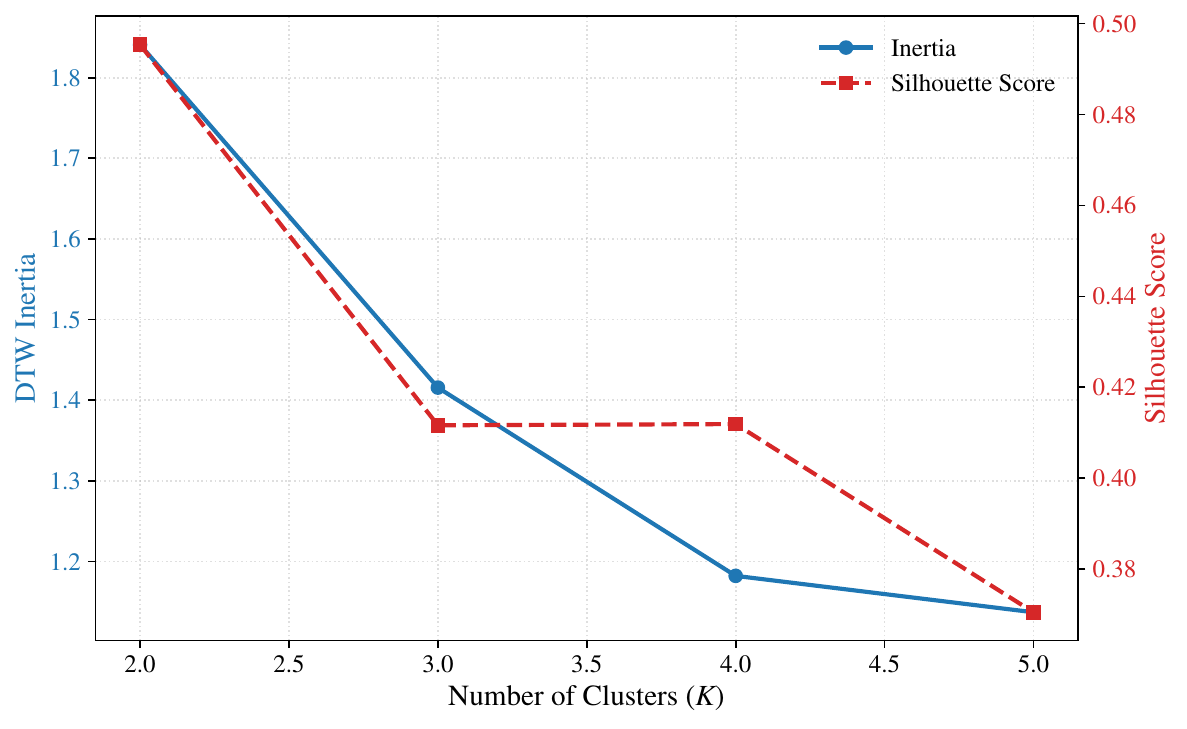}
        \caption{Llama-3.1-8B-Instruct, GRPO}
    \end{subfigure}\hfill
    \begin{subfigure}{0.32\textwidth}
        \includegraphics[width=\linewidth]{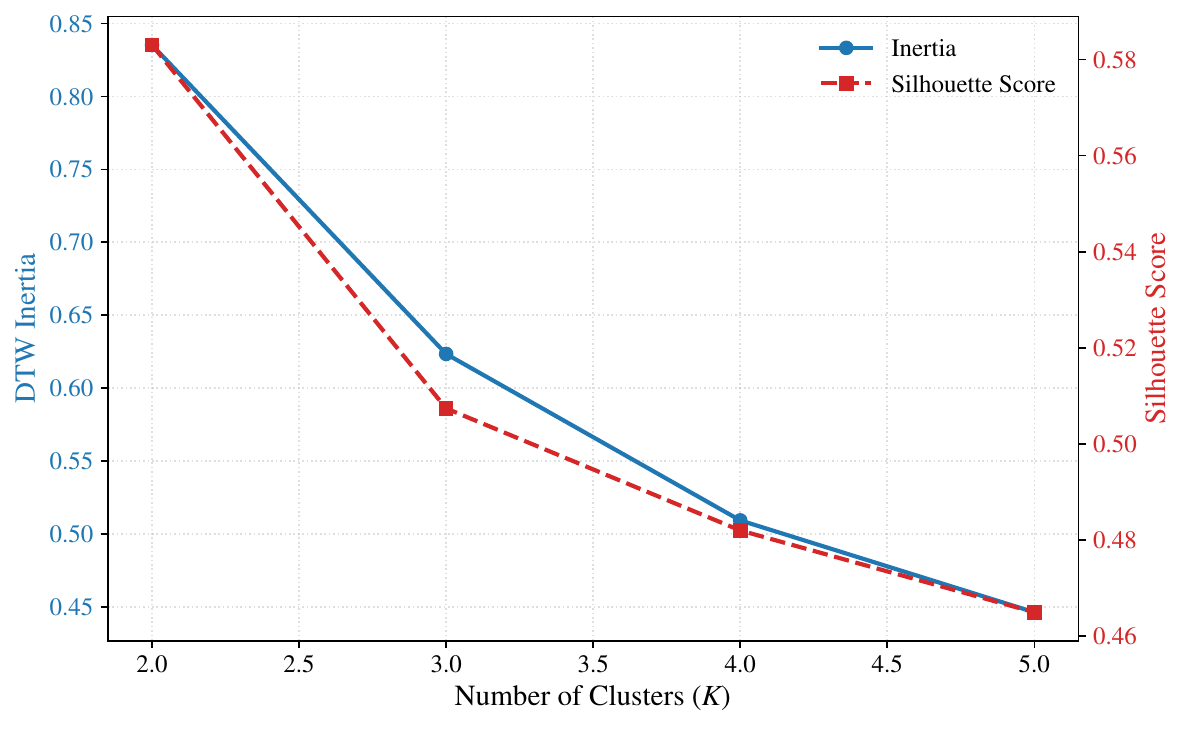}
        \caption{Llama-3.1-8B-Instruct, Ent}
    \end{subfigure}\hfill
    \begin{subfigure}{0.32\textwidth}
        \includegraphics[width=\linewidth]{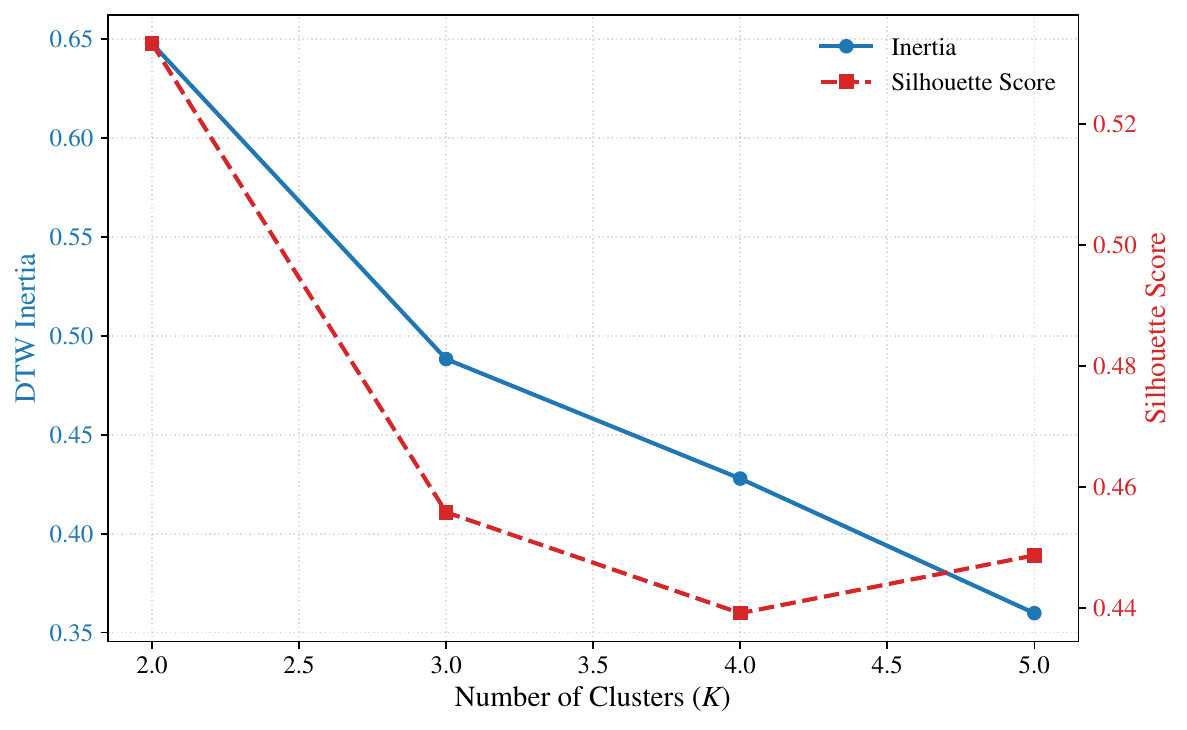}
        \caption{Llama-3.1-8B-Instruct, LP}
    \end{subfigure}

    \caption{\textbf{Evaluation of the number of clusters.} The plots report DTW Inertia and Silhouette Scores for $K\in\{2,3,4,5\}$. $K=2$ often gives the highest Silhouette Score, while $K=3$ is commonly located near the DTW Inertia elbow.}
    \label{fig:comprehensive_k_evaluation}
\end{figure*}

Figure~\ref{fig:comprehensive_k_evaluation} reports DTW Inertia and Silhouette Scores for $K\in\{2,3,4,5\}$. The two criteria favor different resolutions. $K=2$ often obtains the highest Silhouette Score, while the DTW Inertia curves show an elbow near $K=3$. We therefore use both $K=2$ and $K=3$ as representations.

\subsection{Entropy-Trajectory Clusters}
\label{sec:appendix_cok_clustering}

\begin{figure*}[t]
    \centering

    \begin{subfigure}{0.32\textwidth}
        \includegraphics[width=\linewidth]{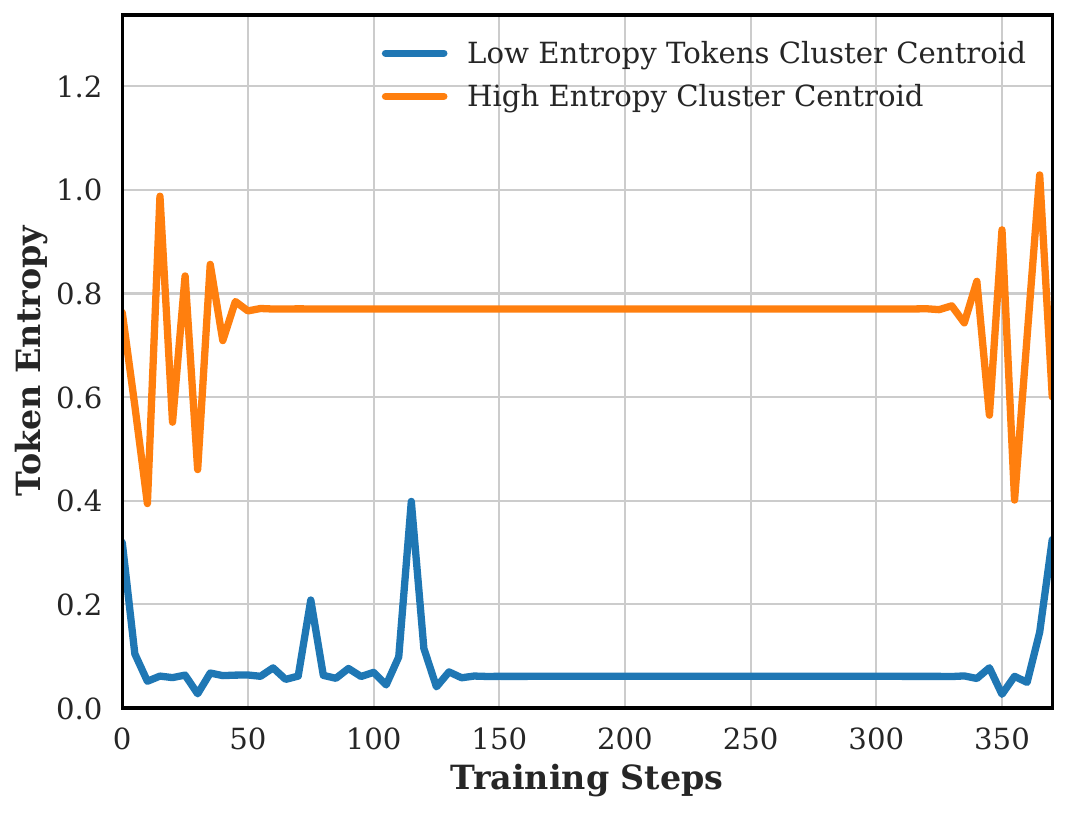}
        \caption{Qwen3-1.7B, GRPO}
    \end{subfigure}\hfill
    \begin{subfigure}{0.32\textwidth}
        \includegraphics[width=\linewidth]{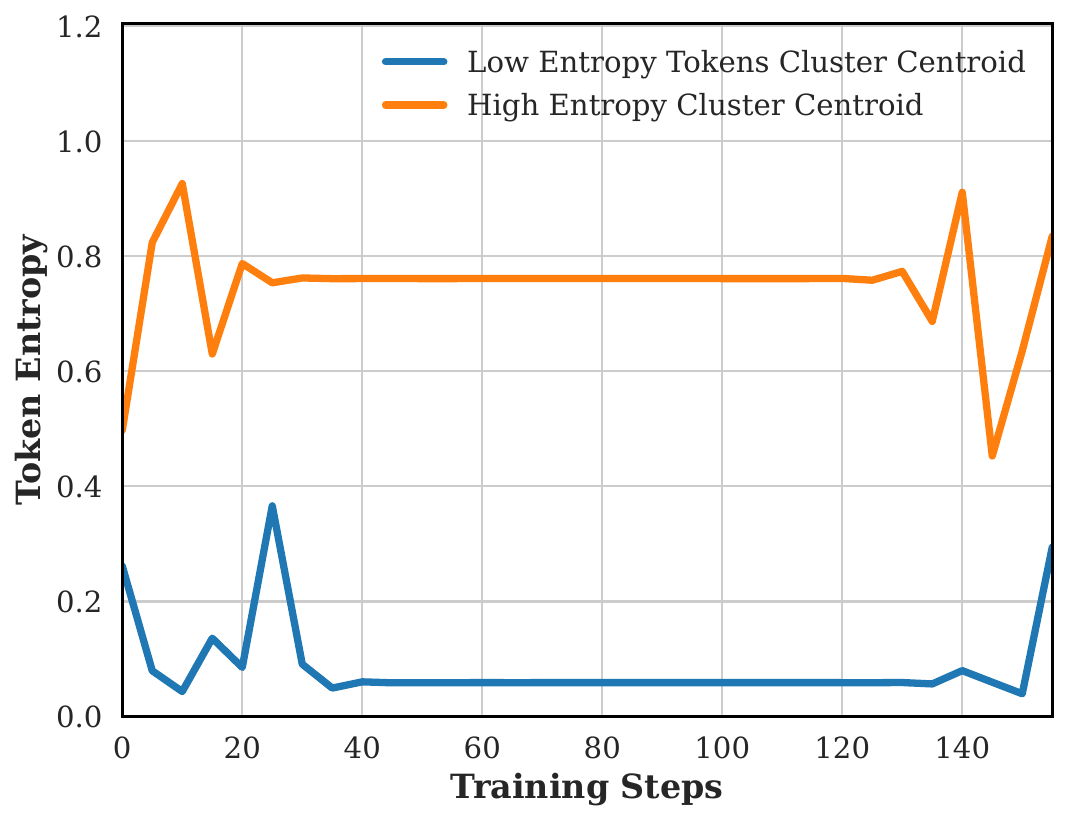}
        \caption{Qwen3-1.7B, Ent}
    \end{subfigure}\hfill
    \begin{subfigure}{0.32\textwidth}
        \includegraphics[width=\linewidth]{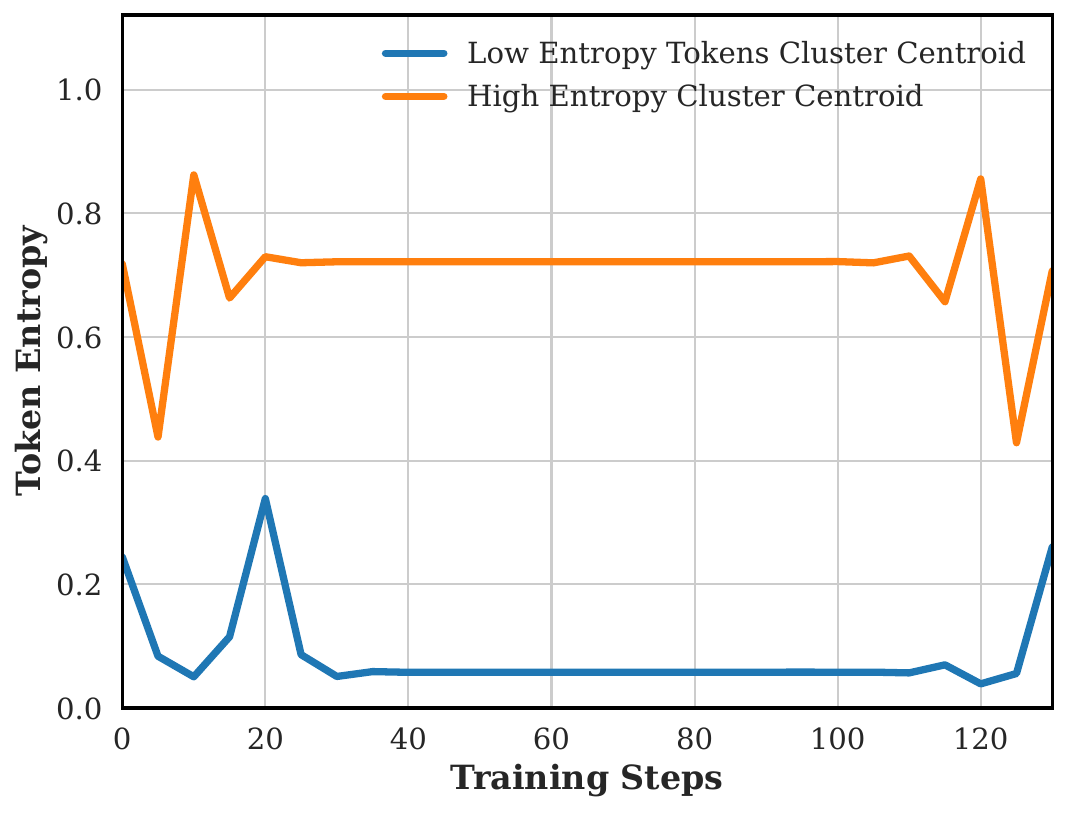}
        \caption{Qwen3-1.7B, LP}
    \end{subfigure}

    \vspace{0.3cm}

    \begin{subfigure}{0.32\textwidth}
        \includegraphics[width=\linewidth]{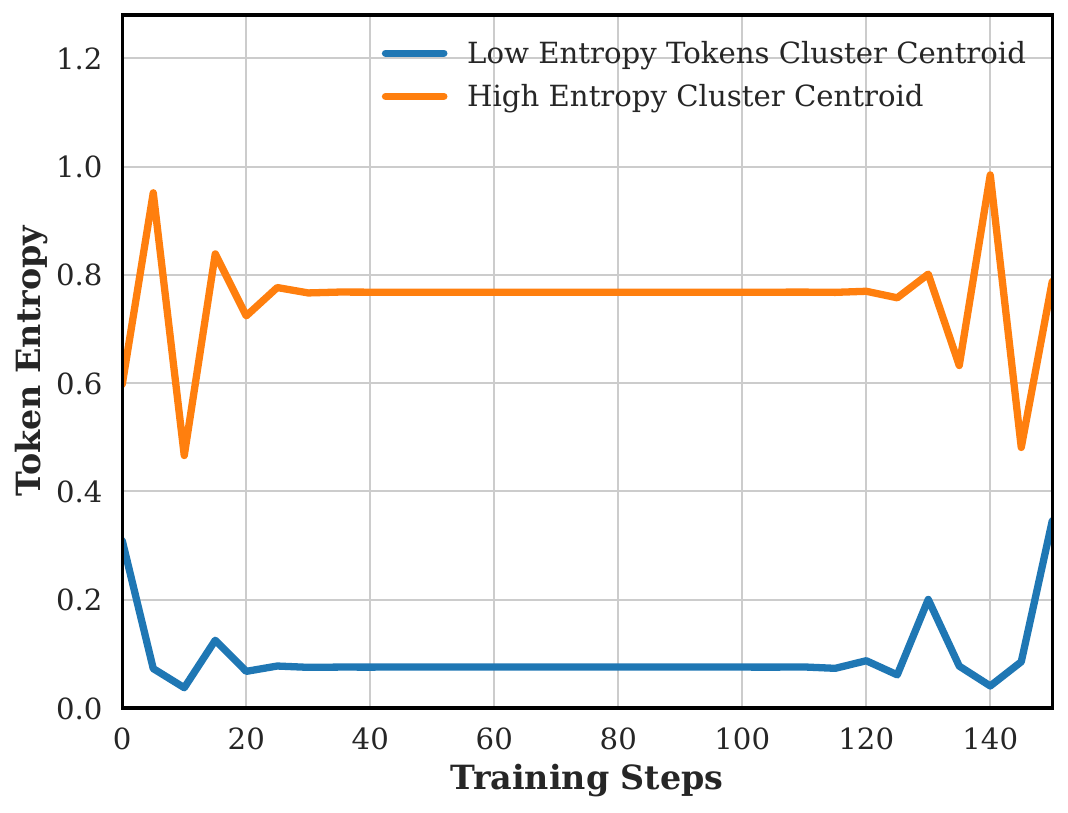}
        \caption{Qwen3-8B, GRPO}
    \end{subfigure}\hfill
    \begin{subfigure}{0.32\textwidth}
        \includegraphics[width=\linewidth]{figure4_k2_individual/figure4_k2_Qwen3-8B-em-dapo-aime-new.pdf}
        \caption{Qwen3-8B, Ent}
    \end{subfigure}\hfill
    \begin{subfigure}{0.32\textwidth}
        \includegraphics[width=\linewidth]{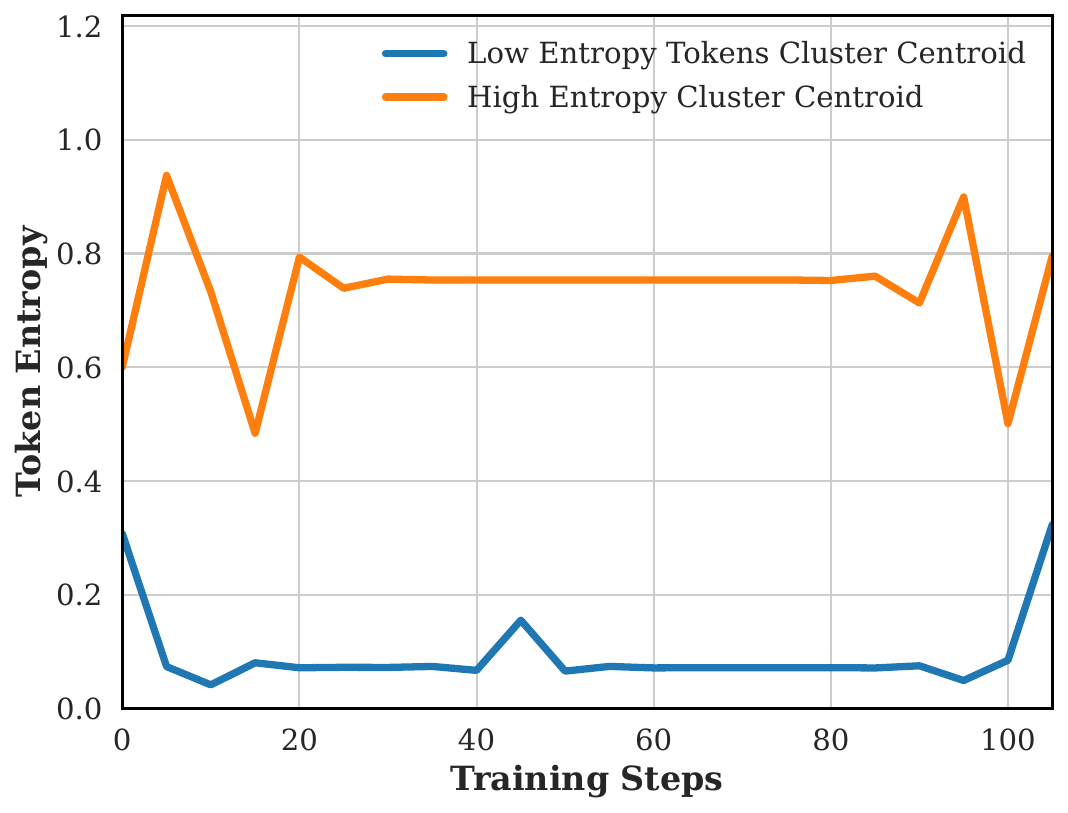}
        \caption{Qwen3-8B, LP}
    \end{subfigure}

    \vspace{0.3cm}

    \begin{subfigure}{0.32\textwidth}
        \includegraphics[width=\linewidth]{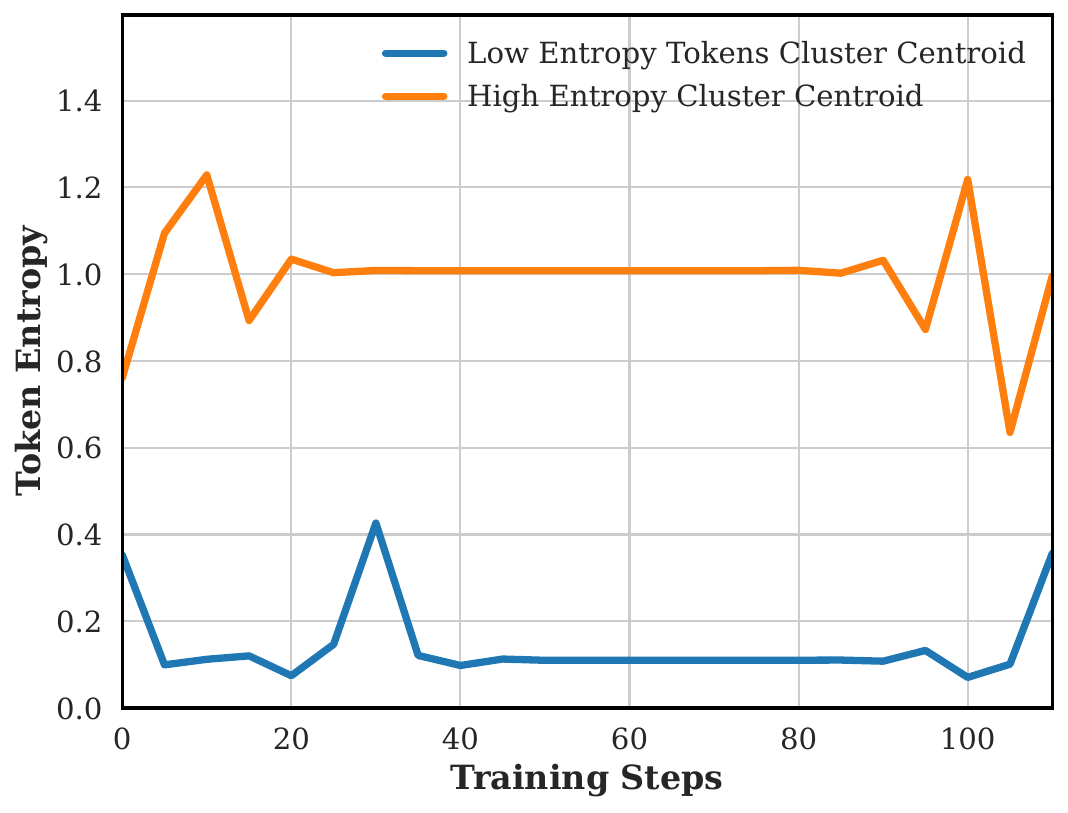}
        \caption{DeepSeek-R1-Distill-Llama-8B, GRPO}
    \end{subfigure}\hfill
    \begin{subfigure}{0.32\textwidth}
        \includegraphics[width=\linewidth]{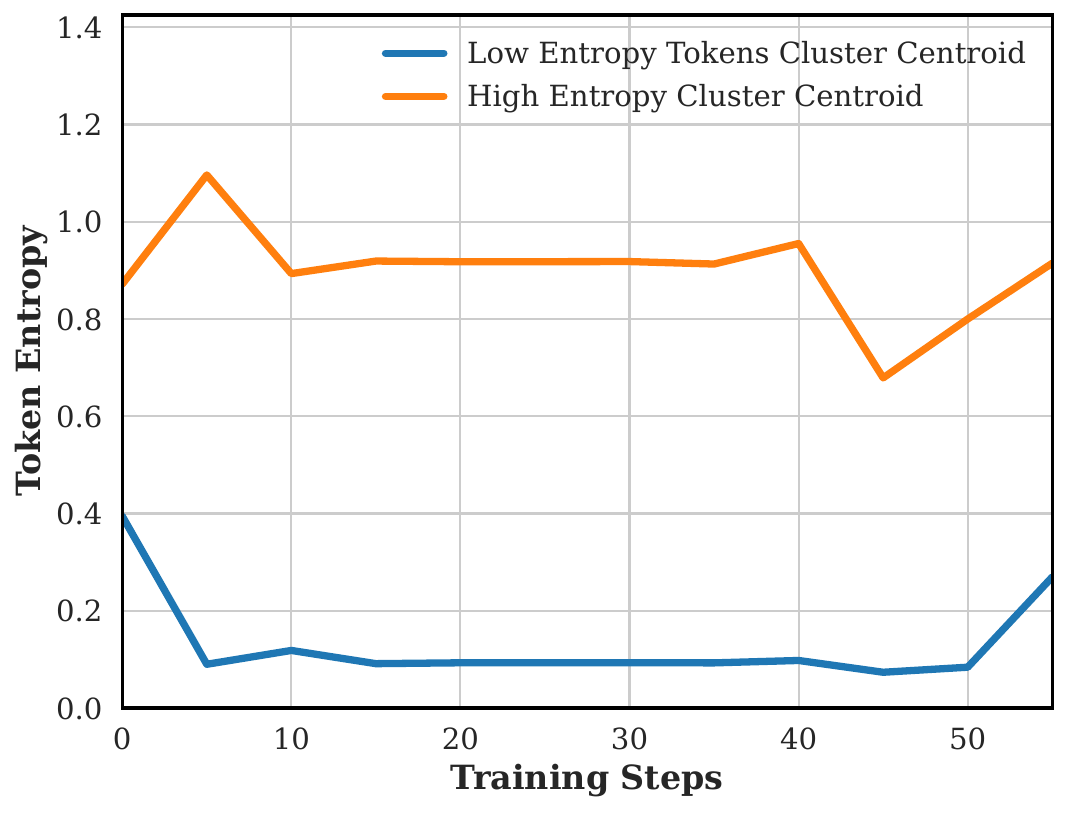}
        \caption{DeepSeek-R1-Distill-Llama-8B, Ent}
    \end{subfigure}\hfill
    \begin{subfigure}{0.32\textwidth}
        \includegraphics[width=\linewidth]{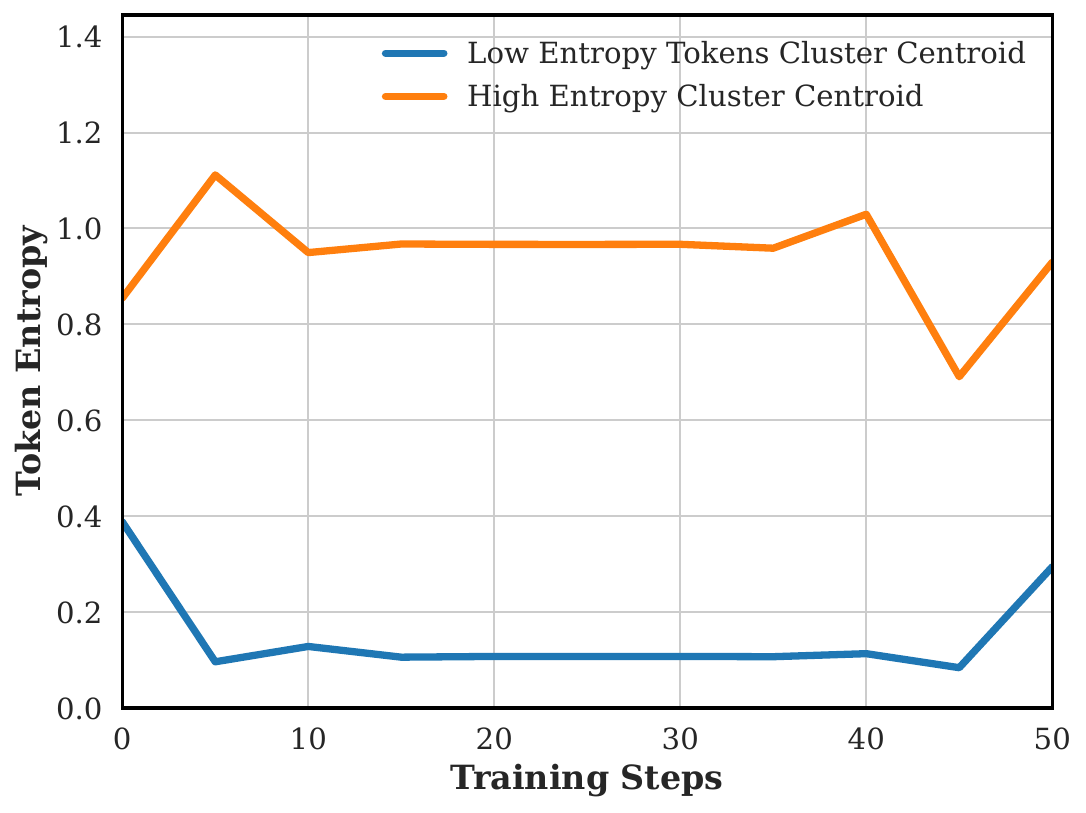}
        \caption{DeepSeek-R1-Distill-Llama-8B, LP}
    \end{subfigure}

    \vspace{0.3cm}

    \begin{subfigure}{0.32\textwidth}
        \includegraphics[width=\linewidth]{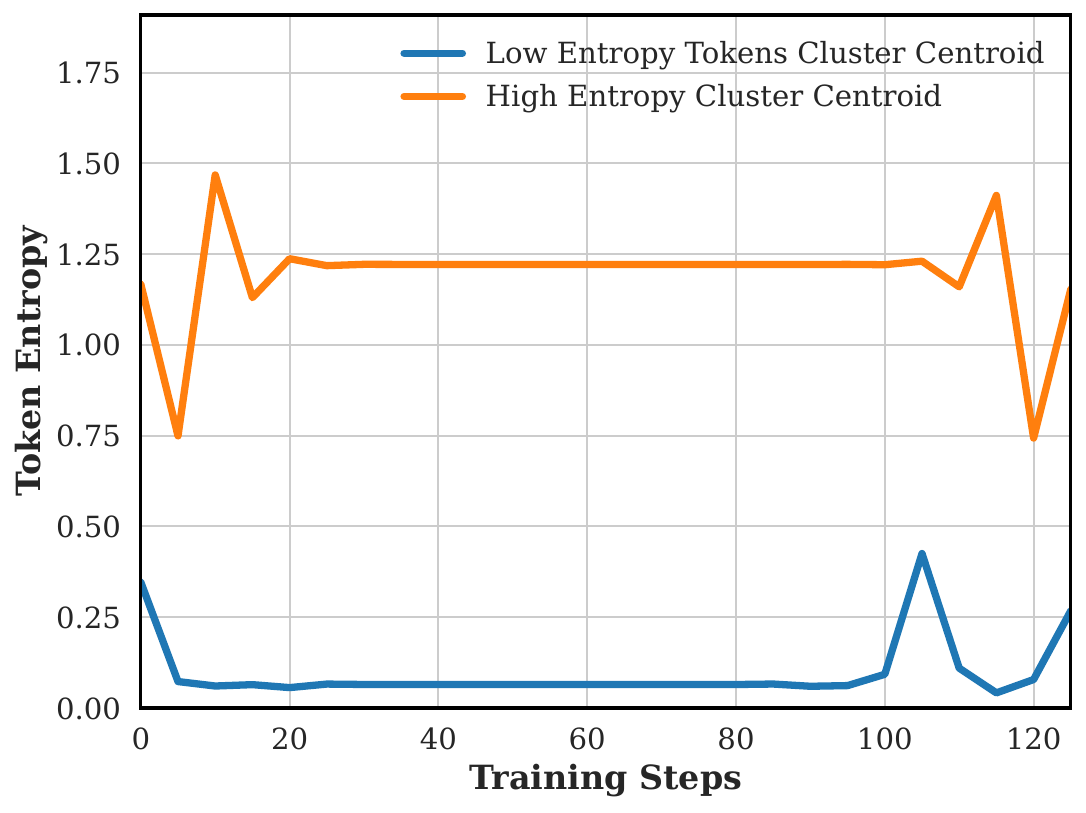}
        \caption{Llama-3.1-8B-Instruct, GRPO}
    \end{subfigure}\hfill
    \begin{subfigure}{0.32\textwidth}
        \includegraphics[width=\linewidth]{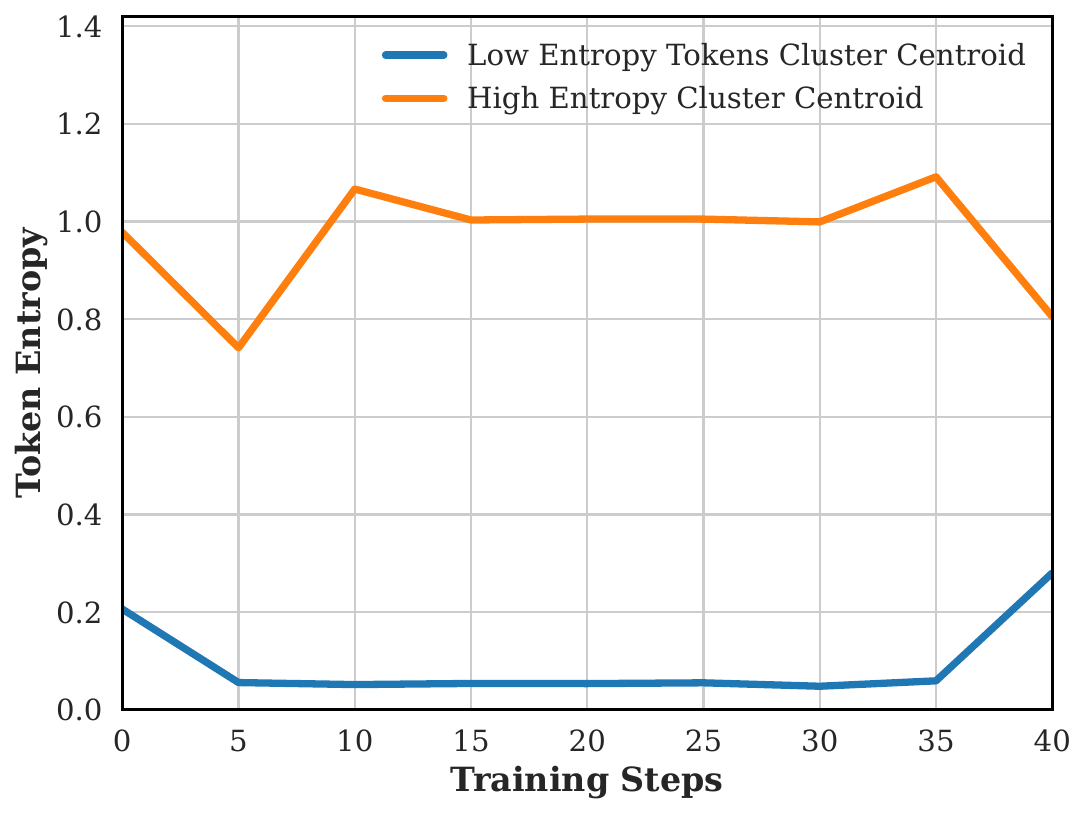}
        \caption{Llama-3.1-8B-Instruct, Ent}
    \end{subfigure}\hfill
    \begin{subfigure}{0.32\textwidth}
        \includegraphics[width=\linewidth]{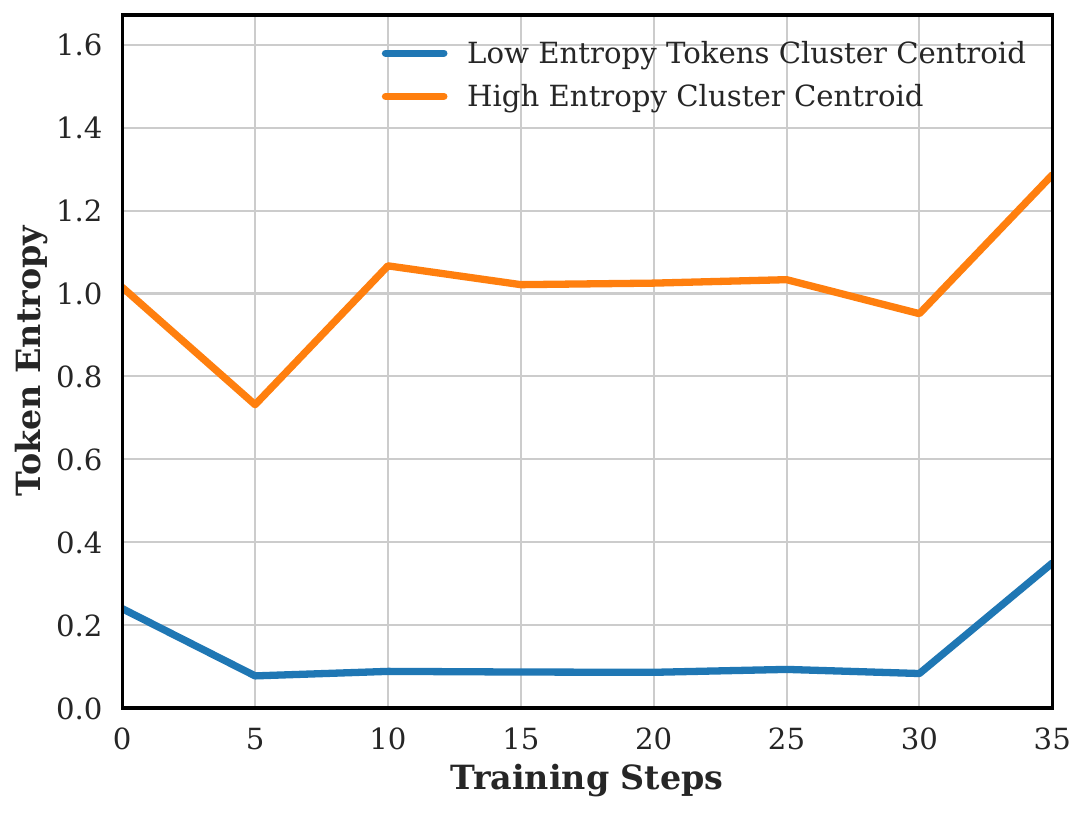}
        \caption{Llama-3.1-8B-Instruct, LP}
    \end{subfigure}

    \caption{\textbf{Token-entropy trajectory clusters with $K=2$.} Each configuration is represented by two trajectory centroids. The two-cluster representation provides a coarse summary of the entropy dynamics across models and training objectives.}
    \label{fig:comprehensive_entropy_clusters_k2}
\end{figure*}

\begin{figure*}[t]
    \centering

    \begin{subfigure}{0.32\textwidth}
        \includegraphics[width=\linewidth]{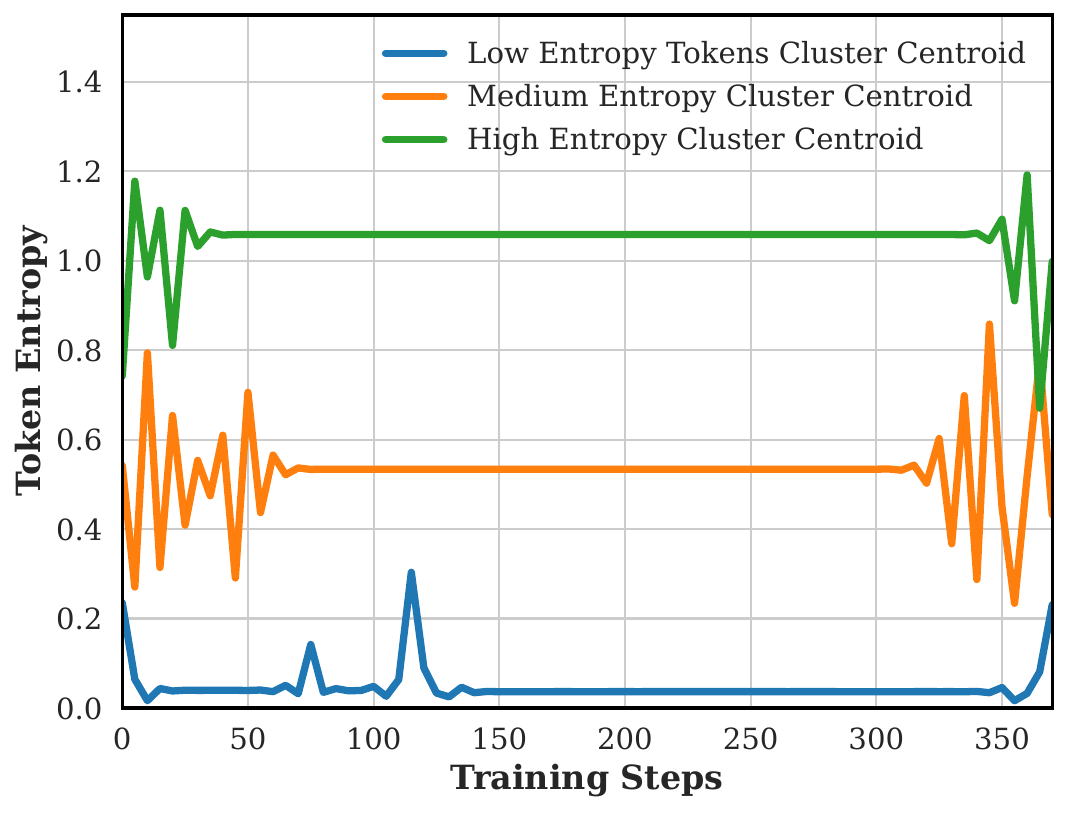}
        \caption{Qwen3-1.7B, GRPO}
    \end{subfigure}\hfill
    \begin{subfigure}{0.32\textwidth}
        \includegraphics[width=\linewidth]{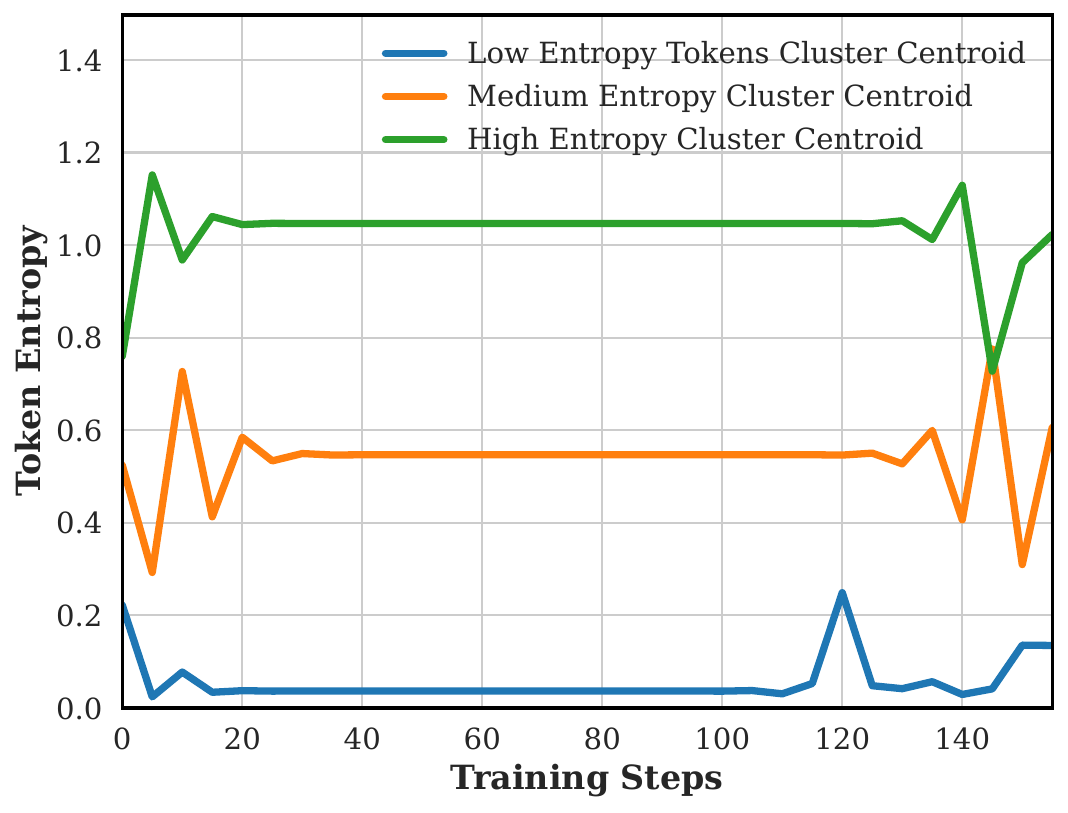}
        \caption{Qwen3-1.7B, Ent}
    \end{subfigure}\hfill
    \begin{subfigure}{0.32\textwidth}
        \includegraphics[width=\linewidth]{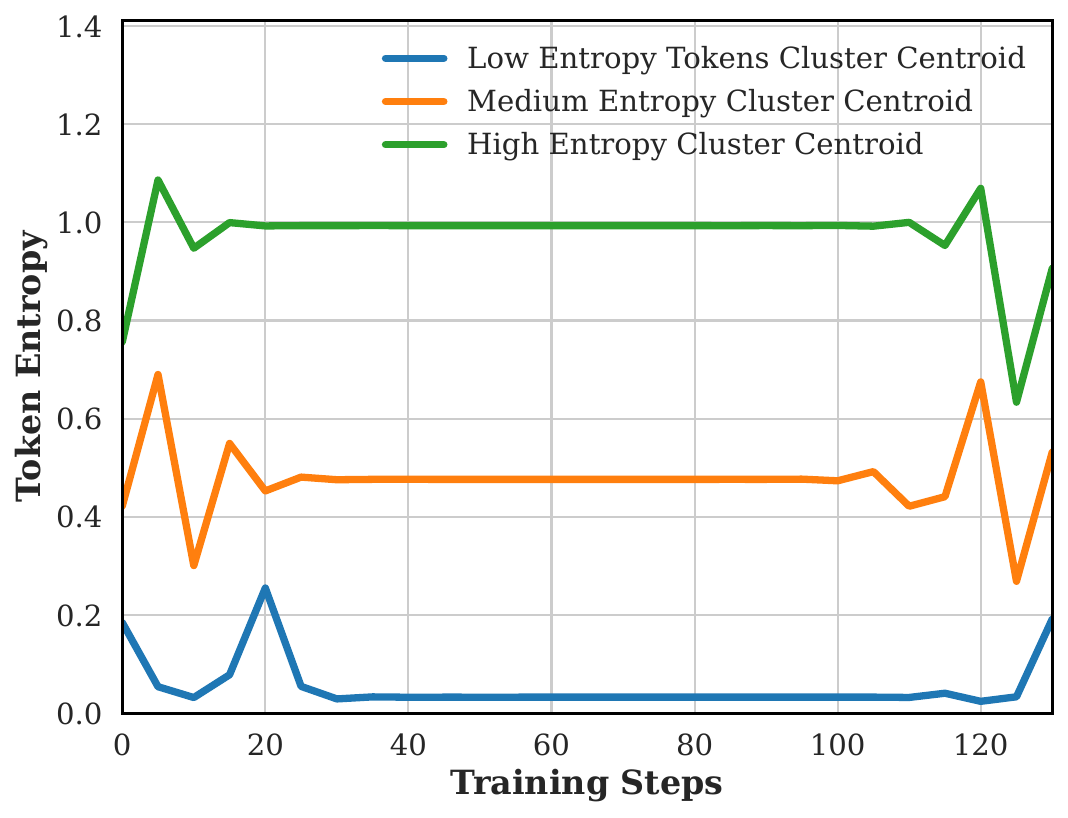}
        \caption{Qwen3-1.7B, LP}
    \end{subfigure}

    \vspace{0.3cm}

    \begin{subfigure}{0.32\textwidth}
        \includegraphics[width=\linewidth]{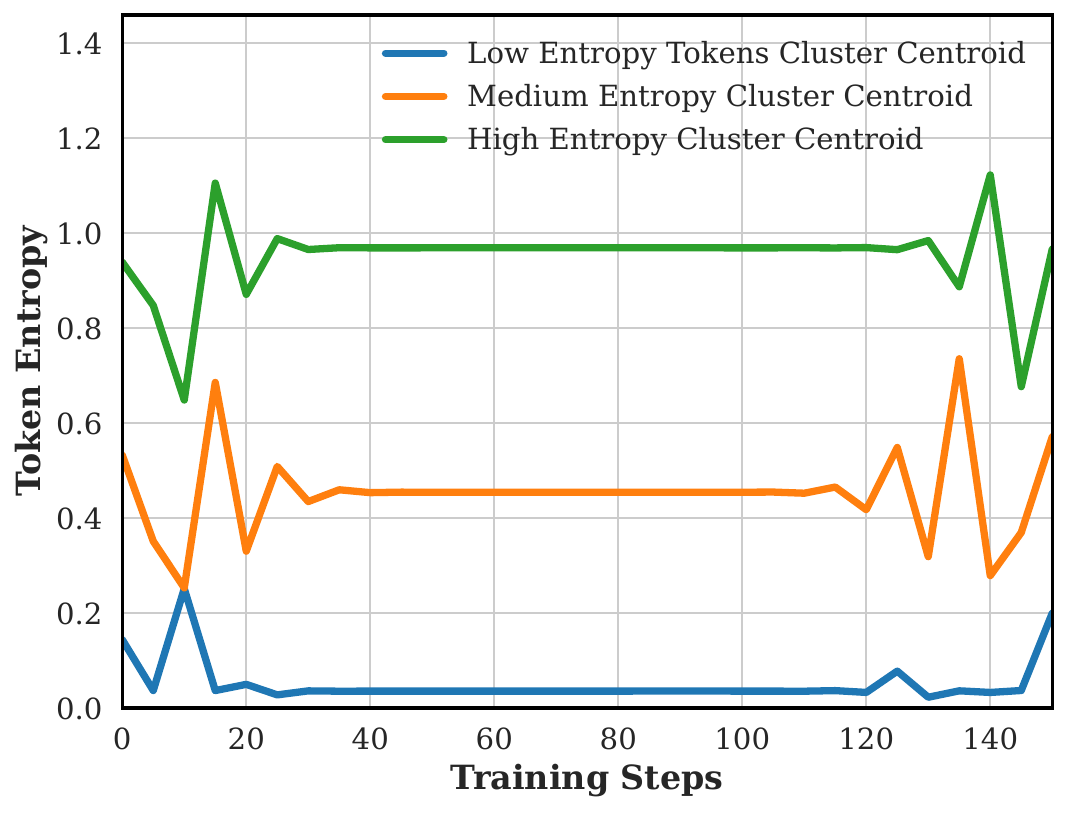}
        \caption{Qwen3-8B, GRPO}
    \end{subfigure}\hfill
    \begin{subfigure}{0.32\textwidth}
        \includegraphics[width=\linewidth]{cluster/entropy_clusters_combined_entropy_Qwen3-8B-em-dapo-aime-new.pdf}
        \caption{Qwen3-8B, Ent}
    \end{subfigure}\hfill
    \begin{subfigure}{0.32\textwidth}
        \includegraphics[width=\linewidth]{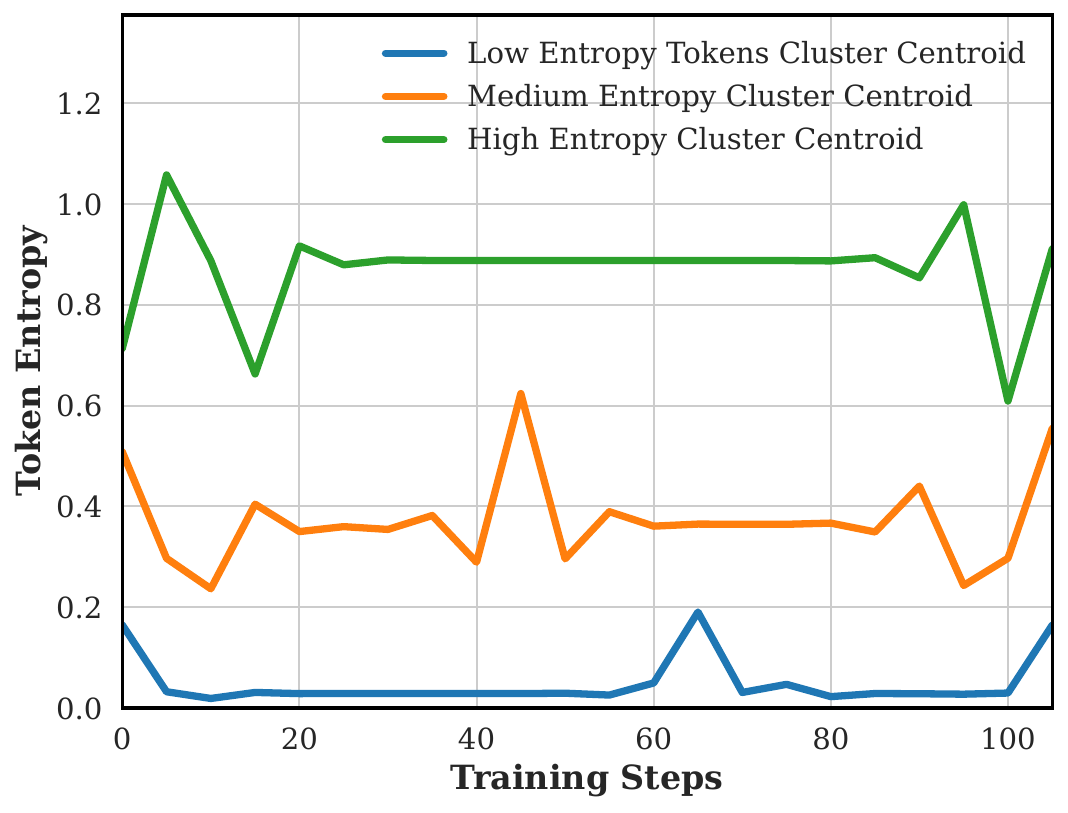}
        \caption{Qwen3-8B, LP}
    \end{subfigure}

    \vspace{0.3cm}

    \begin{subfigure}{0.32\textwidth}
        \includegraphics[width=\linewidth]{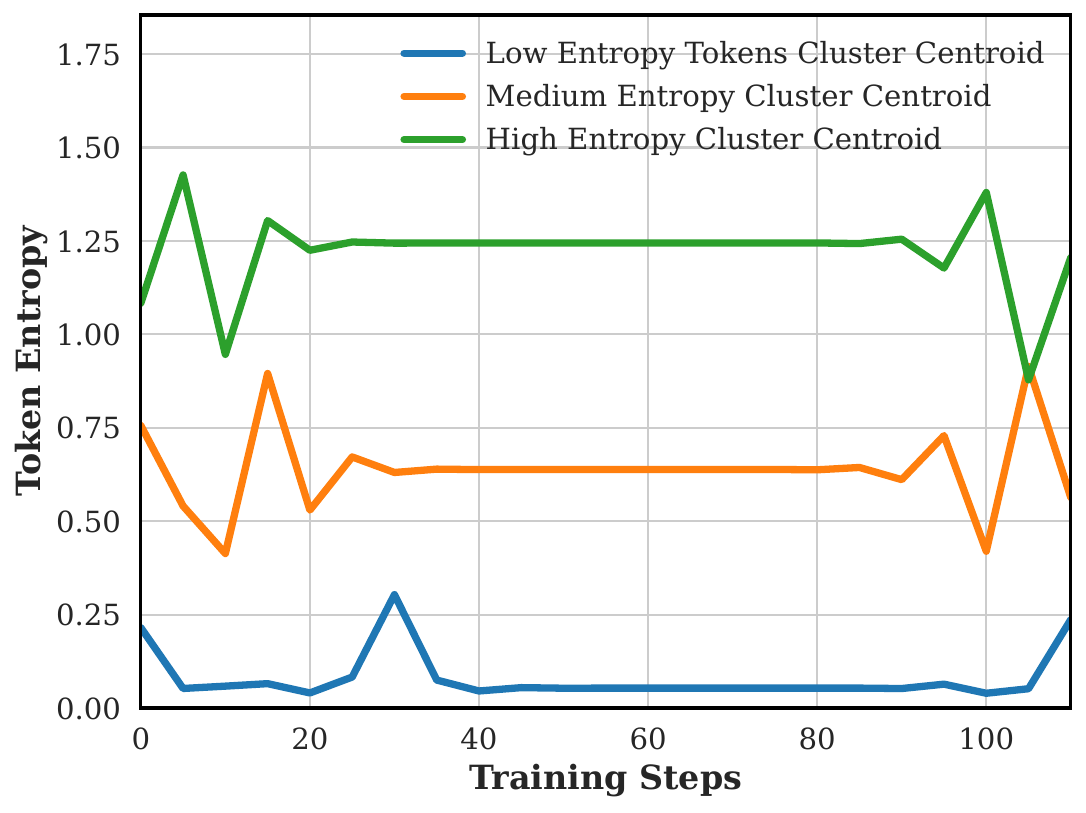}
        \caption{DeepSeek-R1-Distill-Llama-8B, GRPO}
    \end{subfigure}\hfill
    \begin{subfigure}{0.32\textwidth}
        \includegraphics[width=\linewidth]{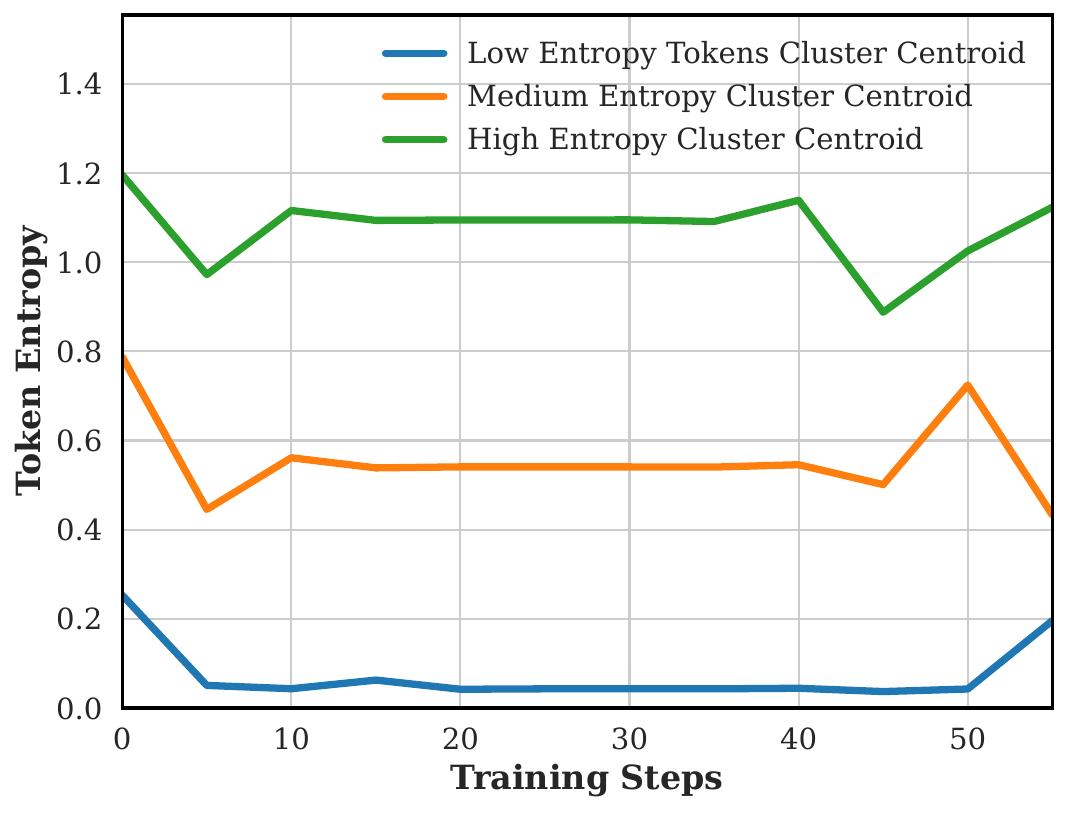}
        \caption{DeepSeek-R1-Distill-Llama-8B, Ent}
    \end{subfigure}\hfill
    \begin{subfigure}{0.32\textwidth}
        \includegraphics[width=\linewidth]{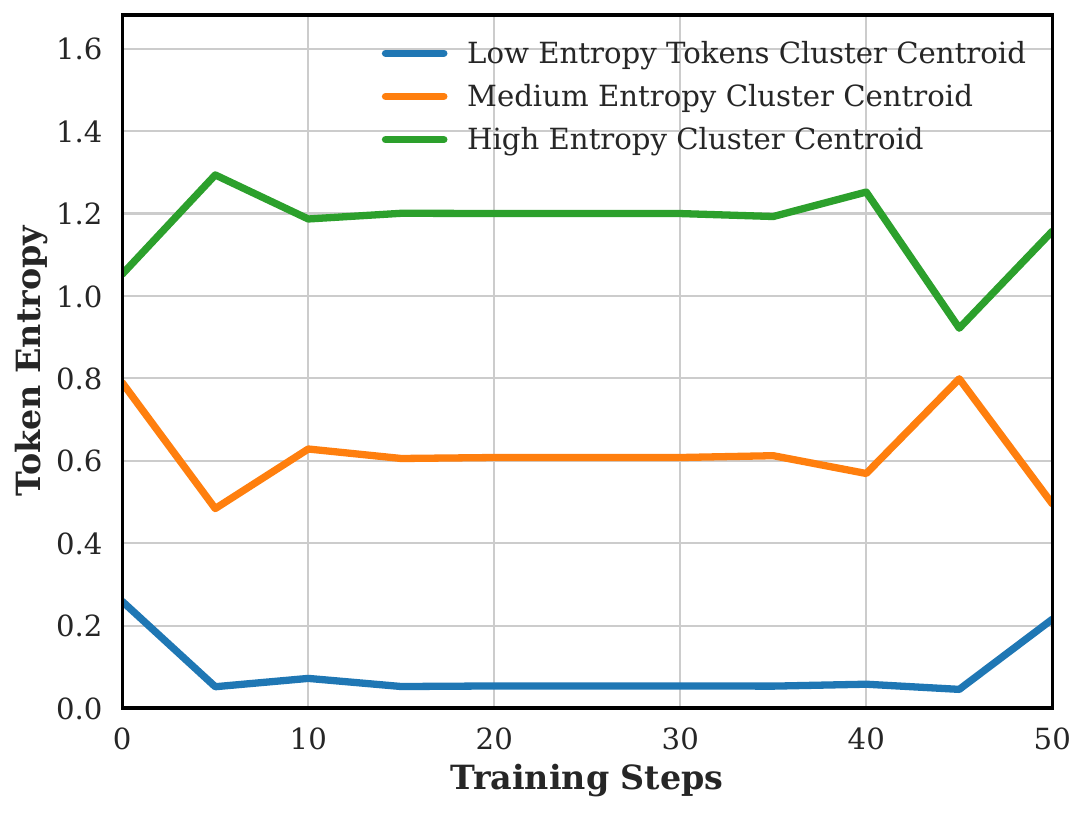}
        \caption{DeepSeek-R1-Distill-Llama-8B, LP}
    \end{subfigure}

    \vspace{0.3cm}

    \begin{subfigure}{0.32\textwidth}
        \includegraphics[width=\linewidth]{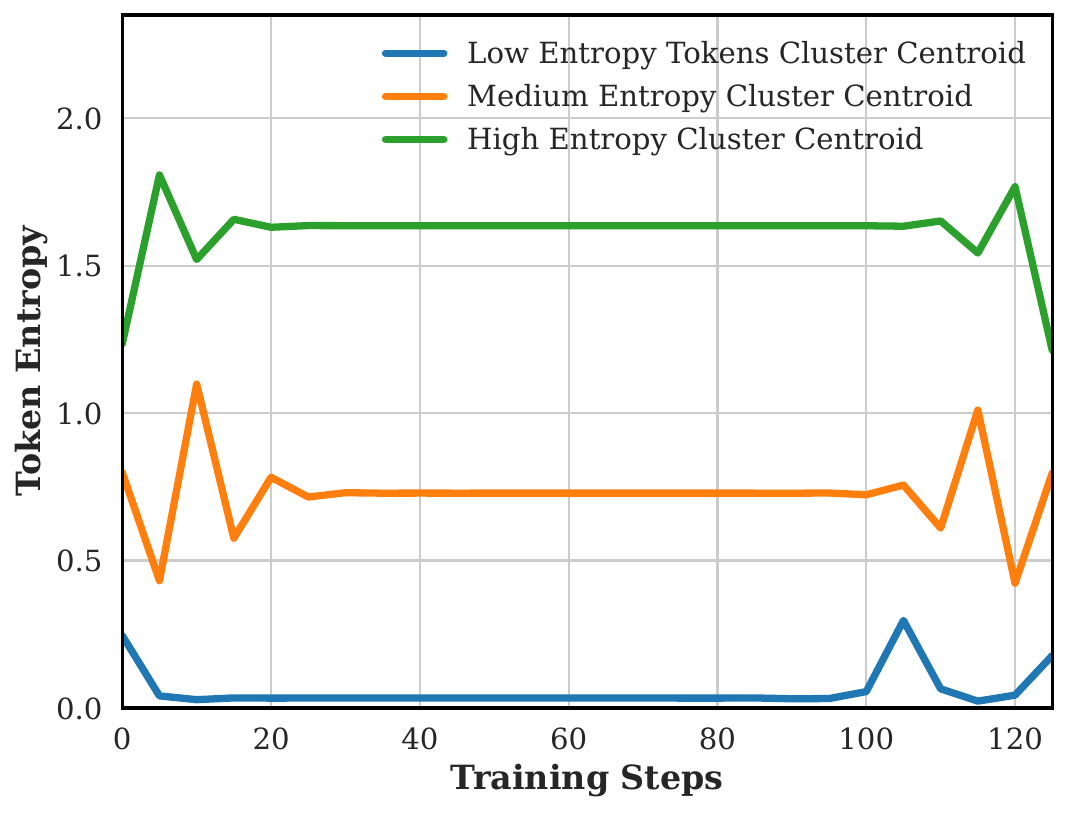}
        \caption{Llama-3.1-8B-Instruct, GRPO}
    \end{subfigure}\hfill
    \begin{subfigure}{0.32\textwidth}
        \includegraphics[width=\linewidth]{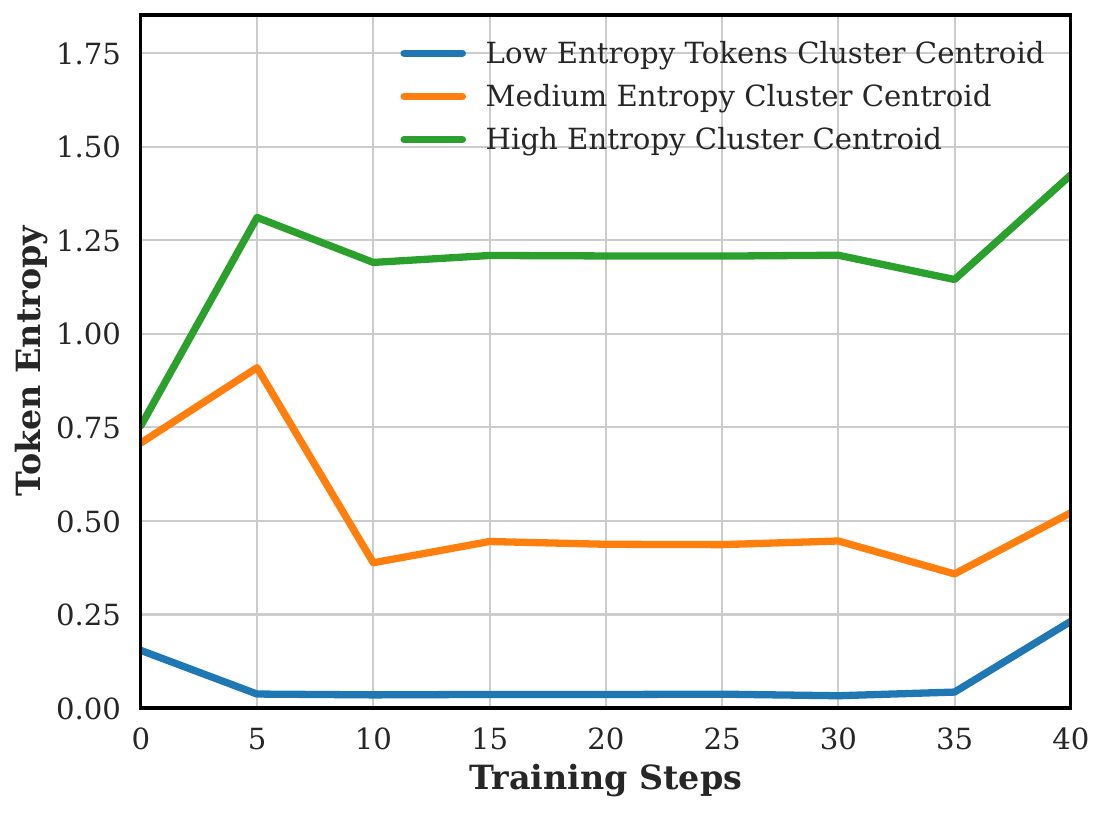}
        \caption{Llama-3.1-8B-Instruct, Ent}
    \end{subfigure}\hfill
    \begin{subfigure}{0.32\textwidth}
        \includegraphics[width=\linewidth]{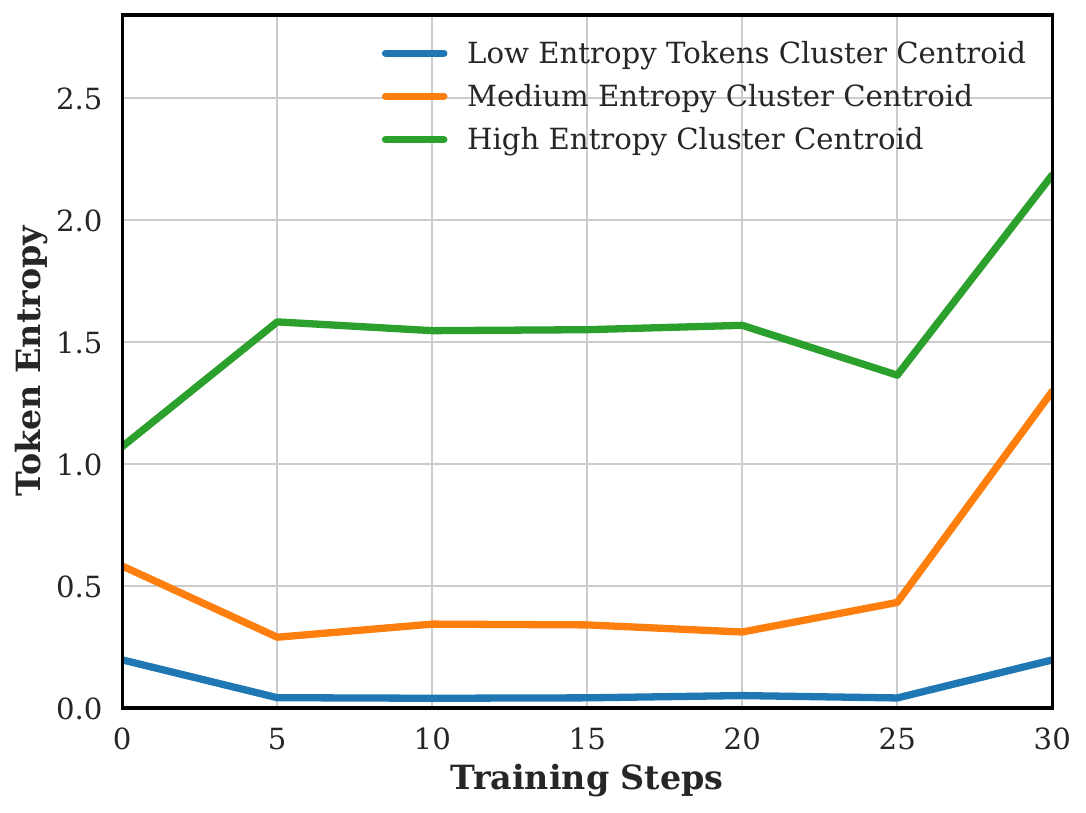}
        \caption{Llama-3.1-8B-Instruct, LP}
    \end{subfigure}

    \caption{\textbf{Token-entropy trajectory clusters with $K=3$.} Each configuration is represented by low-, mid-, and high-entropy trajectory centroids. This setting provides a finer partition than $K=2$.}
    \label{fig:comprehensive_entropy_clusters_k3}
\end{figure*}

Figure~\ref{fig:comprehensive_entropy_clusters_k2} presents the $K=2$ clusters across all models and objectives. Figure~\ref{fig:comprehensive_entropy_clusters_k3} presents the corresponding $K=3$ clusters. Both settings separate trajectories with different entropy levels and temporal patterns. The geometric analysis is therefore evaluated under both resolutions.

\subsection{Post-hoc Token Interpretation}
\label{sec:appendix_cok_semantics}

Representative-token inspection 
\Rev{CR-token-table-ref}{(Table~\ref{tab:comprehensive_token_semantics})}%
{} provides a qualitative interpretation of the $K=3$ entropy groups. The low-entropy groups contain more arithmetic and formatting tokens, the mid-entropy groups contain more transition tokens, and the high-entropy groups contain more uncertainty and planning tokens. These patterns motivate the labels ``Execution,'' ``Logic,'' and ``Thinking.'' The labels are post-hoc interpretations and are not used in the geometric analysis.

\begin{table*}[t]
    \centering
    \small
    \resizebox{\textwidth}{!}{
    \begin{tabular}{lll p{6.5cm} p{6.5cm}}
        \toprule
        \textbf{Entropy Group} & \textbf{Model} & \textbf{Method} & \textbf{Closest-to-Centroid Tokens} & \textbf{Frequent Neighborhood Tokens} \\
        \midrule

        \multirow{12}{*}{\shortstack[l]{\textbf{Low Entropy}\\ \textit{Execution}}}
        & \multirow{3}{*}{DeepSeek-R1-Distill-Llama-8B}
          & GRPO & \texttt{)/}, \texttt{16}, \texttt{|}, \texttt{y} & \texttt{by}, \texttt{0}, \texttt{3}, \texttt{1} \\
        & & Ent & \texttt{\textbackslash}, \texttt{k}, \texttt{]\textbackslash n}, \texttt{than} & \texttt{\textbackslash\textbackslash}, \texttt{]\textbackslash n}, \texttt{**}, \texttt{\^{}} \\
        & & LP & \texttt{]\textbackslash n}, \texttt{c}, \texttt{=}, \texttt{equation} & \texttt{of}, \texttt{=}, \texttt{,}, \texttt{to} \\
        \cmidrule{2-5}
        & \multirow{3}{*}{Llama-3.1-8B-Instruct}
          & GRPO & \texttt{b}, \texttt{x}, \texttt{+}, \texttt{45} & \texttt{1}, \texttt{6}, \texttt{3}, \texttt{4} \\
        & & Ent & \texttt{\$}, \texttt{25}, \texttt{for}, \texttt{24} & \texttt{-}, \texttt{\}}, \texttt{\$}, \texttt{=} \\
        & & LP & \texttt{)\$}, \texttt{*}, \texttt{+}, \texttt{is} & \texttt{of}, \texttt{\$}, \texttt{to}, \texttt{is} \\
        \cmidrule{2-5}
        & \multirow{3}{*}{Qwen3-1.7B}
          & GRPO & \texttt{+}, \texttt{*}, \texttt{3}, \texttt{(} & \texttt{to}, \texttt{=}, \texttt{be}, \texttt{)} \\
        & & Ent & \texttt{example}, \texttt{y}, \texttt{k}, \texttt{\^{}} & \texttt{be}, \texttt{=}, \texttt{)}, \texttt{to} \\
        & & LP & \texttt{example}, \texttt{+}, \texttt{))}, \texttt{*} & \texttt{be}, \texttt{)}, \texttt{=}, \texttt{*} \\
        \cmidrule{2-5}
        & \multirow{3}{*}{Qwen3-8B}
          & GRPO & \texttt{8}, \texttt{B}, \texttt{b}, \texttt{error} & \texttt{by}, \texttt{)}, \texttt{=}, \texttt{be} \\
        & & Ent & \texttt{\^{}}, \texttt{\textbackslash}, \texttt{\$}, \texttt{k} & \texttt{=}, \texttt{me}, \texttt{\textbackslash}, \texttt{\$\$} \\
        & & LP & \texttt{y}, \texttt{by}, \texttt{(-}, \texttt{than} & \texttt{=}, \texttt{by}, \texttt{)}, \texttt{|} \\

        \midrule

        \multirow{12}{*}{\shortstack[l]{\textbf{Mid Entropy}\\ \textit{Logic}}}
        & \multirow{3}{*}{DeepSeek-R1-Distill-Llama-8B}
          & GRPO & \texttt{Then}, \texttt{that}, \texttt{here}, \texttt{compute} & \texttt{Let}, \texttt{So}, \texttt{but}, \texttt{that} \\
        & & Ent & \texttt{again}, \texttt{such}, \texttt{about}, \texttt{correct} & \texttt{?}, \texttt{here}, \texttt{So}, \texttt{complicated} \\
        & & LP & \texttt{these}, \texttt{you}, \texttt{says}, \texttt{an} & \texttt{but}, \texttt{'m}, \texttt{Wait}, \texttt{?} \\
        \cmidrule{2-5}
        & \multirow{3}{*}{Llama-3.1-8B-Instruct}
          & GRPO & \texttt{found}, \texttt{den}, \texttt{follows}, \texttt{some} & \texttt{help}, \texttt{need}, \texttt{).}, \texttt{scenario} \\
        & & Ent & \texttt{Simpl}, \texttt{at}, \texttt{do}, \texttt{will} & \texttt{Now}, \texttt{let}, \texttt{In}, \texttt{one} \\
        & & LP & \texttt{Sub}, \texttt{add}, \texttt{states}, \texttt{answer} & \texttt{has}, \texttt{).}, \texttt{all}, \texttt{at} \\
        \cmidrule{2-5}
        & \multirow{3}{*}{Qwen3-1.7B}
          & GRPO & \texttt{don}, \texttt{Yes}, \texttt{let}, \texttt{what} & \texttt{But}, \texttt{we}, \texttt{there}, \texttt{Wait} \\
        & & Ent & \texttt{But}, \texttt{No}, \texttt{'s}, \texttt{Since} & \texttt{But}, \texttt{Therefore}, \texttt{So}, \texttt{Wait} \\
        & & LP & \texttt{Wait}, \texttt{But}, \texttt{?}, \texttt{I} & \texttt{But}, \texttt{?}, \texttt{Wait}, \texttt{this} \\
        \cmidrule{2-5}
        & \multirow{3}{*}{Qwen3-8B}
          & GRPO & \texttt{Yes}, \texttt{these}, \texttt{another}, \texttt{which} & \texttt{?}, \texttt{but}, \texttt{Wait}, \texttt{it} \\
        & & Ent & \texttt{Now}, \texttt{check}, \texttt{Similarly} & \texttt{Wait}, \texttt{?}, \texttt{this}, \texttt{Let} \\
        & & LP & \texttt{).\textbackslash n\textbackslash n}, \texttt{about}, \texttt{them}, \texttt{recall} & \texttt{Let}, \texttt{but}, \texttt{:\textbackslash n\textbackslash n}, \texttt{Then} \\

        \midrule

        \multirow{12}{*}{\shortstack[l]{\textbf{High Entropy}\\ \textit{Thinking}}}
        & \multirow{3}{*}{DeepSeek-R1-Distill-Llama-8B}
          & GRPO & \texttt{If}, \texttt{Also}, \texttt{since}, \texttt{perhaps} & \texttt{maybe}, \texttt{let}, \texttt{perhaps}, \texttt{wait} \\
        & & Ent & \texttt{wait}, \texttt{Hmm}, \texttt{write}, \texttt{maybe} & \texttt{maybe}, \texttt{perhaps}, \texttt{wait}, \texttt{But} \\
        & & LP & \texttt{think}, \texttt{Hmm}, \texttt{when}, \texttt{maybe} & \texttt{maybe}, \texttt{perhaps}, \texttt{wait}, \texttt{Hmm} \\
        \cmidrule{2-5}
        & \multirow{3}{*}{Llama-3.1-8B-Instruct}
          & GRPO & \texttt{Given}, \texttt{First}, \texttt{Calcul}, \texttt{Let} & \texttt{Calcul}, \texttt{Let}, \texttt{When}, \texttt{Thus} \\
        & & Ent & \texttt{now}, \texttt{Let}, \texttt{determine}, \texttt{'re} & \texttt{To}, \texttt{The}, \texttt{will}, \texttt{express} \\
        & & LP & \texttt{assume}, \texttt{From}, \texttt{So}, \texttt{First} & \texttt{First}, \texttt{start}, \texttt{denote}, \texttt{To} \\
        \cmidrule{2-5}
        & \multirow{3}{*}{Qwen3-1.7B}
          & GRPO & \texttt{???}, \texttt{probably}, \texttt{under}, \texttt{smarter} & \texttt{wait}, \texttt{???}, \texttt{Divide}, \texttt{)"} \\
        & & Ent & \texttt{Thus}, \texttt{confirm}, \texttt{larger}, \texttt{relationship} & \texttt{perhaps}, \texttt{???}, \texttt{Hence}, \texttt{relationship} \\
        & & LP & \texttt{adding}, \texttt{assuming}, \texttt{regardless}, \texttt{What} & \texttt{perhaps}, \texttt{Thus}, \texttt{tackle}, \texttt{Hence} \\
        \cmidrule{2-5}
        & \multirow{3}{*}{Qwen3-8B}
          & GRPO & \texttt{Well}, \texttt{Which}, \texttt{setting}, \texttt{lead} & \texttt{perhaps}, \texttt{Thus}, \texttt{consider}, \texttt{Hence} \\
        & & Ent & \texttt{analyze}, \texttt{contrad}, \texttt{Hmm} & \texttt{perhaps}, \texttt{maybe}, \texttt{suppose} \\
        & & LP & \texttt{Thus}, \texttt{express}, \texttt{Sup}, \texttt{Therefore} & \texttt{However}, \texttt{maybe}, \texttt{If}, \texttt{perhaps} \\

        \bottomrule
    \end{tabular}
    }
    \caption{\textbf{Representative tokens in the $K=3$ entropy groups.} The semantic labels are post-hoc interpretations based on tokens near each cluster centroid.}
    \label{tab:comprehensive_token_semantics}
\end{table*}

\subsection{
\RevNoLine{ADD-cluster-head}{Implementation Details}%
}
\label{sec:ADD-cluster-head}
\Rev{ADD-cluster}{At each observed checkpoint, the entropy value for a prompt--token pair is averaged over its occurrences in the eight responses. The resulting variable-length value sequences are passed directly to DTW clustering. Clustering uses one initialization, random state 42, at most 10 clustering iterations, at most 100 barycenter iterations and inertia tolerance $10^{-6}$. Here $x_i$ denotes a trajectory, $N$ the number of trajectories, and $H(x)$ the token entropy in a response. These definitions specify the quantities used in the coordinate and clustering calculations.}%


\section{Additional Compression Evidence and Controls}
\label{sec:appendix_controls}

\subsection{Direct Accuracy--Length Trajectory}
\label{sec:appendix_accuracy_length}

To verify that the fixed-budget result corresponds to response shortening rather than only a change in truncation frequency, we run LP on DeepSeek-R1-Distill-Qwen-1.5B with a 16K generation budget and evaluate AIME24 throughout training. Table~\ref{tab:accuracy_length_16k} reports the complete trajectory. Average response length decreases from 11309.9 tokens at step 0 to 8015.4 at step 65, while accuracy remains nearly unchanged from 25.3\% to 25.5\%. Continued shortening is then accompanied by accuracy collapse. This directly documents a transient response-compression phase without implying preservation of the original reasoning trace.

\begin{table*}[t]
\centering
\begin{minipage}[t]{0.59\textwidth}
\vspace{0pt}
\setlength{\tabcolsep}{5pt}
\centering
\small
\begin{tabular}{rrr@{\hspace{2em}}rrr}
\toprule
\textbf{Step} & \textbf{Accuracy} & \textbf{Avg. Length} &
\textbf{Step} & \textbf{Accuracy} & \textbf{Avg. Length} \\
\midrule
0 & 0.253 & 11309.9 & 65 & 0.255 & 8015.4 \\
5 & 0.250 & 11481.5 & 70 & 0.245 & 7396.4 \\
10 & 0.262 & 11584.8 & 75 & 0.238 & 6648.0 \\
15 & 0.268 & 11270.3 & 80 & 0.209 & 6408.6 \\
20 & 0.265 & 10961.4 & 85 & 0.180 & 4195.0 \\
25 & 0.264 & 10666.5 & 90 & 0.083 & 954.4 \\
30 & 0.302 & 10317.3 & 95 & 0.065 & 800.2 \\
35 & 0.262 & 10245.8 & 100 & 0.065 & 548.2 \\
40 & 0.253 & 9956.4 & 105 & 0.050 & 357.8 \\
45 & 0.262 & 9359.6 & 110 & 0.021 & 248.6 \\
50 & 0.277 & 8688.9 & 115 & 0.006 & 37.5 \\
55 & 0.259 & 8324.9 & 120 & 0.000 & 5.3 \\
60 & 0.273 & 8126.7 & 125 & 0.000 & 0.0 \\
\bottomrule
\end{tabular}
\caption{\textbf{Accuracy--length trajectory under LP with a 16K response budget.} Results are reported for DeepSeek-R1-Distill-Qwen-1.5B on AIME24.}
\label{tab:accuracy_length_16k}
\end{minipage}\hfill
\begin{minipage}[t]{0.39\textwidth}
\vspace{0pt}
\centering
\small
\begin{tabular}{rr@{\hspace{3em}}rr}
\toprule
\textbf{Step} & \textbf{Accuracy} & \textbf{Step} & \textbf{Accuracy} \\
\midrule
0 & 0.0208 & 95 & 0.2250 \\
5 & 0.0208 & 100 & 0.2583 \\
10 & 0.0167 & 105 & 0.3000 \\
15 & 0.0208 & 110 & 0.2667 \\
20 & 0.0208 & 115 & 0.2833 \\
25 & 0.0250 & 120 & 0.3167 \\
30 & 0.0333 & 125 & 0.3125 \\
35 & 0.0333 & 130 & 0.3250 \\
40 & 0.0375 & 135 & 0.3333 \\
45 & 0.0458 & 140 & 0.3208 \\
50 & 0.0750 & 145 & 0.3625 \\
55 & 0.0875 & 150 & 0.3833 \\
60 & 0.1208 & 155 & 0.3750 \\
65 & 0.1125 & 160 & 0.3458 \\
70 & 0.1542 & 165 & 0.3875 \\
75 & 0.1875 & 170 & 0.3250 \\
80 & 0.1792 & 175 & 0.0042 \\
85 & 0.2000 & 180 & 0.0000 \\
90 & 0.2333 &  &  \\
\bottomrule
\end{tabular}
\caption{\textbf{LP control 
\RevNoLine{CR-control-caption}{for Qwen3-8B}%
{} with global token-mean loss aggregation.} The validation trajectory retains a transient improvement followed by collapse.}
\label{tab:token_mean_control}
\end{minipage}
\end{table*}

\subsection{Reward-Alignment and Loss-Aggregation Controls}
\label{sec:appendix_reward_controls}

The main experiments use sequence-mean/token-mean aggregation,
\begin{equation}
\mathcal{L}_{\mathrm{seq}}
=\frac{1}{B}\sum_{i=1}^B\frac{1}{T_i}\sum_{t=1}^{T_i}\ell_{i,t},
\end{equation}
where $B$ is the number of responses, $T_i$ is the number of valid response tokens, and $\ell_{i,t}$ is the clipped token-level policy-gradient loss. This objective gives equal total weight to each response and therefore gives larger per-token weight to shorter responses. We rerun LP 
\Rev{CR-control-model}{on Qwen3-8B}%
{} with global token-mean aggregation,
\begin{equation}
\mathcal{L}_{\mathrm{token}}
=\frac{\sum_{i=1}^B\sum_{t=1}^{T_i}\ell_{i,t}}
{\sum_{i=1}^B T_i},
\end{equation}
which assigns equal weight to every valid response token. All other settings are unchanged.

Table~\ref{tab:token_mean_control} reports the validation trajectory. Accuracy increases from 2.08\% at step 0 to 38.75\% at step 165 and then collapses to 0\% at step 180. The improve-then-collapse pattern therefore persists after removing the original per-response inverse-length weighting.

We also shuffle LP rewards across responses within each optimization batch before advantage normalization, preserving the reward distribution while breaking the response--reward alignment. The shuffled-LP run still exhibits a transient accuracy peak before collapse. This control indicates that the transient gain should not be attributed solely to LP identifying semantically better reasoning. We therefore interpret LP and Ent together with their optimizer configuration as experimental conditions and avoid a reward-specific causal claim.

\end{document}